\documentclass[lettersize,journal,twoside]{IEEEtran}
\usepackage{xcolor}
\usepackage{amsmath}
\usepackage{amsfonts}
\usepackage{algorithmic}
\usepackage{amssymb}
\usepackage{ulem}
\usepackage{mathtools}
\usepackage{mathrsfs}
\usepackage{multirow}
\usepackage{indentfirst}
\usepackage{booktabs}
\usepackage{tabularx}
\usepackage{threeparttable}
\usepackage[font=small]{caption}
\usepackage[font=small]{subcaption}
\usepackage{array}
\usepackage{textcomp}
\usepackage{stfloats}
\usepackage{url}
\usepackage{verbatim}
\usepackage{graphicx}
\usepackage{amsthm}
\usepackage{cite}
\usepackage{hyperref}
\usepackage{diagbox}
\usepackage{makecell}
\usepackage{enumitem}
\usepackage{bm}

\def\BibTeX{{\rm B\kern-.05em{\sc i\kern-.025em b}\kern-.08em
    T\kern-.1667em\lower.7ex\hbox{E}\kern-.125emX}}
\usepackage{balance}

\usepackage{amsthm}
\theoremstyle{definition}
\newtheorem{theorem}{\bf Theorem}
\newtheorem{proposition}{\bf Proposition}
\newtheorem{definition}{\bf Definition}
\newtheorem{lemma}{\bf Lemma}

\newtheorem{remark}{\bf Remark}
\usepackage{pifont}
\newcommand{\cmarkbold}{\ding{52}}%
\newcommand{\xmark}{\ding{53}}%

\usepackage{xcolor}

\begin{document}
\title{QP Chaser: Polynomial Trajectory Generation for Autonomous Aerial Tracking}

\author{Yunwoo Lee$^{1}$, Jungwon Park$^{2}$, Seungwoo Jung$^{3}$, Boseong Jeon$^{4}$, Dahyun Oh$^{3}$, and H. Jin Kim$^{3}$
\thanks{Manuscript received 02 December 2023; revised  11 July 2024,  6 November 2024, 22 May 2025, 1 Sep  2025; accepted 30 October 2025. This work was supported by the Unmanned Vehicles Core Technology Research and Development Program through the National Research Foundation of Korea (NRF) and Unmanned Vehicle Advanced Research Center (UVARC) funded
by the Ministry of Science and ICT under Grant NRF-2020M3C1C1A010864. \textit{(Corresponding author: H. Jin Kim)}}
\thanks{$^{1}$The author is with the Robotics Institute, Carnegie Mellon University, Pittsburgh PA, United States, (e-mail: yunwool@andrew.cmu.edu).}\
\thanks{$^{2}$The author is with the Department of Mechanical System Design Engineering, Seoul National University of Science and Technology, Seoul, South Korea (e-mail: jungwonpark@seoultech.ac.kr).}
\thanks{$^{3}$The authors are with the Department of Aerospace Engineering, Seoul National University, Seoul, South Korea (e-mail: tmddn833@snu.ac.kr, qlass33@snu.ac.kr, hjinkim@snu.ac.kr).}
\thanks{$^{4}$The author is with Samsung Research, Samsung Electronics, Seoul, South Korea (e-mail: junbs95@gmail.com).}
}
\markboth{IEEE Transactions on Automation Science and Engineering, Accepted October, 2025}{Lee \MakeLowercase{\textit{et al.}}: QP Chaser: Polynomial Trajectory Generation for Autonomous Aerial Tracking}

\maketitle

\begin{abstract}
Maintaining the visibility of the target is one of the major objectives of aerial tracking missions.
This paper proposes a target-visible trajectory planning pipeline using quadratic programming (QP).
Our approach can handle various tracking settings, including 1) single- and dual-target following and 2) both static and dynamic environments, unlike other works that focus on a single specific setup.
In contrast to other studies that fully trust the predicted trajectory of the target and consider only the visibility of the target’s center, our pipeline considers error in target path prediction and the entire body of the target to maintain the target visibility robustly. 
First, a prediction module uses a sample-check strategy to quickly calculate the reachable areas of moving objects, which represent the areas their bodies can reach, considering obstacles.
Subsequently, the planning module formulates a single QP problem, considering path homotopy, to generate a tracking trajectory that maximizes the visibility of the target's reachable area among obstacles.
The performance of the planner is validated in multiple scenarios, through high-fidelity simulations and real-world experiments.
\end{abstract}

\def\abstractname{Note to Practitioners}
\begin{abstract}
This paper proposes an aerial target tracking framework applicable to both single- and dual-target following missions. 
This paper proposes the prediction of the reachable area of moving objects and the generation of a target-visible trajectory, both of which are computed in real-time.
Since the proposed planner considers the possible reach area of moving objects, the generated trajectory of the drone is robust to the prediction inaccuracy in terms of the target visibility. Our system can be utilized in crowded environments with multiple moving objects and extended to multiple-target scenarios. We extensively validate our system through several real-world experiments to show practicality.
\end{abstract}

\IEEEpeerreviewmaketitle

\section{Introduction}
\label{sec:introduction}
\IEEEPARstart{V}{ision}-aided multi-rotors \cite{history1,history2,sphere} are widely employed in applications such as surveillance \cite{covert_tracking} and cinematography \cite{Bonatti}, and autonomous target chasing is essential in such tasks. In target-chasing missions, various situations exist in which a single drone has to handle both single- and multi-target scenarios without occlusion. 
For example, in film shooting, there are scenes in which one or several actors should be shot in a single take, without being visually disturbed by structures in the shooting set. Moreover, the occlusion of main actors by background actors is generally prohibited. 
Additionally, autonomous shooting can be used in sporting events to provide aerial views of athletes' dynamic play. Similarly, the athletes' movement should not be obstructed by crowds. Therefore, a tracking strategy capable of handling both single and multiple targets among static and dynamic obstacles can benefit various scenarios in chasing tasks.

Despite great attention and research over the recent decade, aerial target tracking remains a demanding task for the following reasons.
First, the motion generator in the chasing system needs to account for visibility obstruction caused by environmental structures and the limited camera field-of-view (FOV), in addition to typical considerations in UAV motion planning, such as collision avoidance, quality of the flight path, and dynamical limits. 
Since the sudden appearance of unforeseen obstacles can cause target occlusion, a trajectory that satisfies all these considerations needs to be generated quickly.
Second, it is difficult to forecast accurate future paths of dynamic objects due to perceptual error from sensors, surrounding environmental structures, and unreliable estimation of intentions. Poor predictions can negatively impact trajectory planning performance and lead to tracking failures.
\begin{figure}[t!]
    \centering
    \begin{subfigure}[b]{0.22\textwidth}
        \label{subfig:thumbnail_single}
        \centering
        \includegraphics[width=\textwidth]{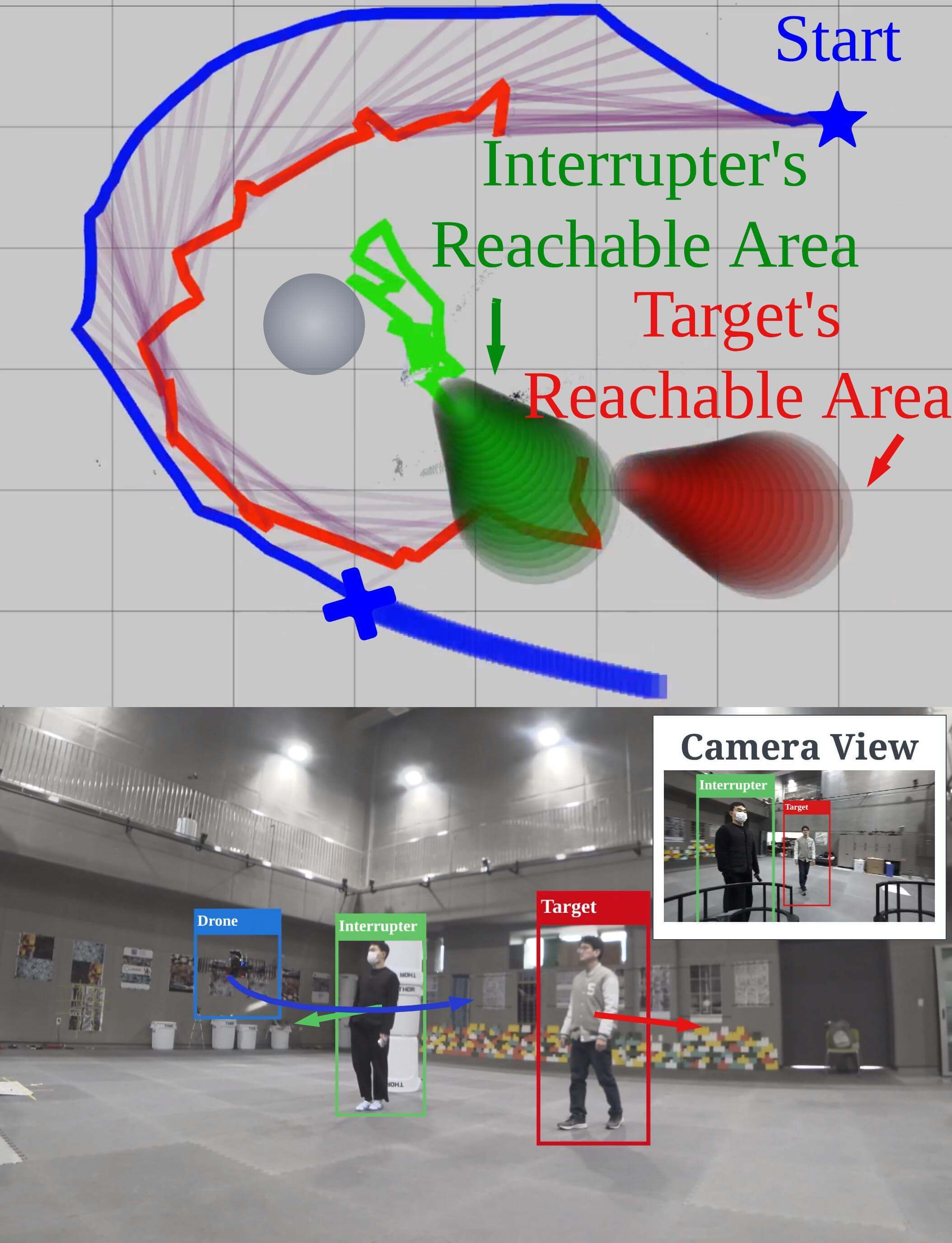}
        \caption{Single-target tracking}
    \end{subfigure}
    \hfill
        \begin{subfigure}[b]{0.22\textwidth}
        \label{subfig:thumbnail_dual}
        \centering
        \includegraphics[width=\textwidth]{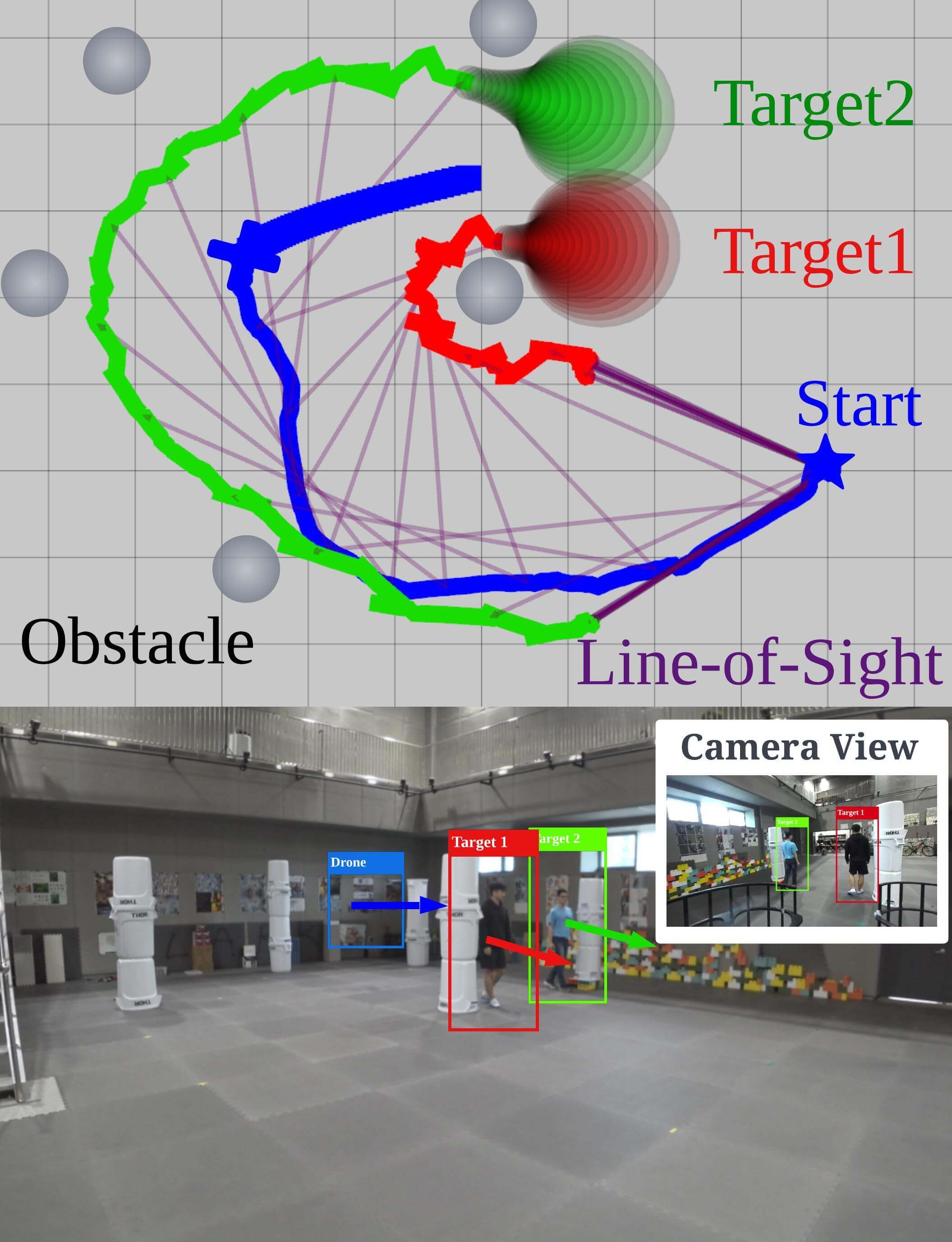}
        \caption{Dual-target tracking}
    \end{subfigure}
    \caption{Target tracking mission in a realistic situation. \textbf{(a)}: A target (red) moves in an indoor arena, and a dynamic obstacle (green) interrupts the visibility of the target. \textbf{(b)}: Two targets (red and green) move among stacked bins (grey). A chaser drone (cross) generates a trajectory (blue) consistently tracking the targets without occlusion.}
    \label{fig:thumbnail}
    \vspace{-4mm}
\end{figure}

In order to address the factors above, we propose a real-time single- and dual-target tracking strategy that 1) can be adopted in both static and dynamic environments and 2) enhances the target visibility against prediction error, as illustrated in Fig. \ref{fig:thumbnail}. The proposed method consists of two parts: \textit{prediction} and \textit{chasing}. 
In the \textit{prediction} module, we calculate reachable areas of moving objects, taking into account the obstacle configuration. 
For efficient calculations, we use a sample-check strategy. First, we sample motion primitives representing possible trajectories, generated by quadratic programming (QP), which is solved in closed form. Then, collisions between primitives and obstacles are checked while leveraging the properties of Bernstein polynomials. We then calculate reachable areas enclosing the collision-free primitives.

In the \textit{chasing} module, we design a chasing trajectory via a single QP. QP is known for being solvable in polynomial time and guaranteeing global optimality. We adopted this method because the optimization was solved within tens of milliseconds during numerous tests. The key idea of our trajectory planning is to define a target-visible region (TVR), a time-dependent spatial set that keeps the visibility of targets. Two essential factors of TVR enhance the target visibility. First, analyzing the homotopy classes of the targets' path helps avoid situations where target occlusion inescapably occurs. Second, TVR is designed to maximize the visible area of the target's reachable area, considering potential visual obstructions from obstacles and ensuring robust target visibility against the prediction error. 
 Moreover, by allowing the entire target body to be viewed instead of specific points, the TVR enhances the target visibility.
Additionally, to handle dual-target scenarios, we define an area where two targets can be simultaneously viewed with a limited FOV camera. Table \ref{table:comparison_with_sota} presents the comparison with the relevant works, and our main contributions are summarized as follows.
\begin{itemize}
    \item We propose a QP-based trajectory planning framework capable of single and dual target-chasing missions amidst static and dynamic obstacles, in contrast with existing works that address either single-target scenarios or static environments only.
    \item (\textit{Prediction}) We propose an efficient sample-check-based method to compute reachable areas of moving objects, leveraging Bernstein polynomials. This method reflects both static and dynamic obstacles, which contrasts with other works that do not fully consider the obstacle configurations in motion predictions.
    \item (\textit{Chasing}) We propose a target-visible trajectory generation method by designing a target-visible region (TVR), in which the target visibility is robustly maintained. This approach considers 1) the path homotopy, 2) the entire body of targets, and 3) the prediction error of moving objects. This contrasts with other works that fully trust potentially erroneous path predictions.
\end{itemize}

The remainder of this paper is arranged as follows. We review the relevant references in Section \ref{sec:related_works}, and the relationship between the target visibility and the path homotopy is studied in Section \ref{sec:preliminary}. The problem statement and the pipeline of the proposed system are presented in Section \ref{sec:overview}, and Section \ref{sec:object_trajectory_prediction} describes the prediction of the reachable areas of moving objects. Section \ref{sec:chasing_trajectory_generation} describes TVR, designs the reference trajectory for the chasing drone, and completes QP formulation. The validation of the proposed pipeline is demonstrated with high-fidelity simulations and real-world experiments in Section \ref{sec:validation}. 

\section{Related Works}
\label{sec:related_works}
In this section, we review UAV trajectory planning research on target tracking.
\subsection{Single Target Chasing in Empty Space}
\label{subsec:related_works_empty_space}
Research on motion planning to follow a single target in a empty space has been widely studied. \cite{control1,control2,control3} propose vision-based controllers to prevent the target from moving out of the camera's FOV.  \cite{empty_learning_1} and \cite{empty_learning_2} propose viewpoint planners for high-quality video creation and accurate 2D human pose estimation, respectively. Although all these works improve target perception, they are only applicable in obstacle-free environments, making them unsuitable for real-world scenarios.
\begin{table}[t!]
\vspace{2mm}
\caption{Comparison with the State-of-the-Art Algorithms.}
\label{table:comparison_with_sota}
    \begin{center}
    \begin{tabularx}{\linewidth}{c|X|X|X|X|X}
    \hline
    \multirow{3}{*}{\centering Method} & \multicolumn{2}{c|}{Scenarios} & \multicolumn{1}{c|}{Prediction}&\multicolumn{2}{c}{Planning} \tabularnewline
    \cline{2-6}
     &\centering Dual-target &\centering Dynamic obstacles & \centering Environ. informed & \centering PE-aware &\centering WB of target \tabularnewline
    \hline
    \cite{fail-safe} & \centering \xmark & \centering \xmark & \centering $\checkmark$ & \centering \xmark & \centering \xmark \tabularnewline
 \cite{Nageli} & \centering $\checkmark$ & \centering $\checkmark$ & \centering \xmark &\centering \xmark &\centering \checkmark \tabularnewline
    \cite{occlusion_cost} & \centering \xmark & \centering $\checkmark$ & \centering \xmark & \centering \xmark & \centering \xmark \tabularnewline
    \cite{multi_convex} & \centering \xmark & \centering $\checkmark$ & \centering \xmark & \centering \xmark & \centering \xmark \tabularnewline 
        \cite{ijcas} & \centering \xmark & \centering $\checkmark$ & \centering $\checkmark$ &\centering \xmark &\centering \xmark \tabularnewline
    \cite{access} & \centering $\checkmark$ & \centering \xmark & \centering $\checkmark$ & \centering \xmark & \centering \xmark \tabularnewline
\cite{cine-mpc} & \centering $\checkmark$ & \centering \xmark & \centering \xmark & \centering \xmark & \centering $\checkmark$ \tabularnewline
    \hline  
    \textbf{Ours} & \centering \cmarkbold & \centering \cmarkbold & \centering \cmarkbold & \centering \cmarkbold  & \centering \cmarkbold \tabularnewline
    \hline
    \end{tabularx}
    \end{center}
\footnotesize{
    $\checkmark$ means that the algorithm explicitly considers the corresponding item. (PE: Perception-error, and WB: Whole body)
    }
\vspace{-2mm}
\end{table}
\subsection{Single Target Chasing in Static Obstacle Environments}
\label{subsec: visibility consideration}
There have been various studies that take the target visibility into account in single-target chasing in static environments. 
\subsubsection{Chasing trajectory planning}
\label{subsubsec:single_static_planning}
\cite{sphere} and \cite{Bonatti} design cost functions that penalize target occlusion by obstacles, while \cite{svpto} designs an environmental complexity cost function to adjust the distance between a target and a drone, implicitly reducing the probability of occlusion.
These methods combine the functions with several other objective functions related to actuation efficiency and collision avoidance.
Such conflicting objectives can yield a sub-optimal solution that compromises tracking motion. 
Therefore, the visibility of the target is not ensured.
In contrast, our approach ensures the target visibility by incorporating it as a hard constraint in the QP problem, which can be solved quickly.
On the other hand, there are works that explicitly consider visibility as hard constraints in optimization.
\cite{elastic} designs sector-shaped visible region and \cite{auto-filmer} designs select the view region among annulus sectors centered around the target. They use their target-visible regions in unconstrained optimization to generate a chasing trajectory.
However, these works focus on the visibility of the center of a target, which may result in partial occlusion. In contrast, our method considers the visibility of the entire body of the target and accounts for the prediction error, thus enhancing robustness in target tracking. 

\subsubsection{Moving target prediction}
\label{subsubsec:single_static_prediction}
Also, there are works that handle target path prediction for aerial tracking. \cite{fast-tracker1} and \cite{fast-tracker2} use past target observations to predict the future trajectory via polynomial regression. On the other hand, \cite{target_prediction_gru} propose prediction method leveraging advantages of Gaussian process regression and gated recurrent unit. However, their methods can generate erroneous results by insufficiently considering surroundings, resulting in path conflicts with obstacles.
With inaccurate target path prediction, planners may not produce effective chasing paths and fail to chase a target without occlusion. 
Some works \cite{fail-safe} and \cite{boseong_icra} predict the future target's movements by formulating optimization problems that encourage the target to move towards free spaces.
However, these approaches are limited to static environments. In contrast, our framework predicts the target's movement considering both static and dynamic obstacles.

\subsection{Single-Target Chasing in Dynamic Environments}
\label{subsec:dynamic_scenarios}
Some studies treat tracking a target in dynamic environments.
There are works that attempt to solve the occlusion problem by approximating the future motion of both dynamic obstacles and targets using a constant velocity model. \cite{Nageli} designs a target-visibility cost that is inspired by GPU ray casting model. \cite{occlusion_cost} uses a learned occlusion model to evaluate the target visibility and tests the planner in dynamic environments. \cite{multi_convex} aims to avoid occlusion by imposing a constraint that prevents a segment connecting a target and the drone from colliding with dynamic obstacles. Since \cite{Nageli} and \cite{occlusion_cost} use the constant velocity model to predict the movements of dynamic objects, there are cases where the target and obstacles overlap, leading to incorrect target-visibility cost evaluation, which can ruin the planning results. Furthermore, the constraint in \cite{multi_convex} cannot be satisfied if the predicted path of the target traverses occupied space (e.g. obstacles), which makes the optimization problem infeasible. In contrast, our approach predicts the movements of objects by fully considering the environment and incorporates the prediction error into planning to enhance the target visibility.

On the other hand, the approach in \cite{ijcas} makes a set of polynomial motion primitives and selects the best path under safety and target visibility constraints to acquire a non-colliding target path and generate a chasing trajectory.
However, when the target and drone are near an obstacle, the planner \cite{ijcas} may select the path toward a region where occlusion is inevitable from the perspective of path homotopy, as will be shown in Section \ref{subsec:simulations}.
In contrast, our method directly applies concepts of the path homotopy to prevent inescapable occlusion.
\subsection{Aerial Tracking of More Than One Target}
\label{subsec:multiple_target_scenarios}
Aerial tracking of more than one target using a single drone has been studied in a few works. \cite{dual1} and \cite{dual2} minimize the change in the target's position in the camera image; however, they do not consider obstacles.  \cite{access} designs a dual visibility score field to handle the visibility of two targets among obstacles and a camera field-of-view (FOV) limit. Because the field is constructed heuristically, the success rate of tracking heavily depends on parameter settings. In contrast, our work handles these issues by applying them as hard constraints that must be satisfied.
\cite{cine-mpc} tracks multiple targets by simultaneously controlling the camera's intrinsic and extrinsic parameters. However, it is limited to systems where the intrinsic parameters can be controlled. 
In contrast, we ensure simultaneous observation of the targets using fixed and limited FOV cameras.
\subsection{Target Following Using Quadratic Programming}
\label{subsec:target_following_qp}
Few aerial tracking works utilize QP for chasing trajectory planning: \cite{boseong_icra} and \cite{shaojie_shen}. \cite{boseong_icra} calculates a series of viewpoints and uses QP optimization to interpolate them, but it does not ensure the target visibility when moving between the viewpoints. \cite{shaojie_shen} generates safe flight corridors along the target's path and pushes the drone's trajectory toward the safe regions, but the target visibility is not considered in its QP problem. In contrast, we formulate the target visibility constraints for the full planning horizon. Also, in conventional drone trajectory generation methods using QP, the trajectory is confined to a static half-space over time intervals. This makes it particularly difficult to keep the target visibility in the presence of dynamic obstacles. In contrast, by incorporating time-varying spatial constraints, our method allows the drone to operate within a broader region, offering an advantage in target tracking.
\begin{figure}[t!]
\centering
\includegraphics[width = 0.90\linewidth]{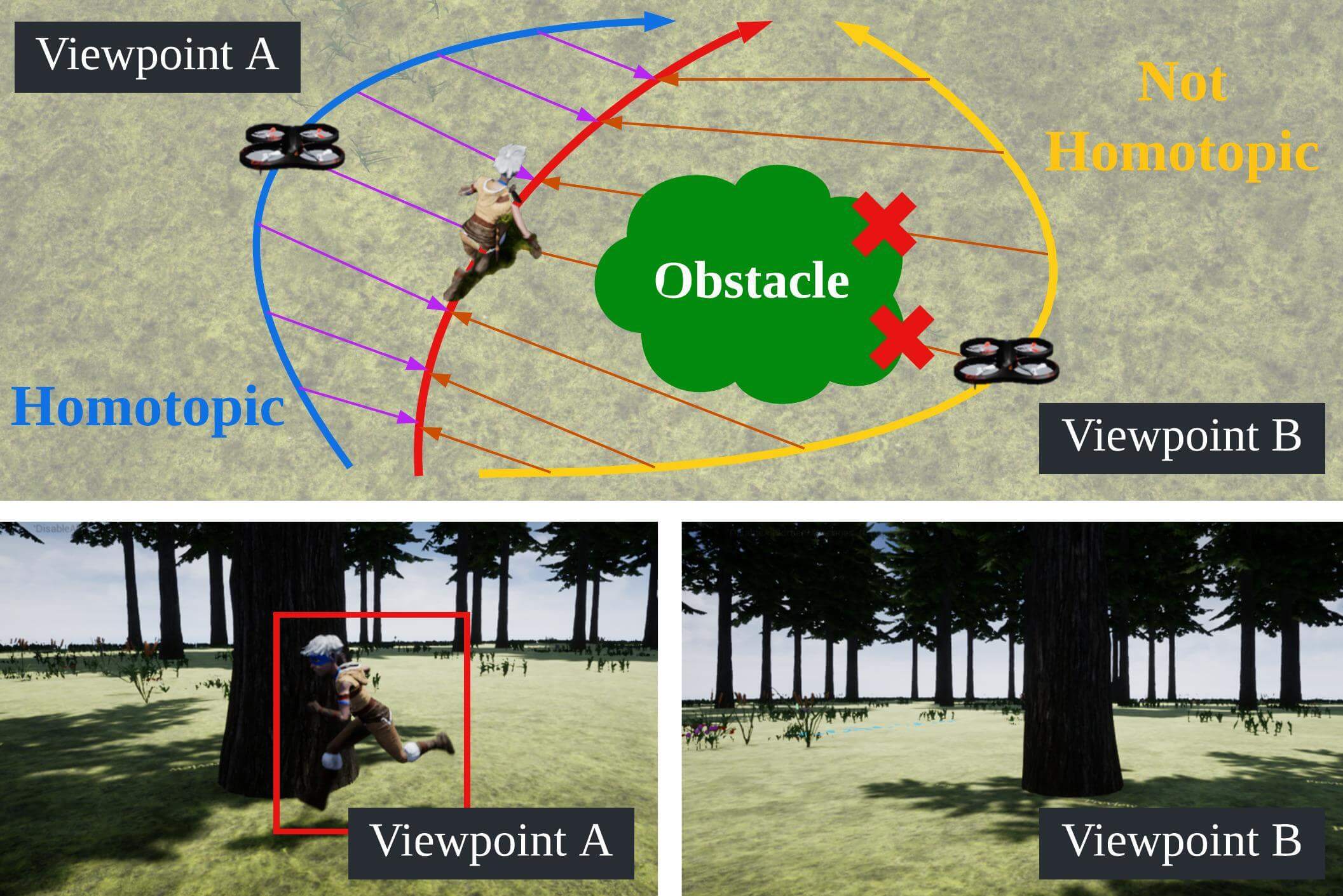}
\caption{Comparison between two viewpoints. The target can be observed at Viewpoint A (on a blue path), and the target is occluded by an obstacle at Viewpoint B (on a yellow path). Each arrow (purple, orange) represents the \textit{Line-of-Sight} between the target and the drone. A red cross means target occlusion by an obstacle.}
\label{fig:motivation}
\end{figure}
\section{Preliminary}
\label{sec:preliminary}
This section presents the relationship between target occlusion and path homotopy. 
In an obstacle environment, there exist multiple path homotopy classes \cite{homotopy}. As shown in Fig. \ref{fig:motivation}, the visibility of the target is closely related to path homotopy. We analyze the relation between them and take it into consideration when generating the chasing trajectory.

As stated in \cite{topology_book}, the definition of path homotopy is as follows.
\begin{definition}
\label{def:path-homotopy}
Paths $\sigma_{1}, \sigma_{2}: I \to X \subset \mathbb{R}^3$ are path-homotopic if there is a continuous map $F: I \times I \to X$ such that
\begin{equation}
\label{eq:path-homotopy}
    F(s,0)=\sigma_{1}(s)\ \text{and} \ F(s,1)=\sigma_{2}(s), 
\end{equation}
where $I$ is unit interval $[0,1]$.
\end{definition}
In this paper, \textit{Line-of-Sight} is of interest, which is defined as follows.
\begin{definition}
\label{def:line-of-sight}
\textit{Line-of-Sight} is a segment connecting two objects $\textbf{x}_1,\textbf{x}_2: [0,\infty) \to \mathbb{R}^3$.
\begin{equation}
\label{eq:line-of-sight}
    \mathcal{L}(\textbf{x}_1(t),\textbf{x}_2(t)) = (1-\epsilon)\textbf{x}_1(t) + \epsilon \textbf{x}_2(t), \quad \forall \epsilon\in I 
\end{equation}
where $t$ represents time.
\end{definition}
Based on the definitions above, we derive a relation between the path homotopy and visibility between two objects. 
\begin{theorem}
\label{theorem:vis_homotopy}
When two objects are reciprocally visible, the paths of objects are path-homotopic.
\end{theorem}
\begin{proof}
Suppose that two objects $\textbf{x}_1(t), \textbf{x}_2(t)$ move along paths $\sigma_1, \sigma_2$ in free space $\mathcal{F}\subset\mathbb{R}^3$ during time interval $[t_0,t_f]$, respectively.
\textit{i.e.} $\sigma_1 = \textbf{x}_1 \circ \xi$, $\sigma_2 = \textbf{x}_2 \circ \xi$, where $\xi: I \to [t_0,t_f]$ is the time mapping function such that $t=\xi(s)$. If the visibility between $\textbf{x}_1$ and $\textbf{x}_2$ is maintained, the \textit{Line-of-Sight} does not collide with obstacles: $\mathcal{L}(\textbf{x}_1(t), \textbf{x}_2(t))\subseteq \mathcal{F}$ for $\forall t\in [t_0,t_f]$. Then, by definition, $\mathcal{L}(\sigma_1(s),\sigma_2(s))=(1-\epsilon)\sigma_1(s)+\epsilon\sigma_2(s)\in \mathcal{F}$ for $\forall \epsilon,s\in I$. Since such a condition satisfies the definition of continuous mapping in Definition \ref{def:path-homotopy}, the two paths are homotopic.
\end{proof}
From Theorem \ref{theorem:vis_homotopy}, when the drone chooses a path with a different homotopy class from the target path, target occlusion inevitably occurs. Therefore, we explicitly consider path-homotopy when planning a chasing trajectory. 

Throughout this article, we use the notation in Table \ref{tab:nomenclature}. The bold small letters represent vectors, calligraphic capital letters denote sets, and italic lowercase letters mean scalar values.

\section{Overview}
\label{sec:overview}
 \begin{table}[t]
 \vspace{2mm}
    \caption{Nomenclature}
    \centering
    \label{tab:nomenclature}
    \begin{tabular}{|c|c|}
    \hline
    \textbf{Symbol} & \textbf{Definition}\\
    \hline
       $\textbf{p}_{c}(t)$  & A trajectory of the drone. \\
       \hline
       $\underbar{\textbf{c}}$  & \makecell[c]{An optimization variable that consists of \\Bernstein coefficients representing $\textbf{p}_{c}(t)$.} \\
       \hline
       $n_{c}$ & The degree of a polynomial trajectory $\textbf{p}_{c}(t)$. \\
       \hline
       $T$  & Planning horizon\\
       \hline
       $\theta_{f}$  & FOV of the camera built on the drone. \\
       \hline
       $v_{\text{max}},a_{\text{max}}$  &Maximum drone speed and acceleration \\
       \hline
       $\mathcal{F}$ & Free space\\
       \hline
       $\boldsymbol{\mathcal{O}}, \mathcal{O}_{j}$ & A set of obstacles. An $j$-th obstacle in $\mathcal{O}$\\
       \hline
       $N_{o}, N_{samp}$ & \makecell[c]{The number of elements of $\mathcal{O}$\\ and samples of end-points in the prediction.}\\
       \hline
       $\mathcal{R}_{\textbf{q}_{j}}(t),\hat{\textbf{q}}_{j}(t),r_{q_{j}}(t)$& \makecell[c]{A reachable area of an $j$-th target. A center\\trajectory and radius of $\mathcal{R}_{\textbf{q}_{j}}(t)$. We omit\\a subscript $j$ to handle an arbitrary target.}\\
       \hline
       $\mathcal{R}_{\textbf{o}_{j}}(t),\hat{\textbf{o}}_{j}(t),r_{o_{j}}(t)$ & \makecell[c]{A reachable area of an $j$-th obstacle. A center\\trajectory and radius of $\mathcal{R}_{\textbf{o}_{j}}(t)$. We omit\\a subscript $j$ to handle an arbitrary obstacle.}\\
       \hline
       $\mathcal{S}_{p}, \mathcal{P}_{p},\mathcal{P}_{s}$ & \makecell[c]{\textit{End Point Set}, a set of candidate \\trajectories and a set of non-colliding \\candidate trajectories.}\\
       \hline
       $\mathcal{L}(\textbf{x}_{1},\textbf{x}_{2})$ & Line-of-Sight between $\textbf{x}_{1}$ and $\textbf{x}_{2}$.\\
       \hline
       $\mathcal{V}_{O}(t), \mathcal{V}_{F}(t)$ & \makecell[c]{Target visible region against  a single\\ obstacle (TVR-O) and target visible region \\that can see the two targets simultaneously \\with camera FOV (TVR-F).}\\
       \hline
       $\mathcal{D}_{z}(t)$ & \makecell[c]{A region where the drone cannot see\\ both targets with the limited camera FOV.\\}\\
       \hline
        \makecell[c]{$\psi(\textbf{p}_{c}(t);\hat{\textbf{q}}(t),\mathcal{O}_{i})$, \\ $\psi(\textbf{p}_{c}(t);\hat{\textbf{q}}(t),\mathcal{O})$} & \makecell[c]{
        Visibility score against the $i$-th obstacle.\\
        Visibility score against all obstacles.
        }\\
       \hline
       ${}_{\textbf{x}_{2}}\textbf{x}_{1}(t)$  & \makecell[c]{Relative position between $\textbf{x}_{1}$ and $\textbf{x}_{2}$ at time $t$.\\$\textbf{x}_{1}(t)-\textbf{x}_{2}(t)$.}\\
       \hline
       $\|\textbf{x}\|_{p}$ & $L_{p}$ norm of $\textbf{x}$\\
       \hline
       $\textbf{x}'(t), \textbf{x}''(t), \textbf{x}'''(t)$ & First, second, third time derivatives of $\textbf{x}(t)$\\
       \hline
       
       $\textbf{a}^{\top}, {B}^{\top}$ & Transposed vector $\textbf{a}$ and matrix $B$ \\
       \hline
       $<, >, \leq, \geq$ & elementwise inequality symbols \\
       \hline
       $\textbf{x}_{(x)}, \textbf{x}_{(y)}$ & $x$- and $y$-components of $\textbf{x}$.\\
       \hline
       $\binom{n}{k}$ & \makecell[c]{Binomial coefficients, the number of\\$k$-combinations from a set of $n$ elements.} \\
       \hline
       $\textbf{0}_{m\times n}$ & An $m\times n$ matrix with all-zero elements.\\
       \hline
       $I_{m\times m}$ & An identity matrix with rank $m$.\\
       \hline
        $\text{tr}(\cdot), \text{det}(\cdot)$ & Trace and determinant of a matrix.\\
        \hline
        $\partial \mathcal{A}$ & A boundary of a closed set $\mathcal{A}$. \\
        \hline
        $\mathcal{B}(\textbf{x},r)$ & A ball with center at $\textbf{x}$ and radius $r$.  \\
        \hline
        $[A;B]$ & Row-wise concatenation of matrices $A$ and $B$\\
        \hline
        $|\mathcal{X}|$ & The number of elements in set $\mathcal{X}$ \\
        \hline
        \makecell[c]{$\overline{AB}$,\\ $\angle{ABC}$} & \makecell[c]{Segment connecting points $A$ and $B$. \\ Angle between $\overline{BA}$ and $\overline{BC}$.}\\
        \hline
    \end{tabular}
    \vspace{-2mm}
\end{table}
\subsection{Problem Setup}
\label{subsec:problem_setup}
\begin{figure}[t!]
\centering
\includegraphics[width = 1.0\linewidth]{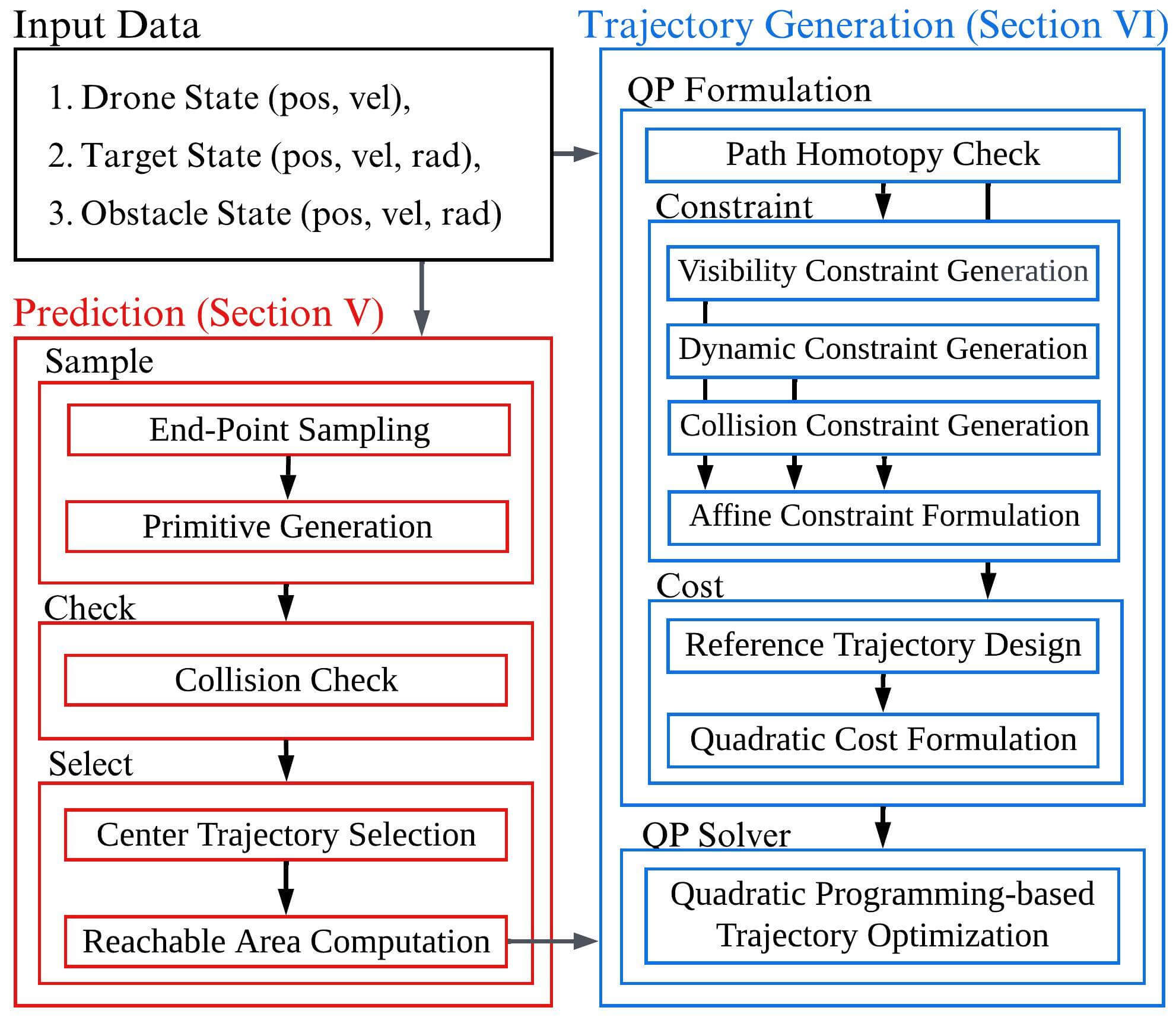}
\caption{Pipeline of the proposed chasing system.}
\label{fig4:Architecture}
\end{figure}

In this section, we formulate the trajectory planning problem for a tracking drone with firmly attached camera sensors that have limited FOVs $\theta_{f}\in (0,\pi)$ [rad]. We consider an environment $\mathcal{W}$, which consists of separate cylindrical static and dynamic obstacles. Our goal is to generate a trajectory of the drone so that it can see the single and dual targets ceaselessly in $\mathcal{W}$ over the time horizon $[0, T]$.
To achieve the goal, the drone has to predict the future movement of moving objects, such as targets and dynamic obstacles, and generate a target-chasing trajectory. 
To reflect the prediction error, we compute the reachable areas of moving objects. 
Then, based on the reachable areas, the planner generates a continuous-time trajectory that satisfies dynamical feasibility and avoids collision while preserving the target visibility. The whole pipeline is summarized in Fig. \ref{fig4:Architecture}. Within this pipeline, we set up two problems: 1) \textit{Prediction} and 2) \textit{Chasing problem}.
\subsubsection{Prediction problem}
\label{subsubsec:prediction_problem}
The prediction module forecasts reachable areas of moving objects such as targets and dynamic obstacles, over a time horizon $t\in[0, T]$. The reachable areas $\mathcal{R}_{\textbf{q}}(t)$ and $\mathcal{R}_{\textbf{o}}(t)$ are sets that encompass the future positions of the targets $\textbf{q}(t)$ and obstacles $\textbf{o}(t)$, respectively. The goal is to calculate the reachable areas considering an obstacle set $\mathcal{O}$. 
Specifically, we represent the $j$-th target ($j=1,2$) and $k$-th obstacle ($k\in\{1,\ldots,N_{o}:=|\mathcal{O}|\}$) as $\textbf{q}_j$ and $\textbf{o}_k$ respectively.
Since the methods to compute $\mathcal{R}_{\textbf{o}}(t)$ and $\mathcal{R}_{\textbf{q}}(t)$ are equivalent, we use a symbol $\textbf{p}$ instead of $\textbf{q}$ and $\textbf{o}$ to represent information of reachable areas of moving objects, in Section \ref{sec:object_trajectory_prediction}.
\subsubsection{Chasing problem}
\label{subsubsec:chasing_problem}
Given the reachable areas of the target and the obstacles obtained by the prediction module, we generate a trajectory of the drone $\textbf{p}_{c}(t)$ that accounts for the following factors.
\begin{itemize}[leftmargin=1em]
    \item Occlusion avoidance: To robustly prevent target occlusion by obstacles, we make the \textit{Lines-of-Sight} between the drone and all points within the target’s reachable area $\mathcal{R}_{q}(t)$ do not intersect with the obstacle’s reachable area $\mathcal{R}_{o}(t)$.
    \begin{equation}
    \label{eq4:visibility_constraint}
     \mathcal{L}\big(\textbf{p}_{c}(t),\mathcal{R}_{\textbf{q}}(t)\big) \cap \mathcal{R}_{\textbf{o}}(t)=\emptyset,\ \forall t\in[0,T]   
    \end{equation}
    \item Field-of-view: While tracking two targets, the drone should keep both targets within the limited FOV $\theta_{f}$ of its camera. The angle made by the two \textit{Lines-of-Sight}, $\mathcal{L}(\textbf{p}_{c}(t),\textbf{q}_{1}(t))$ and $\mathcal{L}(\textbf{p}_{c}(t),\textbf{q}_{2}(t))$ should not exceed $\theta_{f}$.
    \begin{equation}
    \label{eq4:fov_constraint}
         \cos ^{-1}\bigg(\frac{{}_{\textbf{p}_{c}}\textbf{q}_{1}(t)^{\top}{}_{\textbf{p}_{c}}\textbf{q}_{2}(t)}{\|{}_{\textbf{p}_{c}}\textbf{q}_{1}(t)\|_{2}\|{}_{\textbf{p}_{c}}\textbf{q}_{2}(t)\|_{2}}\bigg)\leq \theta_{f},\ \forall t\in[0,T] 
    \end{equation}
    \item Collision avoidance: To ensure safety, the drone should not collide with both targets and obstacles.
    \begin{equation}
    \label{eq4:collision_constraint}
        \mathcal{B}(\textbf{p}_{c}(t),r_{c}) \cap \big\{\mathcal{R}_{\textbf{o}}(t)\cup \mathcal{R}_\textbf{q}(t)\big\}=\emptyset,\ \forall t\in[0,T]
    \end{equation}
     $r_c$ is the radius of the drone, and $\mathcal{B}(\cdot,\cdot)$ is the operator representing a ball with a center and radius.
    \item Dynamic feasibility: Considering the drone’s actuator limits, the planned trajectory should not exceed the maximum velocity $v_{\max}$ and acceleration $a_{\max}$.
    \begin{equation}
    \label{eq4:dynamic_constraint}
        \|\textbf{p}_{c}'(t)\|_{2} \leq v_{\text{max}}, \ \| \textbf{p}_{c}''(t)\|_{2} \leq a_{\text{max}},\ \forall t\in[0,T]
    \end{equation}
\end{itemize}
\subsubsection{Assumptions}
\label{subsusbsec:assumption}
For the \textit{Prediction problem}, we assume that (\textit{AP1}) the moving objects do not collide with obstacles, and (\textit{AP2}) they do not move in a jerky way. In the \textit{Chasing problem}, we assume that (\textit{AC1}) the maximum velocity $v_{\text{max}}$ and the maximum acceleration $a_{\text{max}}$ of the drone are higher than the target and obstacles. 
Furthermore, based on Theorem \ref{theorem:vis_homotopy}, when the targets move along a path with different homotopy classes, occlusion unavoidably occurs, so we assume that (\textit{AC2}) all targets move along homotopic paths against obstacles. \\
In addition, we set the flying height of the drone to a fixed level for the acquisition of consistent images of the target. From the problem settings, we focus on the design of chasing trajectory in the $x-y$ plane. 

\subsubsection{Trajectory representation}
\label{subsubsec:trajectory_representation}
Due to the virtue of differential flatness of quadrotor dynamics \cite{snap}, the trajectory of multi-rotors can be expressed with a polynomial function of time $t$. In this paper, Bernstein basis is employed to express polynomials. Bernstein bases of $n$-th order polynomial for time interval $[t_a, t_b]$ are defined as follows.
\begin{equation}
\label{eq:bernstein_basis}
    b_{k,n}(t;t_{a},t_{b}) = \binom{n}{k}\frac{(t_b-t)^{n-k}(t-t_a)^{k}}{(t_b-t_a)^n},\ 0\leq k\leq n
\end{equation}
Since the bases defined above are non-negative in the time interval $[t_{a},t_{b}]$, a linear combination with non-negative coefficients makes the total value become non-negative. We utilize this property in the following sections.

The trajectory of the drone, $\textbf{p}_{c}(t)\in \mathbb{R}^2$, is represented as an $M$-segment piecewise Bernstein polynomial.
\begin{equation}
\label{eq:segment_representation}
  \textbf{p}_{c}(t)=\begin{cases}
    &\textbf{C}_{1}^{\top} \textbf{b}_{n_{c},1}(t)\ \ \ \ \ t\in [T_{0},T_{1}]\\
    &\textbf{C}_{2}^{\top} \textbf{b}_{n_{c},2}(t)\ \ \ \ \ t\in [T_{1},T_{2}]\\
    &\ldots \\
    &\textbf{C}_{M}^{\top} \textbf{b}_{n_{c},M}(t)\ \ \ t\in [T_{M-1},T_{M}]\\
  \end{cases}
\end{equation}
$\textbf{b}_{n_{c},m}(t)\in\mathbb{R}^{(n_{c}+1)}$ is a vector that stacks the Bernstein basis functions $b_{k,n_{c}}(t;T_{m-1},T_{m})$ vertically for $k=0$ to $n$, and $\textbf{C}_{m}=[\textbf{c}_{m(x)}, \textbf{c}_{m(y)}]\in \mathbb{R}^{(n_{c}+1)\times2}$ is a coefficient matrix with $n_{c}+1$ control points stacked row-wise, where each column represents the $x$ and $y$ coordinates of the trajectory: $\textbf{p}_{c}(t)=[\textbf{c}_{m(x)}^{\top}\textbf{b}_{n_{c},m}(t),\textbf{c}_{m(y)}^{\top}\textbf{b}_{n_{c},m}(t)]^{\top}$ for $t\in[T_{m-1},T_{m}]$.
We define $\textbf{c}_{m}=[\textbf{c}_{m(x)}^{\top},\textbf{c}_{m(y)}^{\top}]^{\top}\in \mathbb{R}^{2(n_c+1)}$, and collect them into  $\underbar{\textbf{c}} = [\textbf{c}_{1}^{\top},\ldots,\textbf{c}_{M}^{\top}]^{\top}\in \mathbb{R}^{2M(n_c+1)}$, which serve as the decision variable of the polynomial trajectory optimization.
\subsubsection{Objectives}
\label{objective_qp}
For the \textit{Prediction Problem}, the prediction module forecasts moving objects' reachable area $\mathcal{R}_{\textbf{p}}(t)$.  
Then, for the \textit{Chasing Problem}, the chasing module formulates a QP problem with respect to $\underbar{\textbf{c}}$ and finds an optimal $\underbar{\textbf{c}}$ so that the drone chases the target without occlusion and collision while satisfying dynamical limits.
\section{Reachable Area Prediction}
\label{sec:object_trajectory_prediction}
This section introduces a method for calculating reachable areas using the sample-check strategy. We sample a set of motion primitives and filter out the primitives that collide with obstacles. Then, we calculate a reachable area that encloses the remaining collision-free primitives. Figs. \ref{subfig5:PredictionSampling}-\ref{subfig5:ReachSet100} visualize the prediction process.
\subsection{Candidate for Future Trajectory of Moving Object}
\label{subsec:candidate_object_trajectory}
We sample the motion primitives of the target by collecting the positions that can be reached at $t=T$ and interpolating them with the current position.
\subsubsection{Endpoint sampling}
\label{subsubsec:end_point_sampling}
The positions of a moving object at time $t=T$ are sampled, using a constant velocity model under disturbance:
\begin{equation}
\label{eq:dynamics_object}
\begin{aligned}
&
\begin{bmatrix}
    \textbf{p}'(t) \\ \textbf{p}''(t)
\end{bmatrix}
= F_p \begin{bmatrix}
    \textbf{p}(t) \\ \textbf{p}'(t)
\end{bmatrix}
+ G_p \textbf{w}, \quad \textbf{w}\sim (0,Q), \\
& F_p =
\begin{bmatrix}
    \textbf{0}_{2\times2} & \textbf{I}_{2\times2}\\
    \textbf{0}_{2\times2} & \textbf{0}_{2\times2}
\end{bmatrix}
, \quad G_p = 
\begin{bmatrix}
    \textbf{0}_{2\times2} \\
    \textbf{I}_{2\times2}
\end{bmatrix}
\end{aligned}        
\end{equation}
where $\textbf{p}(t)\in \mathbb{R}^{2}$ is position of the moving object, and $\textbf{w} \in \mathbb{R}^2$ is white noise with power spectral density $Q$. Then, with no measurement update, estimation error covariance $P(t)$ in continuous-time Kalman filter propagates with time \cite{error_propagation}.
\begin{equation}
\label{eq:error_propagation}
P'(t) = F_p(t)P(t)+P(t)F_{p}^{\top}(t)+G_{p}QG_{p}^{\top}
\end{equation}
We collect $N_{samp}$ points from a 2-dimensional Gaussian distribution $\mathcal{N}(\hat{\textbf{p}}_{0}+\hat{\textbf{p}}_{0}'T,P(T))$
where $\hat{\textbf{p}}_{0}$ and $\hat{\textbf{p}}_{0}'$ are position and velocity at current time $t=0$, respectively. It is illustrated in Fig. \ref{subfig5:PredictionSampling}, where the samples are shown as red points, and we refer to the set of the samples the
\textit{Endpoint Set} $\mathcal{S}_{p}=\{\textbf{s}_{p,i}\in\mathbb{R}^2 | i=1,\ldots,N_{samp} \}.$
\subsubsection{Primitive generation}
\label{subsubsec:primitive_generation}
\begin{figure}[t!]
\centering
\begin{subfigure}[t]{0.23\textwidth}
\centering\includegraphics[width=\textwidth]{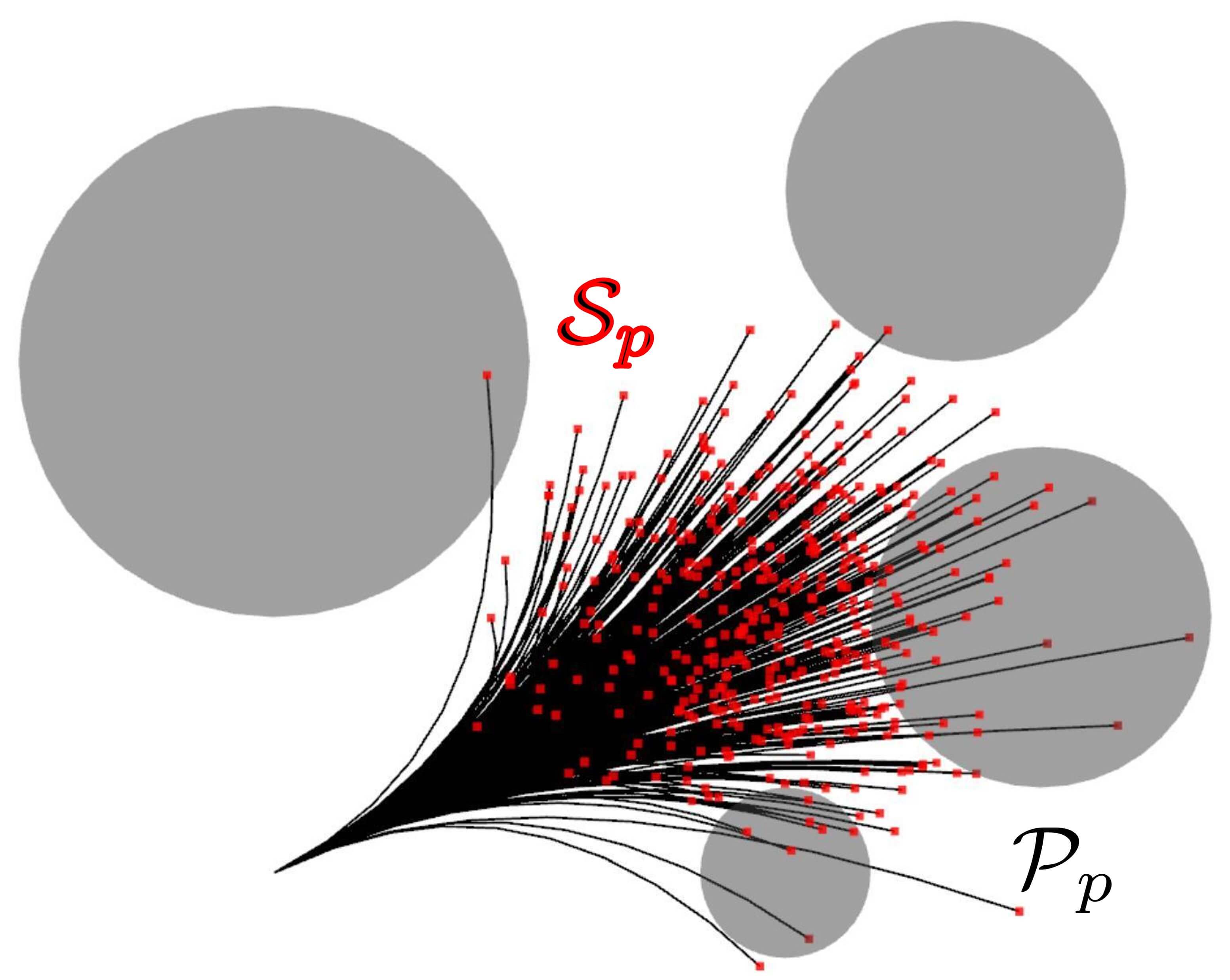}
\caption{Primitive Sampling}
\label{subfig5:PredictionSampling}
\end{subfigure}
\hfill
\begin{subfigure}[t]{0.23\textwidth}
\centering\includegraphics[width=\textwidth]{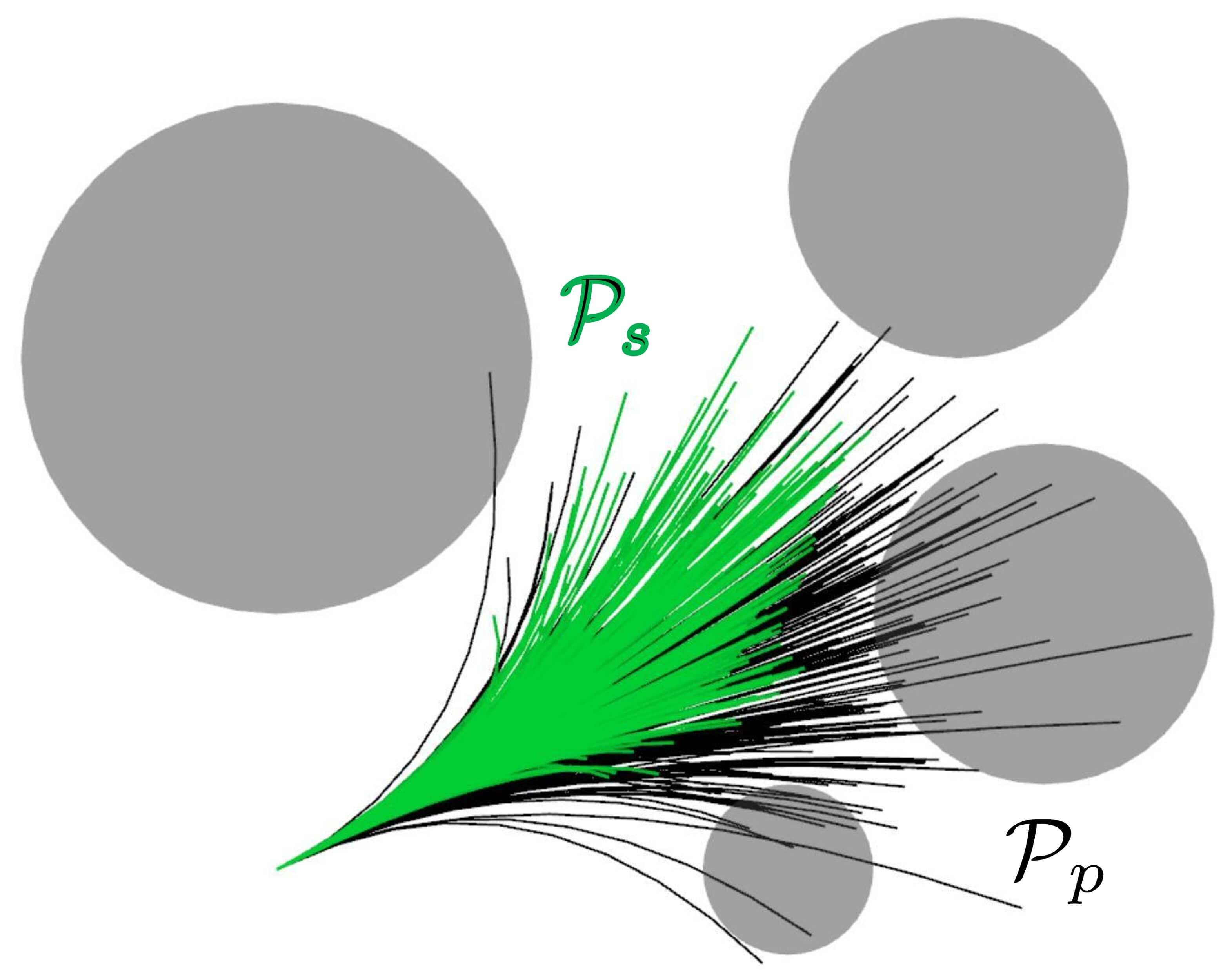}
\caption{Collision Check}
\label{subfig5:collision_check}
\end{subfigure}
\hfill
\begin{subfigure}[b]{0.23\textwidth}
\centering\includegraphics[width=\textwidth]{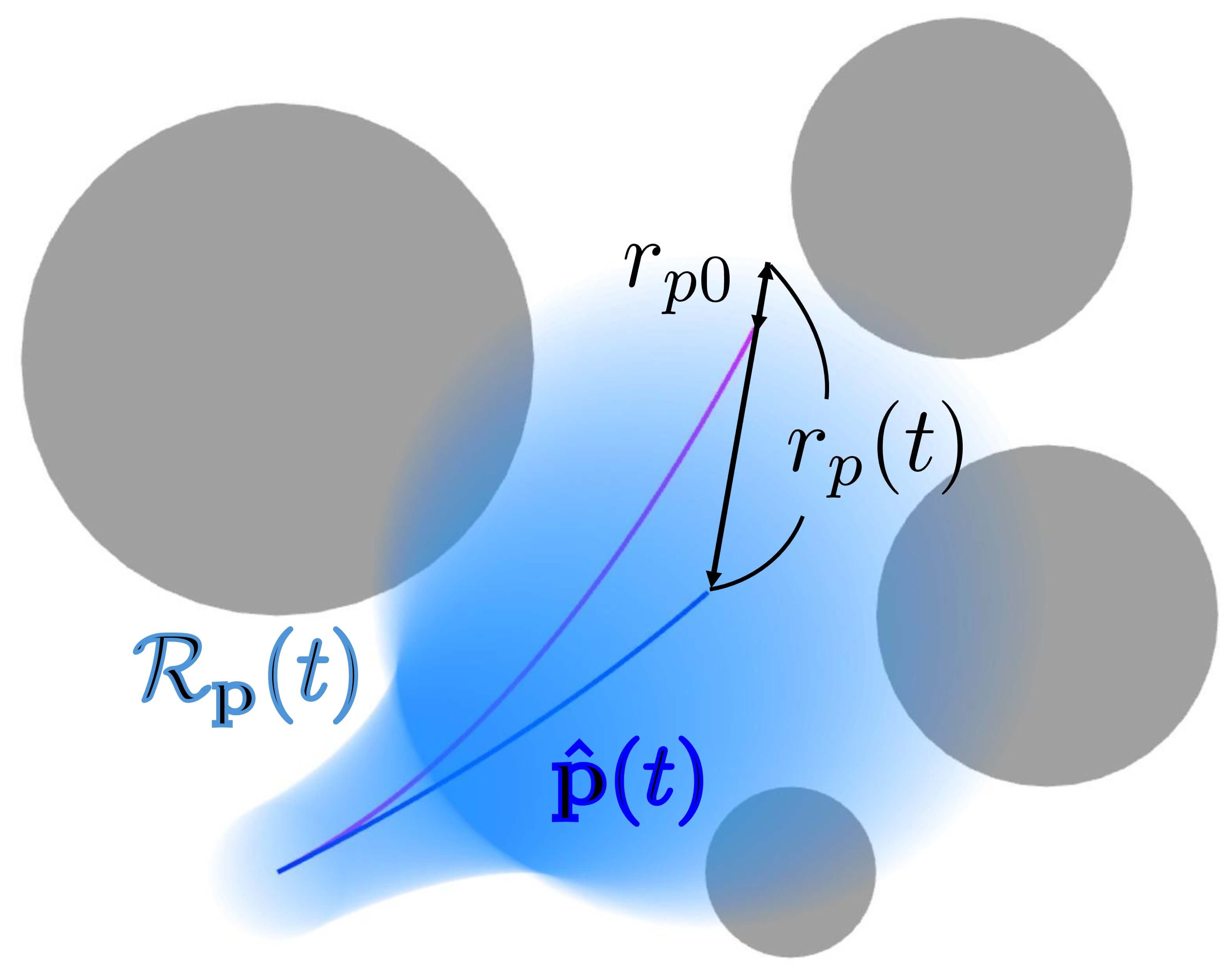}
\caption{Reachable Area}
\label{subfig5:ReachSet100}
\end{subfigure}
\hfill
\begin{subfigure}[b]{0.23\textwidth}
\centering\includegraphics[width=\textwidth]{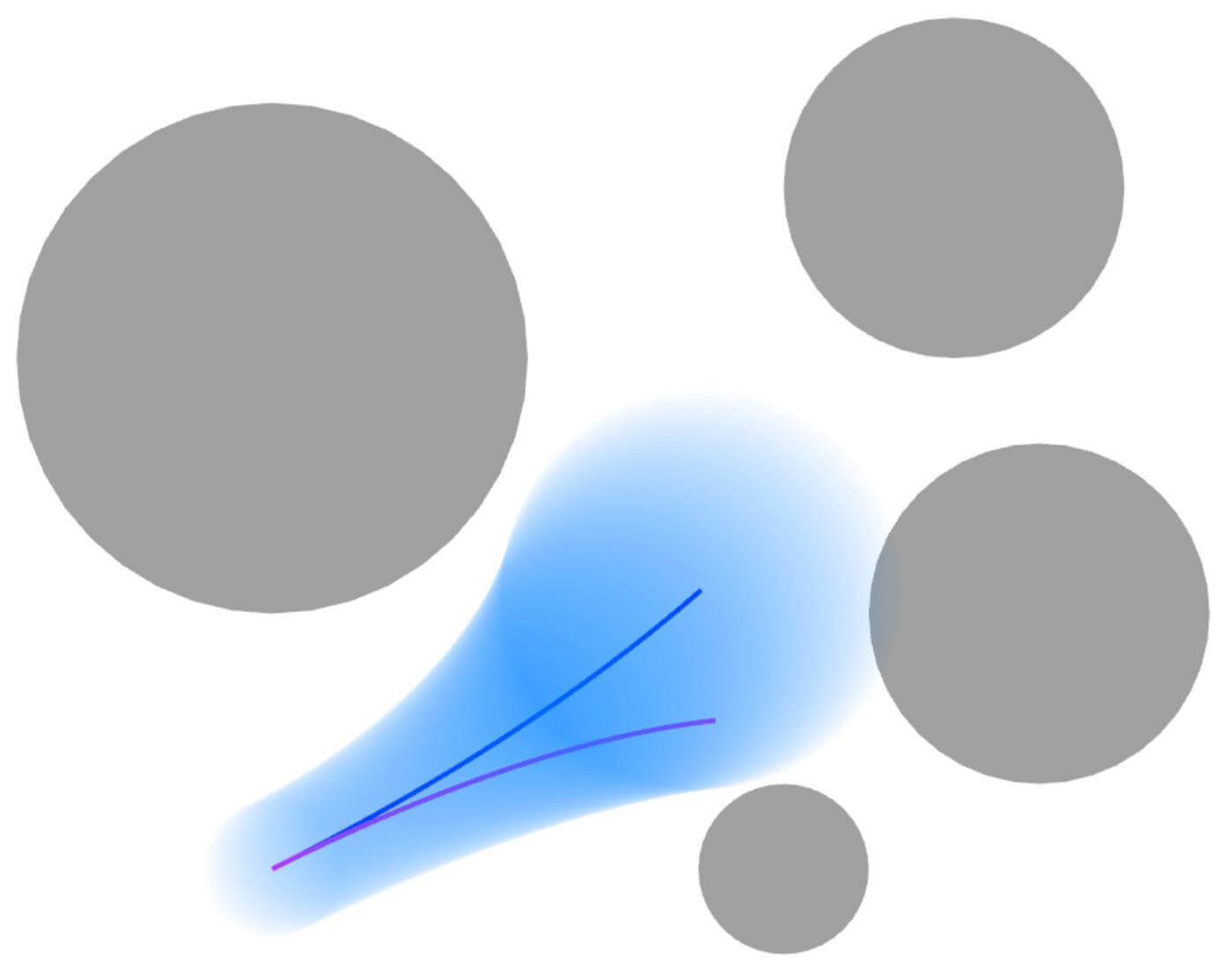}
\caption{Reachable Area, $\epsilon_{p}=0.3$}
\label{subfig5:ReachSet80}
\end{subfigure}
\caption{\textbf{(a)}: Sampled endpoints $\mathcal{S}_{p}$ (red) and primitives $\mathcal{P}_{p}$ (black) among obstacles (grey). \textbf{(b)}: Primitives $\mathcal{P}_{p}$ (black) and non-colliding primitives $\mathcal{P}_{s}$ (green). 
\textbf{(c)}: The best primitive $\hat{\textbf{p}}(t)$ (blue), reachable area $\mathcal{R}_{\textbf{p}}(t)$ (light blue), and the farthest primitive (magenta). \textbf{(d)}: The prediction with the outlier factor $\epsilon_{p} = 0.3$.}
\vspace{-4mm}
\end{figure}

Given the initial position, velocity, and end positions $\textbf{s}_{p,i}\in \mathcal{S}_{p}$, we design trajectory candidates $\hat{\textbf{p}}_{i}(t)$, $i=1,\ldots,N_{samp}$ for moving objects. Under the assumption (\textit{AP2}) that moving objects do not move in a jerky way, we establish the following problem.
\begin{equation}
\label{eq:primitive_optimization_problem}
    \begin{aligned}
        &\underset{\hat{\textbf{p}}_{i}(t)}{\text{min}} &&\int_{0}^{T} \|\hat{\textbf{p}}_{i}'''(t)\|^{2}_{2}dt \\
        & \text{s.t.} && \hat{\textbf{p}}_{i}(0) = \hat{\textbf{p}}_{0},\ 
        \hat{\textbf{p}}_{i}'(0) = \hat{\textbf{p}}_{0}',\ 
        \hat{\textbf{p}}_{i}(T) = \textbf{s}_{p,i}
    \end{aligned}
\end{equation}
Recalling that the trajectory is represented with Bernstein polynomial, we write an $i$-th candidate trajectory as $\hat{\textbf{p}}_{i}(t)=\textbf{P}^{\top}_{i}\textbf{b}_{n_{p}}(t)$, $i=1,\ldots, N_{samp}$ where $\textbf{P}_{i}= [\textbf{p}_{i(x)},\textbf{p}_{i(y)}]\in \mathbb{R}^{(n_{p}+1) \times 2}$ and $\textbf{b}_{n_{p}}(t)=[b_{0,n_{p}}(t;0,T),\ldots,b_{n_{p},n_{p}}(t;0,T)]^{\top}$. By defining $\underbar{\textbf{p}}_{i}:=[\textbf{p}_{i(x)}^{\top},\textbf{p}_{i(y)}^{\top}]^{\top}$ as an optimization variable, (\ref{eq:primitive_optimization_problem}) becomes a QP problem
\begin{equation}
    \label{eq:primitive_qp}
    \begin{aligned}
    &\underset{\underbar{\textbf{p}}_{i}}{\text{min}} && \frac{1}{2}\underbar{\textbf{p}}_{i}^{\top}Q_{p}\underbar{\textbf{p}}_{i}\\
    &\text{s.t.} && A_{p}\underbar{\textbf{p}}_{i}=\textbf{b}_{p,i}\\
    \end{aligned}
\end{equation}
where $Q_{p}$ is a positive semi-definite matrix, $A_{p}$ is a $6\times(n_{p}+1)$ matrix, and $\textbf{b}_{p,i}$ is a vector composed of $\hat{\textbf{p}}_{0}$, $\hat{\textbf{p}}_{0}'$, and $\textbf{s}_{p,i}$.
(\ref{eq:primitive_qp}) can be converted into unconstrained QP, whose optimal $\textbf{p}_{i}$ is a closed-form solution as follows.
\begin{equation}
\label{eq:KKT_matrix}
    \begin{bmatrix}
        \underbar{\textbf{p}}_{i}\\
        \boldsymbol{\lambda}
    \end{bmatrix}
    =
    \begin{bmatrix}
        Q_p & A_{p}^{\top}\\
        A_{p} & \textbf{0}_{6\times6}
    \end{bmatrix}^{-1}
    \begin{bmatrix}
    \textbf{0}_{2(n_p+1)\times1} \\
    \textbf{b}_{p,i}
    \end{bmatrix}
\end{equation}
where $\boldsymbol{\lambda}$ is a lagrange multiplier. A set of candidate trajectories of the moving object is defined as $\mathcal{P}_{p}$, and Fig. \ref{subfig5:PredictionSampling} shows an example, where the trajectories are shown as black splines.
\subsection{Collision Check}
\label{subsec:colli_check}
Under the assumption (\textit{AP1}) that moving objects do not collide, trajectory candidates $\hat{\textbf{p}}_{i}(t)\in \mathcal{P}_{p}$ that violate the below condition are filtered out:
\begin{equation}
    \label{eq:collision_constraint_pred}
    \|\hat{\textbf{p}}_{i}(t)-\hat{\textbf{o}}_{j}(t) \|_{2}^{2}-(r_{p0}+r_{o_{j}}(t))^{2} \geq 0,\ \forall t \in [0,T]
\end{equation}
where $\hat{\textbf{o}}_{j}(t)$ and $r_{o_{j}}(t)$ denote the predicted trajectory and radius, including the prediction error bounds, of the j-th obstacle, respectively, while $r_{p0}$ represents the radius of the dynamic object of interest.
Due to the fact that all terms in (\ref{eq:collision_constraint_pred}) can be represented in polynomials, and Bernstein bases are non-negative in the time period $[0, T]$, entirely non-negative coefficients make the left-hand side of (\ref{eq:collision_constraint_pred}) non-negative. 
We examine all coefficients of (\ref{eq:collision_constraint_pred}) for the primitives belonging to $\mathcal{P}_{p}$.
For the mathematical details, see Appendix \ref{appendix:collision_check}. We call a set of primitives that pass the test $\mathcal{P}_{s}$. Fig. \ref{subfig5:collision_check} shows an example,  where the green splines are collision-free out of all the black trajectories.
\subsection{Prediction with Error Bounds}
\label{subsec:define_reachable_sets}
With the set $\mathcal{P}_{s}$, the reachable area $\mathcal{R}_{\textbf{p}}(t)$ is defined as a time-varying ball enclosing all the primitives in $\mathcal{P}_{s}$.
\begin{equation}
\label{eq:reachable_set}
    \mathcal{R}_{\textbf{p}}(t) = \mathcal{B}(\hat{\textbf{p}}(t),r_{p}(t))
\end{equation}
We determine a center trajectory $\hat{\textbf{p}}(t)$ as the primitive with the smallest sum of distances to the other primitives in $\mathcal{P}_{s}$:
\begin{equation}
\label{eq:best_primitive}
    \hat{\textbf{p}}(t)= \underset{\hat{\textbf{p}}_{i}(t)\in\mathcal{P}_s}{\text{argmin}}
    \sum_{j\neq i, \hat{\textbf{p}}_{j}(t)\in\mathcal{P}_s}^{|\mathcal{P}_{p}|} \int_{0}^{T}\|\hat{\textbf{p}}_{i}(t)-\hat{\textbf{p}}_{j}(t) \|_{2}dt.
\end{equation}
\begin{proposition}
\label{prop:center_trajectory}
The optimization problem (\ref{eq:best_primitive}) is equivalent to the following problem.
\begin{equation}
    \label{eq:best_end_point}
    \begin{aligned}
    &\hat{\textbf{p}}(t) = \hat{\textbf{p}}_{i^{*}}(t),\ i^{*} =\underset{i}{\text{argmin}}\sum_{j=0}^{|\mathcal{P}_{p}|} \|\textbf{s}_{p,i}- \textbf{s}_{p,j}\|_2, \\
    &\text{where}\ \textbf{s}_{p,i}=\hat{\textbf{p}}_{i}(T),\  \textbf{s}_{p,j}=\hat{\textbf{p}}_{j}(T), \ \  \hat{\textbf{p}}_{i}(t),\hat{\textbf{p}}_{j}(t) \in \mathcal{P}_{s}
    \end{aligned}
\end{equation}
\end{proposition} 
\begin{proof}
    See Appendix \ref{appendix:prediction_proof}.
\end{proof}
From Proposition \ref{prop:center_trajectory}, $\hat{\textbf{p}}(t)$ is determined by simple arithmetic operations and the process has a time complexity of $O(|\mathcal{P}_{s}|^{2})$. Then, we define $r_p(t)$ so that $\mathcal{R}_{\textbf{p}}(t)$ encloses all the primitives in $\mathcal{P}_{s}$ for $\forall{t} \in [0,T]$.
\begin{equation}
    \label{eq:reachable_set_radius}
        r_{p}(t) = \underset{\hat{\textbf{p}}_{j}(t)\in \mathcal{P}_{s}}{\max} \|\hat{\textbf{p}}(t)-\hat{\textbf{p}}_{j}(t)\|_{2}+r_{p0}
\end{equation}
The second term in the right hand side of (\ref{eq:reachable_set_radius}) allows the whole body to be contained in $\mathcal{R}_{\textbf{p}}(t)$.
Fig. \ref{subfig5:ReachSet100} illustrates how $\mathcal{R}_{\textbf{p}}(t)$, shown as the blue area, is determined.
The use of $r_{q}(t)$ and $r_{o}(t)$ in \textit{chasing problem} allows for the consideration of the visibility of the entire bodies of the targets, as well as the potential movements of moving targets and dynamic obstacles.
\subsection{Evaluation}
\label{subsec:prediction_evaluation}
We measure the computation time of prediction in obstacle environments. We set $n_p$ as 3 and test 1000 times for different scenarios $(N_{samp}, N_{o})=$ (500,2), (500,4), (2000,2), (2000,4). The prediction module is computed on a laptop with an Intel i7 CPU, with a single thread implementation, and execution time is summarized in Table \ref{tab:prediction_time}. As $N_{samp}$ and $N_{o}$ increase, the time needed for calculating primitives and collision checks increases.
However, environments where obstacles are densely located around moving objects, as shown in Fig. \ref{subfig5:collision_check}, may result in a small $\mathcal{P}_{s}$.
Accordingly, the time to compute a reachable area decreases as the number of close obstacles increases. On the other hand, if the object moves while staying far from the obstacles, the computation time for the reachable area increases as the number of obstacles grows.

\begin{figure}[t!]
\centering
\begin{subfigure}[t]{0.24\textwidth}
\centering\includegraphics[width=\textwidth]{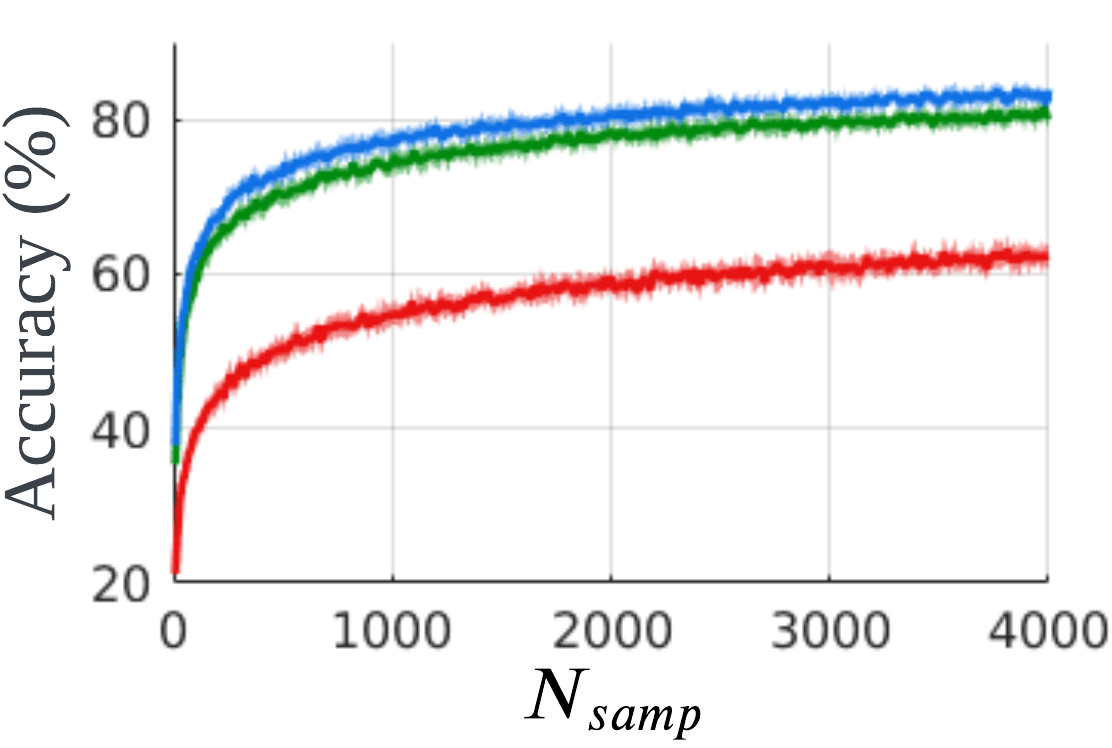}
\caption{$\forall t\in [0.1T,T]$}
\label{fig:prediction_accuracy_90}
\end{subfigure}
\hfill
\begin{subfigure}[t]{0.24\textwidth}
\centering\includegraphics[width=\textwidth]{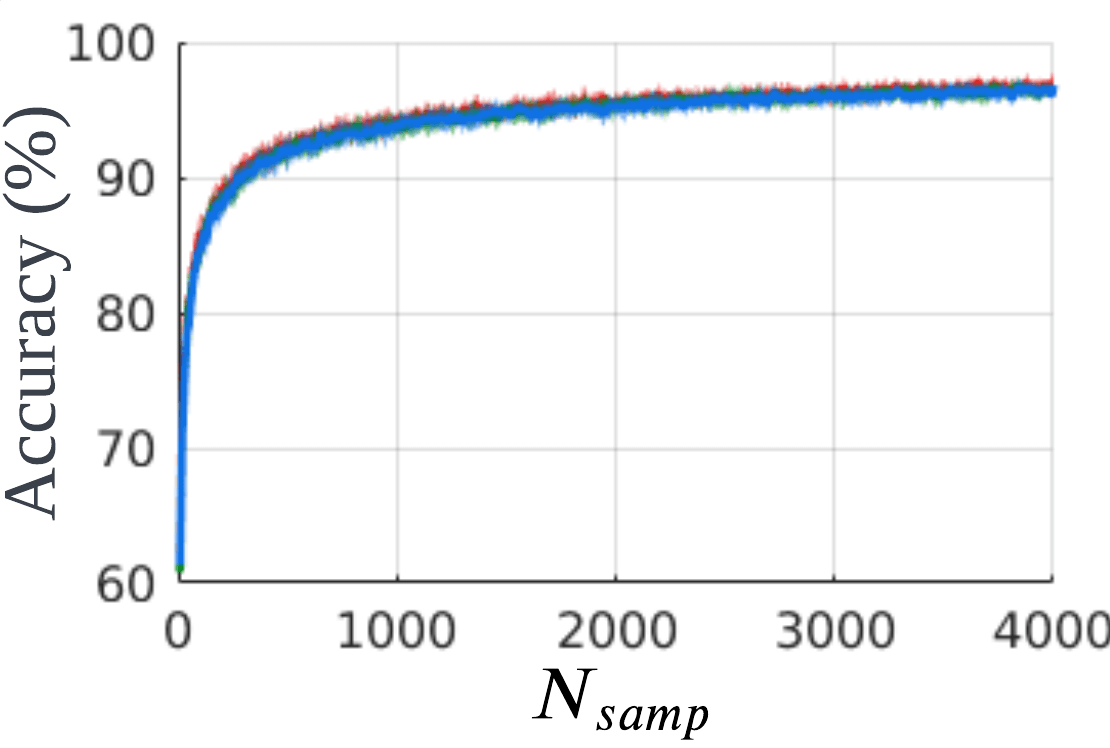}
\caption{$\forall t\in [0.5T,T]$}
\label{fig:Prediction_Accuracy_50}
\end{subfigure}
\caption{Prediction accuracy. Red, green, and blue lines represent $Q=$ 0.1, 0.5, 1.0 [$\text{m}^{2}/\text{s}^{3}$] cases, respectively.}
\label{fig:Prediction_Accuracy}
\end{figure}

Also, we test the accuracy of the presented prediction methods. For evaluation, discretized models of (\ref{eq:dynamics_object}) with power spectral density of white noise $Q=$ 0.1, 0.5, 1.0 $[\text{m}^{2}/\text{s}^{3}]$ are considered. We confirm whether $\mathcal{B}(\textbf{p}(t),r_{p0})\subset \mathcal{R}_{\textbf{p}}(t)$ is satisfied in obstacle-free situations while progressively increasing the $N_{samp}$. For each $N_{samp}$, the tests are executed 10000 times, and the acquired accuracy is shown in Fig. \ref{fig:Prediction_Accuracy}. Contrary to the assumption (\textit{AP2}) that a moving object follows a smooth trajectory, the discretized model used for evaluation can exhibit jerky movements. This indicates that our prediction approach is capable of handling non-ideal target motions when a sufficient number of samples $N_{samp}$ is used.
As $N_{samp}$ increases, the accuracy improves, as shown in Fig. \ref{fig:Prediction_Accuracy}, but the computation time also increases, as presented in Table \ref{tab:prediction_time}.
Therefore, $N_{samp}$ should be determined according to the computation resources for a balance between accuracy and running time. 
In this paper, we set $N_{samp}$ as 2000.
With this setting, $\mathcal{B}(\textbf{p}(t),r_{p0})\subset \mathcal{R}_{\textbf{p}}(t)$, $\forall t\in (0,T]$ was satisfied 98.8\% empirically in the simulations in Section \ref{subsec:simulations}.

As elaborated in Section \ref{subsec:problem_setup}, we specifically calculate the reachable areas of moving targets and obstacles: $\mathcal{R}_{\textbf{q}}(t)=\mathcal{B}(\hat{\textbf{q}}(t),r_{q}(t))$ and $\mathcal{R}_{\textbf{o}}(t)=\mathcal{B}(\hat{\textbf{o}}(t),r_{o}(t))$.
Subsequently, in the following section, we generate a trajectory that maximizes the visibility of $\mathcal{R}_{\textbf{q}}(t)$ while avoiding $\mathcal{R}_{\textbf{o}}(t)$.
\section{Chasing Trajectory Generation}
\label{sec:chasing_trajectory_generation}
This section formulates a QP problem to make a chasing trajectory occlusion-free, collision-free, and dynamically feasible. We first define the target visible region (TVR), the region with maximum visibility of the target's reachable area, and formulate constraints regarding collision avoidance and dynamic limits. Then, we design the reference trajectory to enhance the target visibility. Regarding the target visibility, all the steps are based on the path homotopy.
\begin{table}[t]
    \vspace{2mm}
    \centering
    \caption{Computation Time in Prediction Process}
    \label{tab:prediction_time}
    \begin{tabular}{c|cccc}
        \toprule
         $(N_{samp},N_{o})$ &(500,2) &(500,4) &(2000,2) &(2000,4)\\ \hline
         \begin{tabular}[c]{@{}c@{}}Endpoints Sampling \& \\ {Primitive Generation\ [}$\mu$s{]}\end{tabular}& 43.97 & 47.23 & 168.3 & 184.1 \\ \hline
         Collision Check\ [$\mu$s] & 146.3 & 185.3 & 487.9 & 750.7 \\ \hline
         \begin{tabular}[c]{@{}c@{}}Reachable Area\\ {Computation\ [}ms{]}\end{tabular}& 0.245 & 0.172 & 2.921 & 2.073 \\ \hline     
         \textbf{Total Time} [ms] & 0.435 & 0.405 & 3.577 & 3.008\\
        \bottomrule
    \end{tabular}
\end{table}
\subsection{Homotopy Class Check}
\label{subsec:topology_check}
In 2-dimensional space, there exist two classes of path homotopy against a single obstacle, as shown in Fig. \ref{fig:TopologyClass}. As stated in Theorem \ref{theorem:vis_homotopy}, the drone should move along a homotopic path with the target path to avoid occlusion.
Based on the relative position between the drone and the obstacle, ${}_{\hat{\textbf{o}}}\textbf{p}_{c}(t)=\textbf{p}_{c}(t)-\hat{\textbf{o}}(t)$, and the relative position between the target and the obstacle, ${}_{\hat{\textbf{o}}}\hat{\textbf{q}}(t)=\hat{\textbf{q}}(t)-\hat{\textbf{o}}(t)$, at the current time $t=0$, the homotopy class of chasing path is determined as follows: 
\begin{equation}
\label{eq:topology_check}
  \begin{cases}
    &\textit{Class O1} \  (\text{if}\ \det ([{}_{\hat{\textbf{o}}}\textbf{p}_{c}^{\top}(t); {}_{\hat{\textbf{o}}}\hat{\textbf{q}}^{\top}(t)]) \geq 0, \text{at}\ t=0)\\
    &\textit{Class O2} \ (\text{if}\ \det ([{}_{\hat{\textbf{o}}}\textbf{p}_{c}^{\top}(t); {}_{\hat{\textbf{o}}}\hat{\textbf{q}}^{\top}(t)]) < 0, \text{at}\ t=0)
  \end{cases}
\end{equation}
where $[A;B]$ means the row-wise concatenation of matrices $A$ and $B$.
After conducting the homotopy check above, we define the visibility constraint and reference chasing trajectory.
\subsection{Target Visible Region}
\label{subsec:target_visible_region}
To robustly maintain the target visibility despite the prediction error, the target visible region TVR is defined considering the reachable areas of the moving objects: $\mathcal{R}_{\textbf{q}}(t)$ and $\mathcal{R}_{\textbf{o}}(t)$.
\subsubsection{Target visibility against obstacles}
\label{subsubsec:target_visibility_obstacles}
We define the target visible region against an obstacle (TVR-O), for each obstacle. To maximize a visible area of the targets' reachable area that is not occluded by the reachable area of obstacles, TVR-O is set to a half-space so that it includes the target's reachable area and minimizes the area that overlaps with the obstacle's reachable area. There are two cases where the reachable areas of the target and obstacles overlap and do not overlap, and TVR-O is defined accordingly.

\begin{figure}[t!]
\begin{subfigure}[b]{0.24\textwidth}
\centering\includegraphics[width=\textwidth]{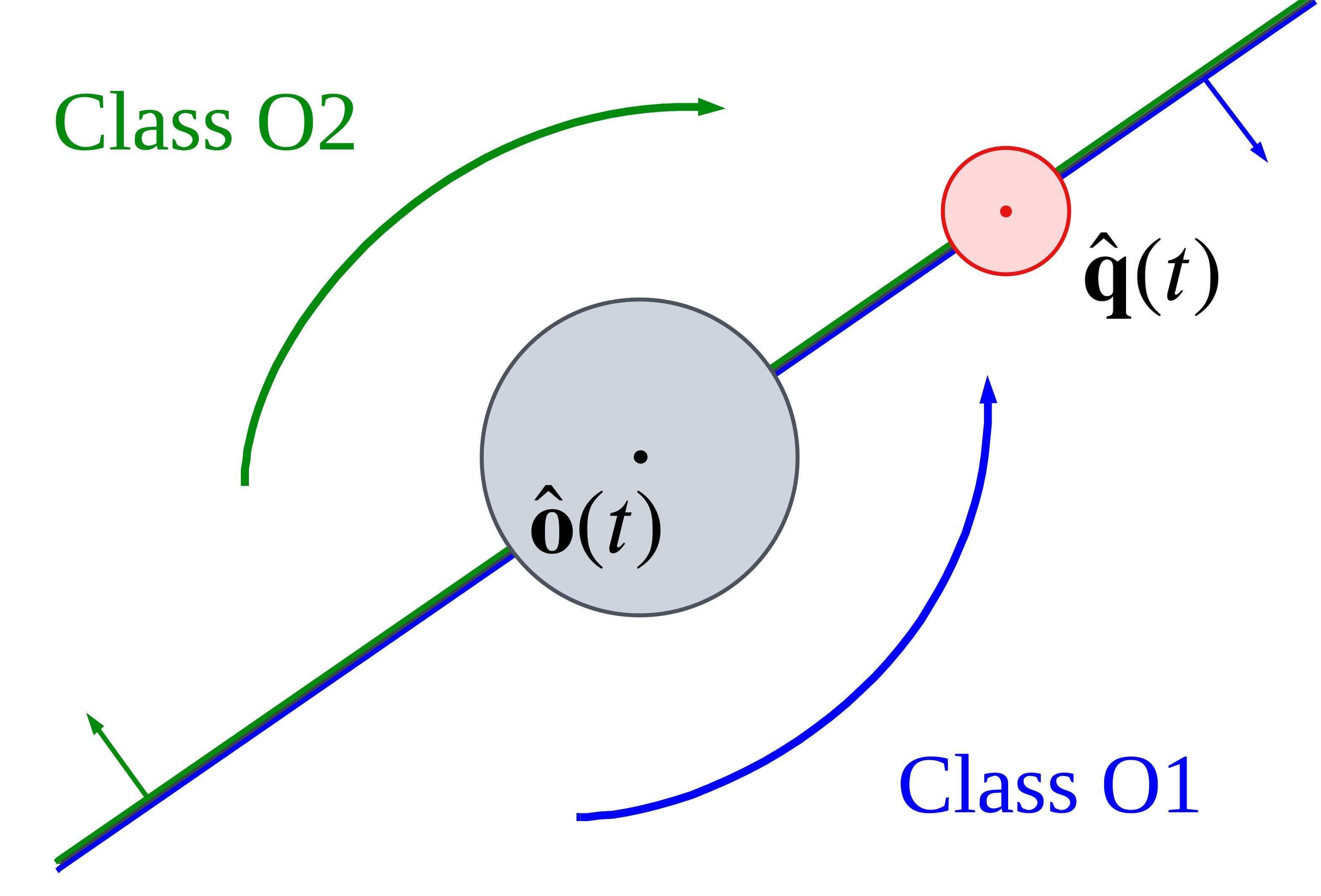}
\caption{Homotopy class}
\label{fig:TopologyClass}
\end{subfigure}
\hfill
\begin{subfigure}[b]{0.24\textwidth}
\centering
\includegraphics[width=\textwidth]{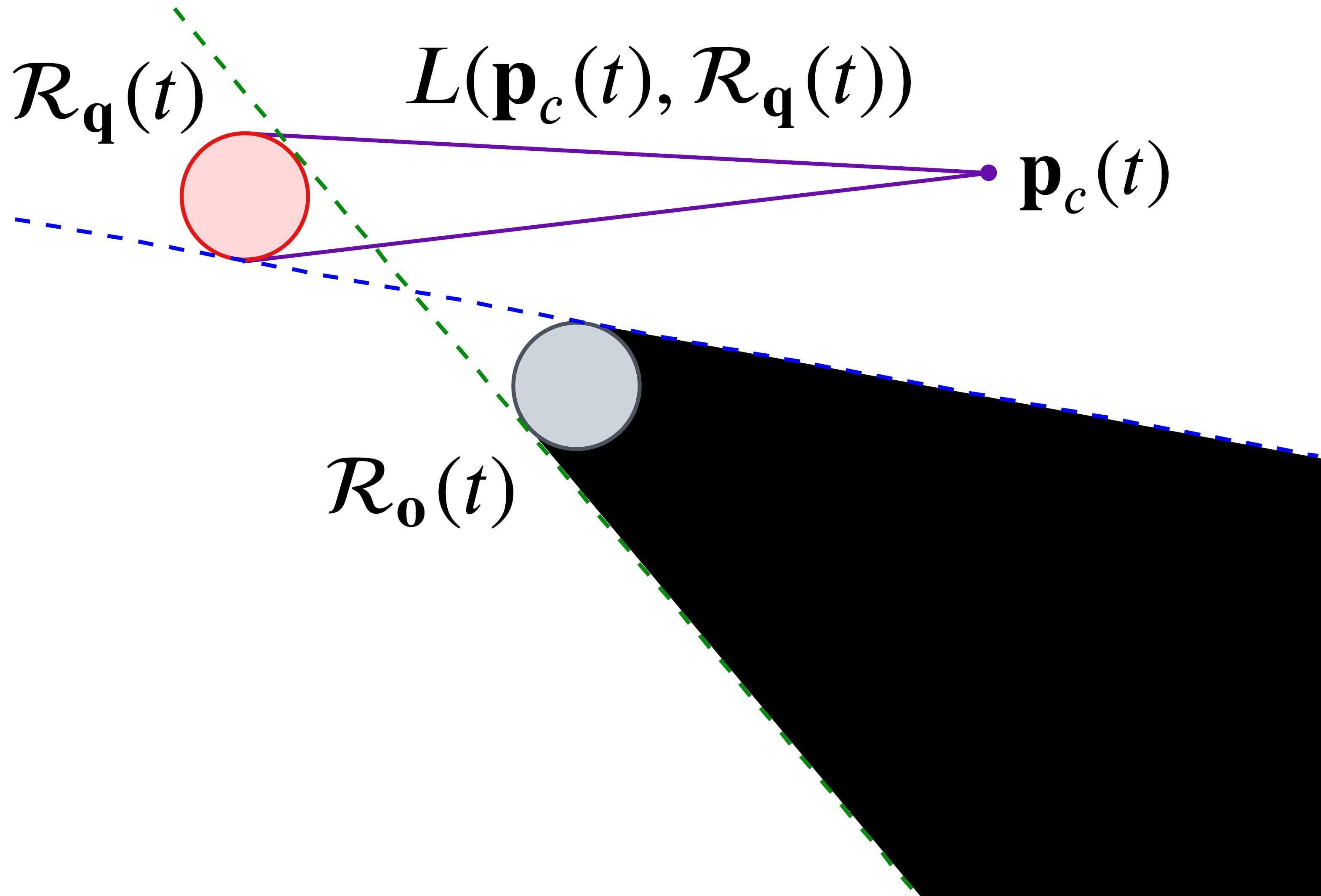}
\caption{Occlusion area}
\label{fig:OcclusionArea}
\end{subfigure}
\hfill
\begin{subfigure}[b]{0.24\textwidth}
\centering
\includegraphics[width=\textwidth]{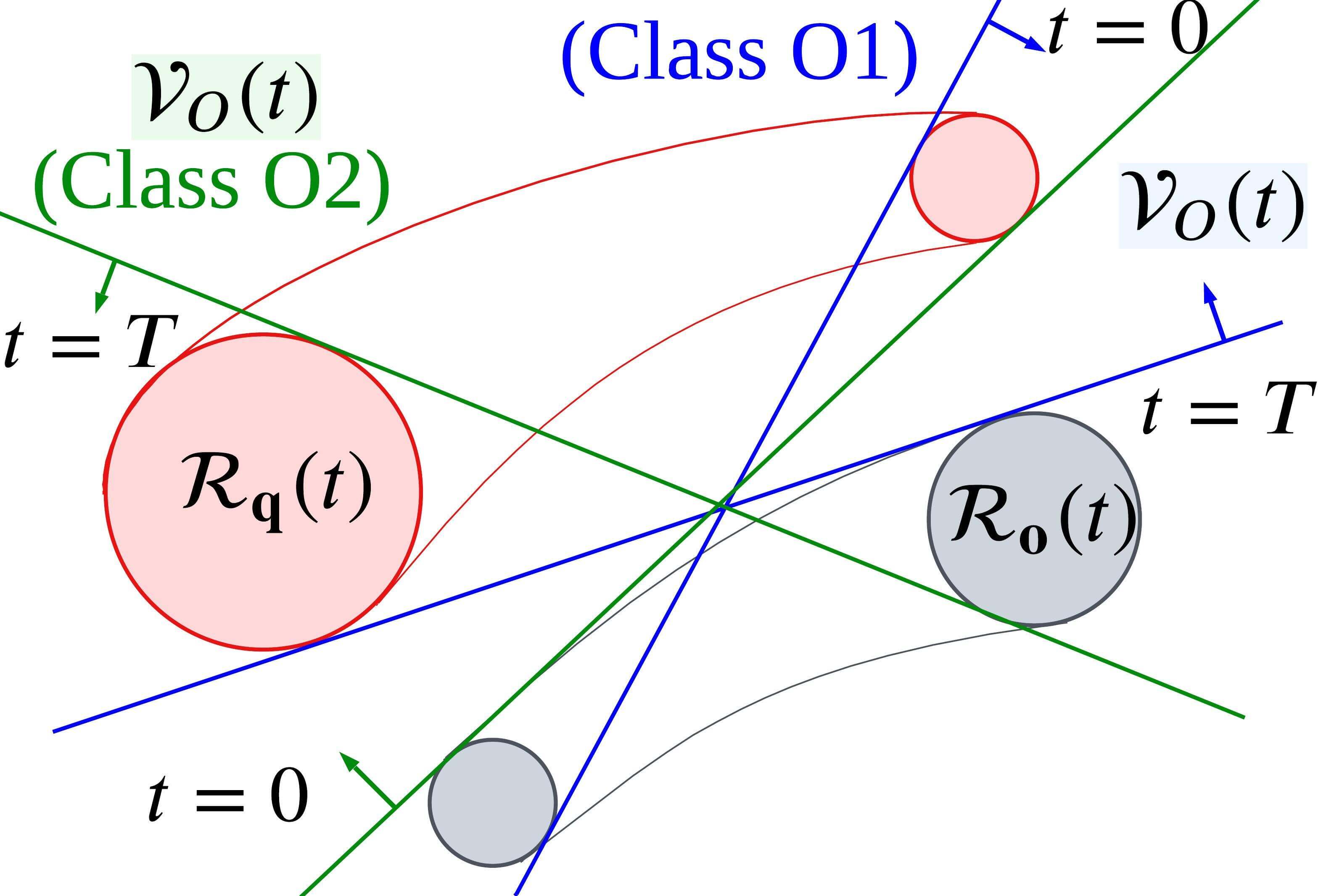}
\caption{TVR-O $(\mathcal{R}_{\textbf{o}}(t) \cap \mathcal{R}_{\textbf{q}}(t) = \emptyset)$}
\label{fig:TVRO1}
\end{subfigure}
\hfill
\begin{subfigure}[b]{0.24\textwidth}
\centering\includegraphics[width=\textwidth]{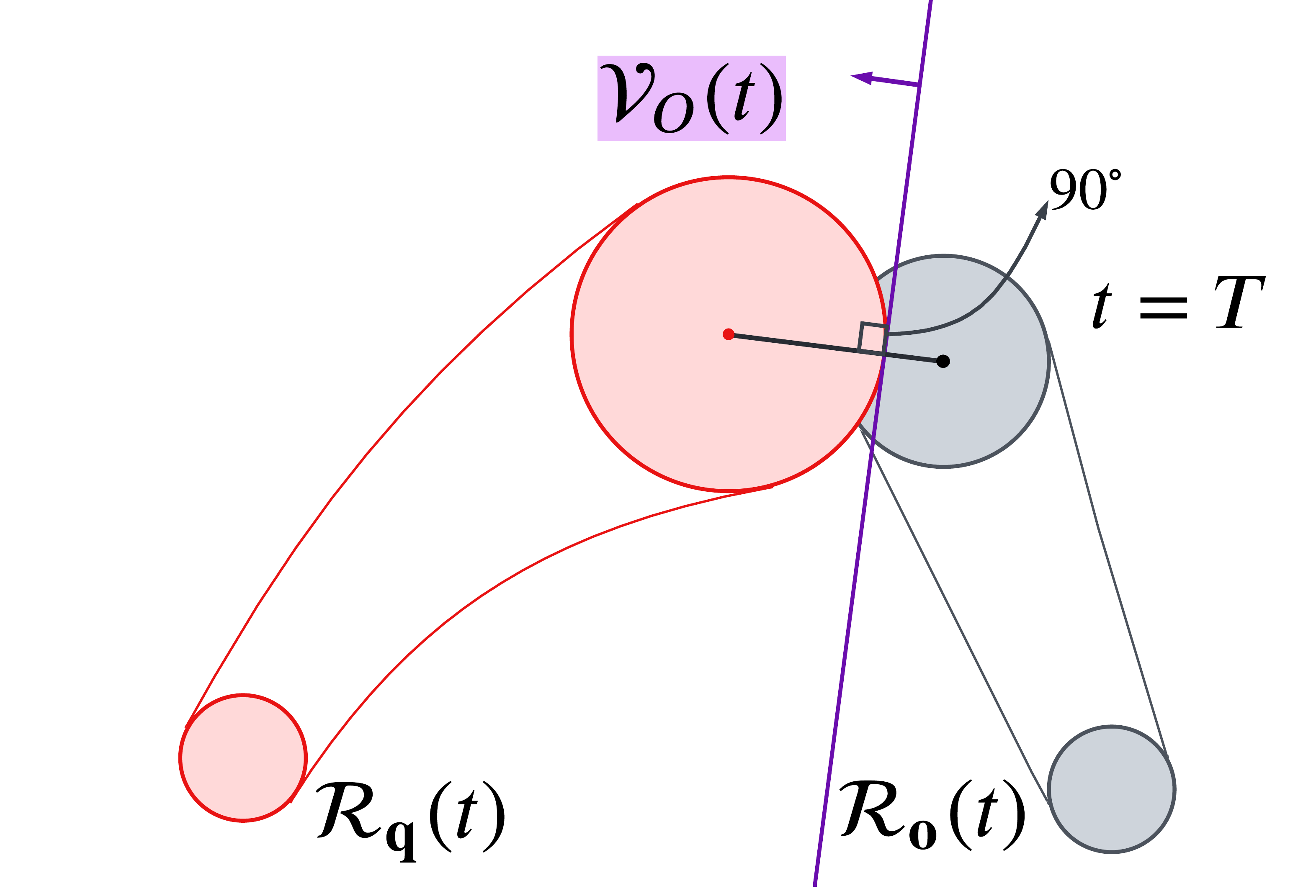}
\caption{TVR-O $(\mathcal{R}_{\textbf{o}}(t) \cap \mathcal{R}_{\textbf{q}}(t) \neq \emptyset)$}
\label{fig:TVRO2}
\end{subfigure}
\hfill
\begin{subfigure}[b]{0.24\textwidth}
\centering\includegraphics[width=\textwidth]{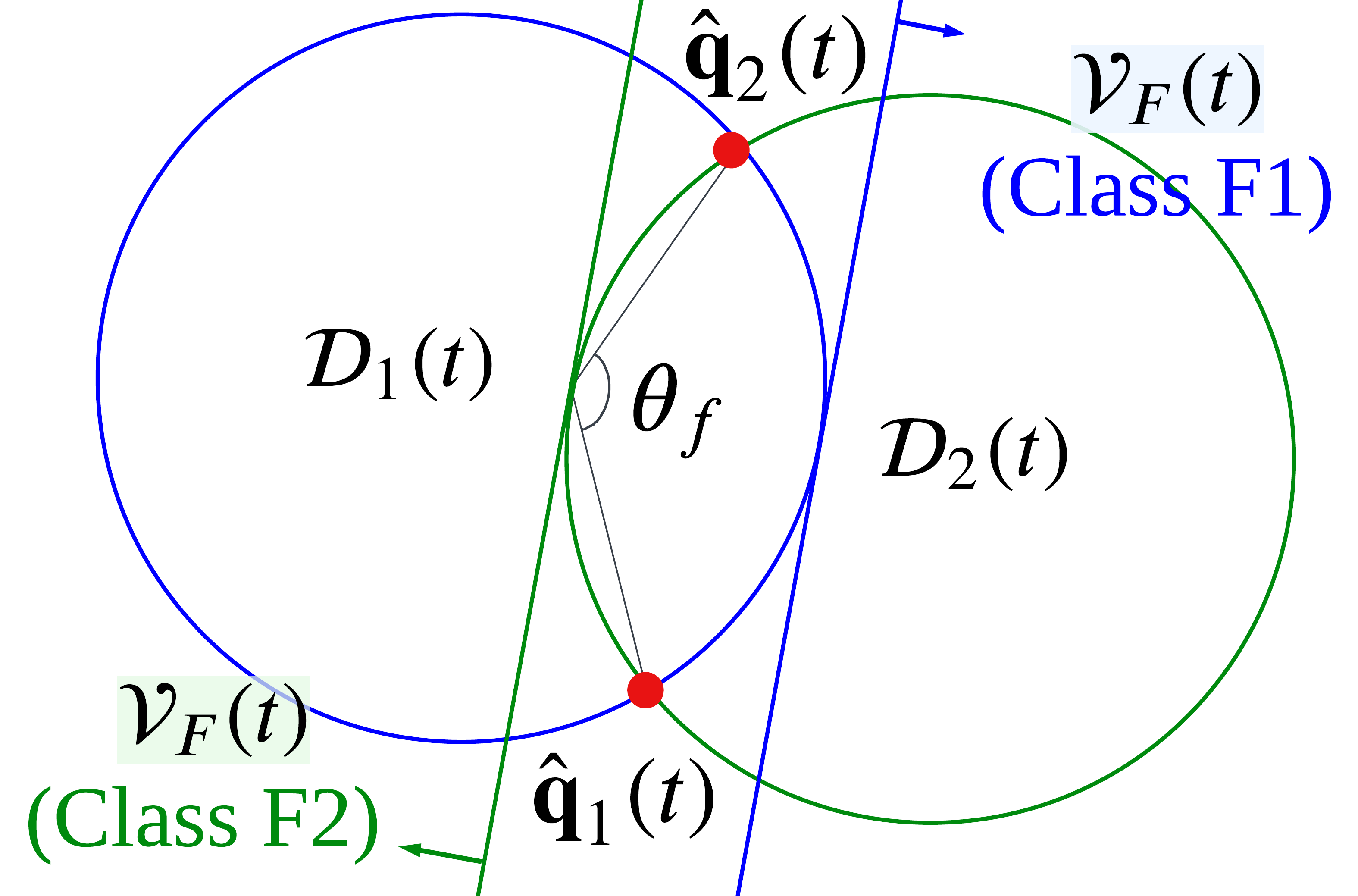}
\caption{TVR-F}
\label{fig:TVRF}
\end{subfigure}
\hfill
\centering
\begin{subfigure}[b]{0.24\textwidth}
\centering
\includegraphics[width=\textwidth]{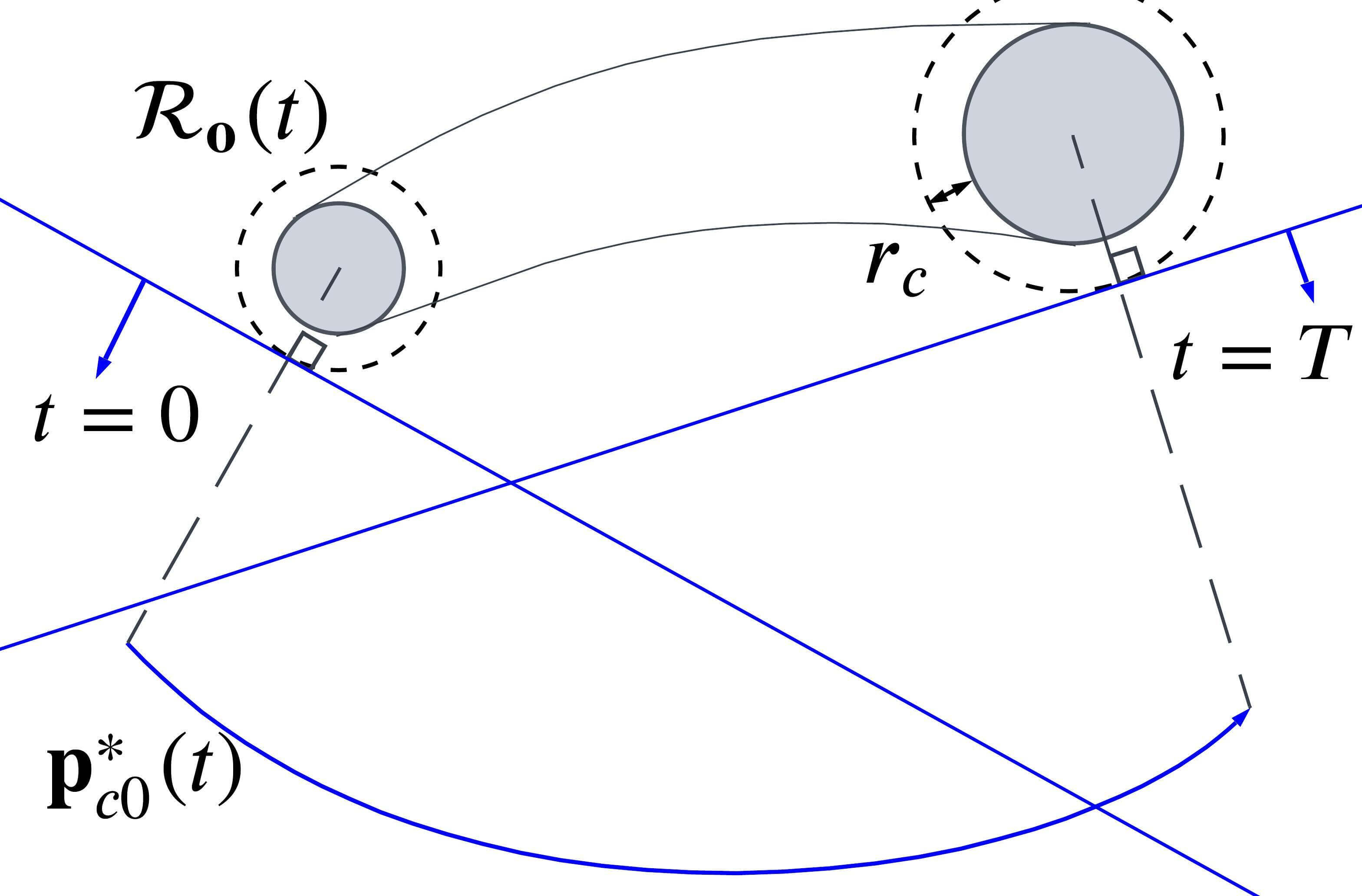}
\caption{Collision-free region}
\label{fig:Collision_constraint}
\end{subfigure}
\caption{\textbf{(a)}: Homotopy class against a single obstacle. \textbf{(b)}: A drone in a black-colored area cannot see the whole reachable area of the target.
\textbf{(c)} TVR-O (blue: \textit{Class O1}, green: \textit{Class O2}) when the reachable area of the target (red) and obstacles (grey) are distant from each other. \textbf{(d)} TVR-O (purple) when the reachable areas of the target (red) and obstacles (grey) overlap. \textbf{(e)} TVR-F (blue: \textit{Class F1}, green: \textit{Class F2}). \textbf{(f)}: Collision-free region (blue) against an obstacle (grey).}
\label{fig:TVR}
\end{figure}

\textbf{Case 1}: In the case where the reachable areas of the target and an obstacle do not overlap, the target invisible region is made by straight lines which are tangential to both $\mathcal{R}_{\textbf{q}}(t)$ and $\mathcal{R}_{\textbf{o}}(t)$, as shown in Fig. \ref{fig:OcclusionArea}. Accordingly, we define TVR-O as a half-space made by a tangential line between $\mathcal{R}_{\textbf{q}}(t)$ and $\mathcal{R}_{\textbf{o}}(t)$ according to the homotopy class (\ref{eq:topology_check}), as shown in Fig. \ref{fig:TVRO1}. $\mathcal{V}_{O}(t)$ denotes TVR-O and is represented as follows.
\begin{equation}
\label{eq:TVR}
\begin{aligned}
    \mathcal{V}_{O}(t) = \Big \{\textbf{x}(t)|\
    {}_{\hat{\textbf{o}}}\hat{\textbf{q}}^{\top}(t)\begin{bmatrix}r_{qo}(t)& \mp d_2(t) \\ \pm d_2(t)& r_{qo}(t) \end{bmatrix}{}_{\hat{\textbf{o}}}\textbf{x}(t)\\
    - r_o(t)d_{1}^{2}(t) \geq 0 \Big\}
\end{aligned}
\end{equation}
where $r_{qo}(t) = r_q(t)+r_o(t)$, $d_1(t) =\|{}_{\hat{\textbf{o}}}\hat{\textbf{q}}(t)\|_{2}$, and $d_2(t) = \sqrt{d_{1}^{2}(t)-r_{qo}^{2}(t)}$. The double signs in (\ref{eq:TVR}) are in the same order, where the lower and upper signs are for \textit{Class O1} and \textit{Class O2}, respectively.
\begin{lemma}
If $\textbf{p}_{c}(t)\in \mathcal{V}_{O}(t)$,\ $\mathcal{L}(\textbf{p}_{c}(t),\mathcal{R}_{\textbf{q}}(t))\cap \mathcal{R}_{\textbf{o}}(t)=\emptyset$ is satisfied.
\end{lemma}
\begin{proof}
$\mathcal{V}_{O}(t)$ is a half-space which is convex, and both $\textbf{p}_{c}(t)$ and $\mathcal{R}_{\textbf{q}}(t)$ belong to $\mathcal{V}_{O}(t)$. By the definition of convexity, all the \textit{Lines-of-Sight} connecting the drone $\textbf{p}_{c}(t)$ and all points in the reachable area of the target $\mathcal{R}_{\textbf{q}}(t)$ are included in $\mathcal{V}_{O}(t)$. Since the TVR-O defined in (\ref{eq:TVR}) is disjoint with $\mathcal{R}_{\textbf{o}}(t)$, $\mathcal{L}(\textbf{p}_{c}(t),\mathcal{R}_{\textbf{q}}(t))\cap \mathcal{R}_{\textbf{o}}(t)=\emptyset$.
\end{proof}
\begin{remark}
In dual-target scenarios, in order to avoid the occlusion of a target by another target, (\ref{eq:TVR}) is additionally defined by considering one of the targets as an obstacle.
\end{remark}

\textbf{Case 2}: For the case where the reachable areas overlap, TVR-O becomes a half-space made by a tangential line to $\mathcal{R}_{\textbf{o}}(t)$, which is perpendicular to a line segment connecting the centers of the target $\hat{\textbf{q}}(t)$ and the obstacles $\hat{\textbf{o}}(t)$, illustrated as straight lines in Fig. \ref{fig:TVRO2}. Then the TVR-O is represented as
\begin{equation}
    \label{eq:TVR_except}
    \mathcal{V}_{O}(t)= \{\textbf{x}(t)|  
    {}_{\hat{\textbf{o}}}\hat{\textbf{q}}^{\top}(t){}_{\hat{\textbf{q}}}\textbf{x}(t)+ r_{q}(t)d_{1}(t)\geq 0 \}
\end{equation}

The reachable areas include the volumes of the moving objects along with the prediction error. Since TVR-O is defined based on the reachable areas, it enhances the visibility of the entire body of the target against the prediction error. For detailed formulations of (\ref{eq:TVR}) and (\ref{eq:TVR_except}), see Appendix \ref{appendix:tvro_formulation}.
\subsubsection{Bernstein polynomial approximation}
\label{subsubsec:bernstein_approximation}
In defining TVR-O, all terms in (\ref{eq:TVR}) and (\ref{eq:TVR_except}) are polynomials except $d_1(t)$ and $d_2(t)$. To make  $\textbf{p}_{c}(t)\in \mathcal{V}_{O}(t)$ affine with respect to the optimization variable $\underbar{\textbf{c}}$, we first convert $d_1(t), d_2(t)$ into Bernstein polynomials using the following numerical technique that originates from \textit{Lagrange interpolation} with standard polynomial representation \cite{lagrange}. 

Suppose that, by using \textit{Lagrange interpolation}, a non-polynomial function $f(t)$ is approximated by an $\tilde{n}$-th degree Bernstein polynomial function $\tilde{f}(t)$ over the time interval $[t_{a},t_{b}]$:
\begin{equation}\label{eq6:lagrangian_approximation}
    \tilde{f}(t)=\tilde{\textbf{f}}_{\tilde{n}}^{\top}\textbf{b}_{\tilde{n}}(t), \quad t\in[t_{a},t_{b}]
\end{equation}
where $\tilde{\textbf{f}}_{\tilde{n}}\in\mathbb{R}^{\tilde{n}+1}$ and $\textbf{b}_{\tilde{n}}(t)\in\mathbb{R}^{\tilde{n}+1}$ are a coefficient vector and a Bernstein basis vector for $[t_{a},t_{b}]$, respectively. To calculate the coefficients of approximated function $\tilde{\textbf{f}}_{\tilde{n}}$, we utilize Bernstein Vandermonde matrix $B_{\tilde{n}}$ \cite{vandermonde} as follows.
\begin{equation}
    \begin{aligned}
        & &&\tilde{\textbf{f}}_{\tilde{n}}= B^{-1}_{\tilde{n}}\bar{\textbf{f}}_{\tilde{n}},\\
        &\text{where}&&B_{\tilde{n}} = \{b_{j,k}\}\in{\mathbb{R}^{(\tilde{n}+1)\times(\tilde{n}+1)}},\\
        & &&b_{j,k}= \binom{\tilde{n}}{k}{\Big(\dfrac{j}{\tilde{n}}\Big)}^{k}{\Big(\dfrac{\tilde{n}-j}{\tilde{n}}\Big)}^{\tilde{n}-k},\ 0\leq j, k \leq \tilde{n},\\
        & &&\bar{\textbf{f}}_{\tilde{n}}= [f_{0},f_{1},\ldots,f_{\tilde{n}}]^{\top},\ f_l =f((1-\frac{l}{\tilde{n}})t_a+\frac{l}{\tilde{n}}t_b)
    \end{aligned}
\end{equation}
Using this technique,  $d_1(t)$ and $d_2(t)$ in (\ref{eq:TVR}) and (\ref{eq:TVR_except}) are approximately represented as $\tilde{d}_1(t)$ and $\tilde{d}_2(t)$, respectively.
With the approximated terms, we represent the target visibility constraint as follows.
\begin{subequations}
\label{eq:visibility_constraint_revised}
\begin{align}
    \label{subeq:tvro1}
    &\textbf{Case 1}:\ t\in [T_{m-1},T_{m}],\ \text{s.t. } \mathcal{R}_{\textbf{q}}(t)\cap \mathcal{R}_{\textbf{o}}(t) = \emptyset : \\ \nonumber
    & \quad \quad {}_{\hat{\textbf{o}}}\hat{\textbf{q}}^{\top}(t)\begin{bmatrix}r_{qo}(t)& \mp \tilde{d}_2(t) \\ \pm \tilde{d}_2(t)& r_{qo}(t) \end{bmatrix}{}_{\hat{\textbf{o}}}\textbf{p}_{c}(t) - r_o(t){d}_{1}^{2}(t) \geq 0 \\
    \label{subeq:tvro2}
    &\textbf{Case 2}:\ t\in [T_{m-1},T_{m}],\ \text{s.t. } \mathcal{R}_{\textbf{q}}(t)\cap \mathcal{R}_{\textbf{o}}(t) \neq \emptyset : \\ \nonumber
    & \quad \quad {}_{\hat{\textbf{o}}}\hat{\textbf{q}}^{\top}(t){}_{\hat{\textbf{q}}}\textbf{p}_{c}(t) - r_{q}(t)\tilde{d}_1(t)\geq 0
\end{align}
\end{subequations}
\subsubsection{FOV constraints}
\label{subsubsec:fov_constraints} 
In addition to (\ref{eq:visibility_constraint_revised}), the drone should avoid the region where it cannot see the two targets with its limited camera FOV. As shown in Fig. \ref{fig:TVRF}, circles whose inscribed angle of an arc tracing points $Q_{1}$ and $Q_{2}$ at $\hat{\textbf{q}}_{1}(t)$, $\hat{\textbf{q}}_{2}(t)$ equals $\theta_{f}$, are represented as follows:
\begin{equation}
\label{eq:FOV_circle}
\begin{aligned}
    &\mathcal{D}_{j}(t) = \big\{\textbf{x}(t)| \|\textbf{x}(t)-\textbf{f}_j(t)\|_{2}\leq r_{f}(t)\big\},\ j=1,2 \ ,\\
    & \textbf{f}_{j}(t) = \frac{1}{2}(\hat{\textbf{q}}_{1}(t)+\textbf{q}_{2}(t))+\frac{\cot{\theta_{f}}}{2}R\Big((-1)^{j}\frac{\pi}{2}\Big){}_{\hat{\textbf{q}}_{1}}\hat{\textbf{q}}_{2}(t)\\
    & r_{f}(t) = \frac{1}{2\sin{\theta_{f}}}\|{}_{\hat{\textbf{q}}_{1}}\hat{\textbf{q}}_{2}(t)\|_{2},\ {}_{\hat{\textbf{q}}_{1}}\hat{\textbf{q}}_{2}(t) = \hat{\textbf{q}}_{2}(t)-\hat{\textbf{q}}_{1}(t)
\end{aligned}
\end{equation}
where $R(\theta)$ is the rotation matrix for a rotation of $\theta$.  According to the geometric property of an inscribed angle, $\angle Q_{1}XQ_{2} > \theta_{f}$, $\angle Q_{1}YQ_{2} < \theta_{f}$  for $\forall X \in \mathcal{D}_{z}(t)\setminus \partial\mathcal{D}_{z}(t)$, $\forall Y \notin \mathcal{D}_{z}(t)$, where $D_{z}(t)$ is represented as follows.
\begin{equation}
    \label{eq:FOV_dead_zone}    
    \mathcal{D}_{z}(t)= \begin{cases}
     &\mathcal{D}_{1}(t)\cap\mathcal{D}_{2}(t) \quad (\text{if}\ \theta_{f}\geq \frac{\pi}{2})\\
     &\mathcal{D}_{1}(t)\cup\mathcal{D}_{2}(t) \quad (\text{if}\ \theta_{f}< \frac{\pi}{2})
     \end{cases}
\end{equation}
Since the camera FOV is $\theta_{f}$, the drone inside $\mathcal{D}_{z}(t)$ misses at least one target in the camera image inevitably, and the drone outside of $\mathcal{D}_{z}(t)$ is able to see both targets.

We define the target visible region considering camera FOV (TVR-F) as a half-space that does not include $\mathcal{D}_{z}(t)$ and is tangential to $\mathcal{D}_{z}(t)$, as shown in Fig. \ref{fig:TVRF}. $\mathcal{V}_{F}(t)$ denotes TVR-F and is represented as follows.
\begin{equation}
    \label{eq:fov_constraint}
    \mathcal{V}_{F}(t)=\{\textbf{x}(t)| \pm\text{det} ( \begin{bmatrix}
        {}_{\hat{\textbf{q}}_1}\textbf{x}^{\top}(t)\\ {}_{\hat{\textbf{q}}_{1}}\hat{\textbf{q}}_{2}^{\top}(t) 
    \end{bmatrix})-\frac{1+\cos{\theta_{f}}}{2\sin{\theta_{f}}}\|{}_{\hat{\textbf{q}}_{1}}\hat{\textbf{q}}_{2}(t) \|_{2}^{2} \geq 0\}
\end{equation}
The upper and lower signs in (\ref{eq:fov_constraint}) are for \textit{Class F1} and \textit{Class F2}, respectively, which are defined as
\begin{equation}
\label{eq:topology_fov_check}
  \begin{cases}
    &\textit{Class F1} \  (\text{if}\ \det ([{}_{\hat{\textbf{q}}_{1}}\textbf{p}_{c}^{\top}(t);{}_{\hat{\textbf{q}}_{1}}\hat{\textbf{q}}_{2}^{\top}(t)]) \geq 0, \text{at}\ t=0)\\
    &\textit{Class F2} \ (\text{if}\ \det ([{}_{\hat{\textbf{q}}_{1}}\textbf{p}_{c}^{\top}(t);{}_{\hat{\textbf{q}}_{1}}\hat{\textbf{q}}_{2}^{\top}(t)]) < 0, \text{at}\ t=0).
  \end{cases}
\end{equation}
We enforce the drone to satisfy $\textbf{p}_{c}(t)\in \mathcal{V}_{F}(t)$ in order to see both targets. For the details, see the Appendix \ref{appendix:tvrf_formulation}.
\subsection{Collision Avoidance}
\label{subsec:collision_avoidance}
To avoid collision with obstacles, we define a collision-free area as a half-space made by a tangential line to a set that is inflated by $r_{c}$ from $\mathcal{R}_{\textbf{o}}(t)$. The tangential line is drawn considering a trajectory $\textbf{p}_{c0}^{*}(t)$, which is the previous planning result. This is illustrated in Fig. \ref{fig:Collision_constraint} and represented as follows.
\begin{equation}
\label{eq:collisiion_constraint}
    \mathcal{F}_{c}(t) = \big\{\textbf{x}(t)| \frac{{}_{\hat{\textbf{o}}}\textbf{p}_{c0}^{*\top}(t)}{\|{}_{\hat{\textbf{o}}}\textbf{p}_{c0}^{*}(t)\|_{2}}(\textbf{x}(t)-\hat{\textbf{o}}(t))\geq (r_{o}(t)+r_{c}) \big\}
\end{equation}
where ${}_{\hat{\textbf{o}}}\textbf{p}_{c0}^{*}(t):=\textbf{p}_{c0}^{*}(t)-\hat{\textbf{o}}(t)$. As in (\ref{eq6:lagrangian_approximation}), we approximate a non-polynomial term  $\| {}_{\hat{\textbf{o}}}\textbf{p}_{c0}^{*}(t)\|_{2}$ to a polynomial $\tilde{d}_{3}(t)$. With the approximated  term $\tilde{d}_{3}(t)$, the collision constraint is defined as follows:
\begin{equation}
\label{eq:collisiion_constraint_revised}
{}_{\hat{\textbf{o}}}\textbf{p}_{c0}^{*\top}(t){}_{\hat{\textbf{o}}}\textbf{p}_{c}(t)- (r_{o}(t)+r_{c})\tilde{d}_{3}(t)\geq 0
\end{equation}

Due to the fact that the multiplication of Bernstein polynomials is also a Bernstein polynomial, the left-hand side of the (\ref{eq:visibility_constraint_revised}), (\ref{eq:fov_constraint}), and (\ref{eq:collisiion_constraint_revised})  can be represented in a Bernstein polynomial form. With the non-negativeness of Bernstein basis, we make coefficients of each basis non-negative in order to keep the left-hand side non-negative, and (\ref{eq:visibility_constraint_revised}), (\ref{eq:fov_constraint}), and (\ref{eq:collisiion_constraint_revised}) turn into affine constraints with the decision vector $\underbar{\textbf{c}}$. The constraints can be written as follows, and we omit details.
\begin{equation}
\label{eq:affine_constraint}
    \begin{aligned}
        A_{\text{TVR-O}}\underbar{\textbf{c}}-\textbf{b}_{\text{TVR-O}} &\geq 0, \\
        A_{\text{TVR-F}}\underbar{\textbf{c}}-\textbf{b}_{\text{TVR-F}} &\geq 0,\\
        A_{\text{Colli}}\underbar{\textbf{c}}-\textbf{b}_{\text{Colli}}& \geq 0
    \end{aligned}
\end{equation}
\subsection{Reference Trajectory for Target Tracking}
\label{subsec:reference_trajectory}
In this section, we propose a reference trajectory for target chasing that enhances the visibility of the targets.
\subsubsection{Single-target case}
\label{subsubsec:reference_single_target}
\begin{figure}[t!]
\centering
\begin{subfigure}[t]{0.24\textwidth}
\centering
\includegraphics[width=\textwidth]{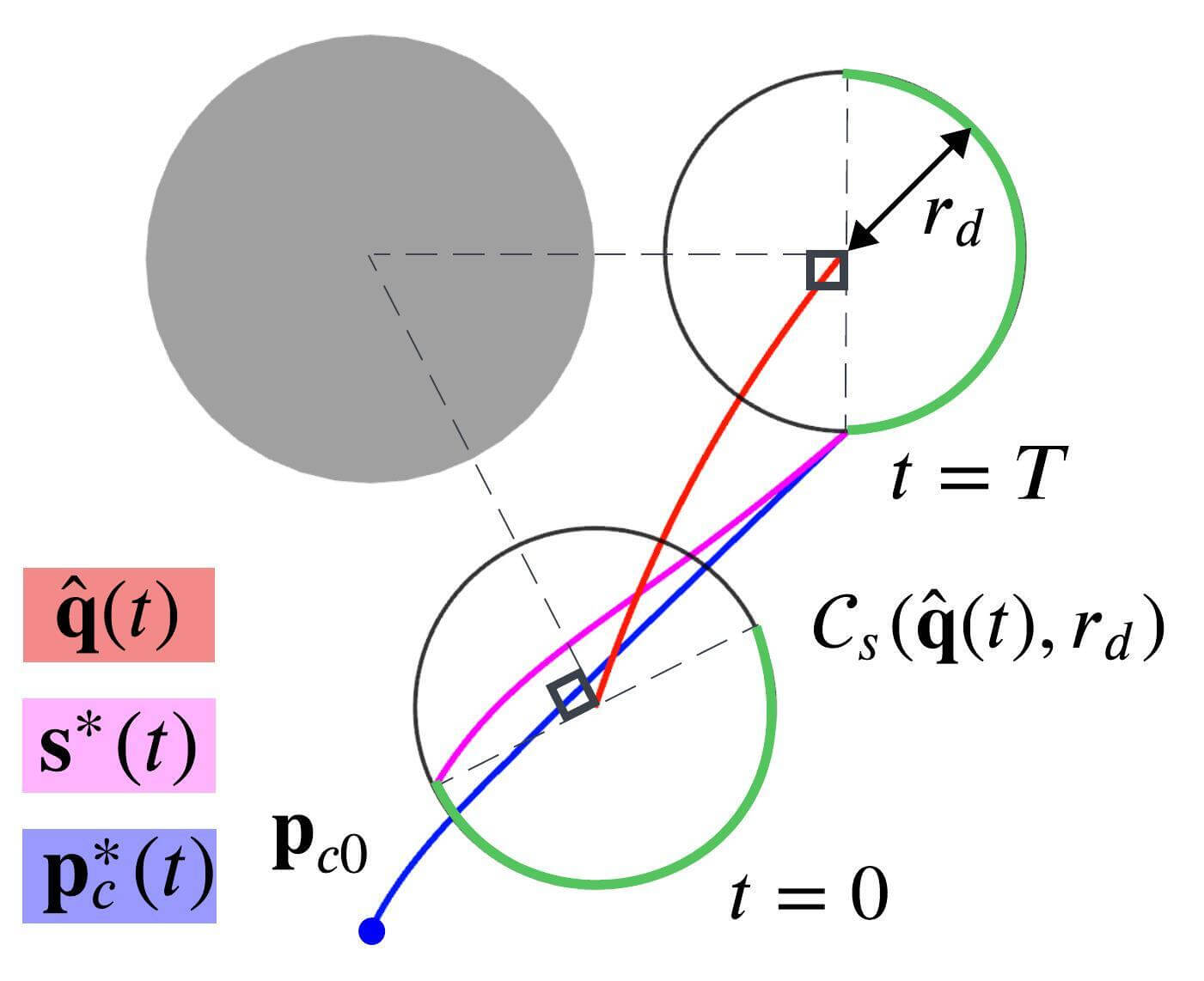}
\caption{Single-target}
\label{fig:SingleRefTraj}
\end{subfigure}
\hfill
\begin{subfigure}[t]{0.24\textwidth}
\centering
\includegraphics[width=\textwidth]{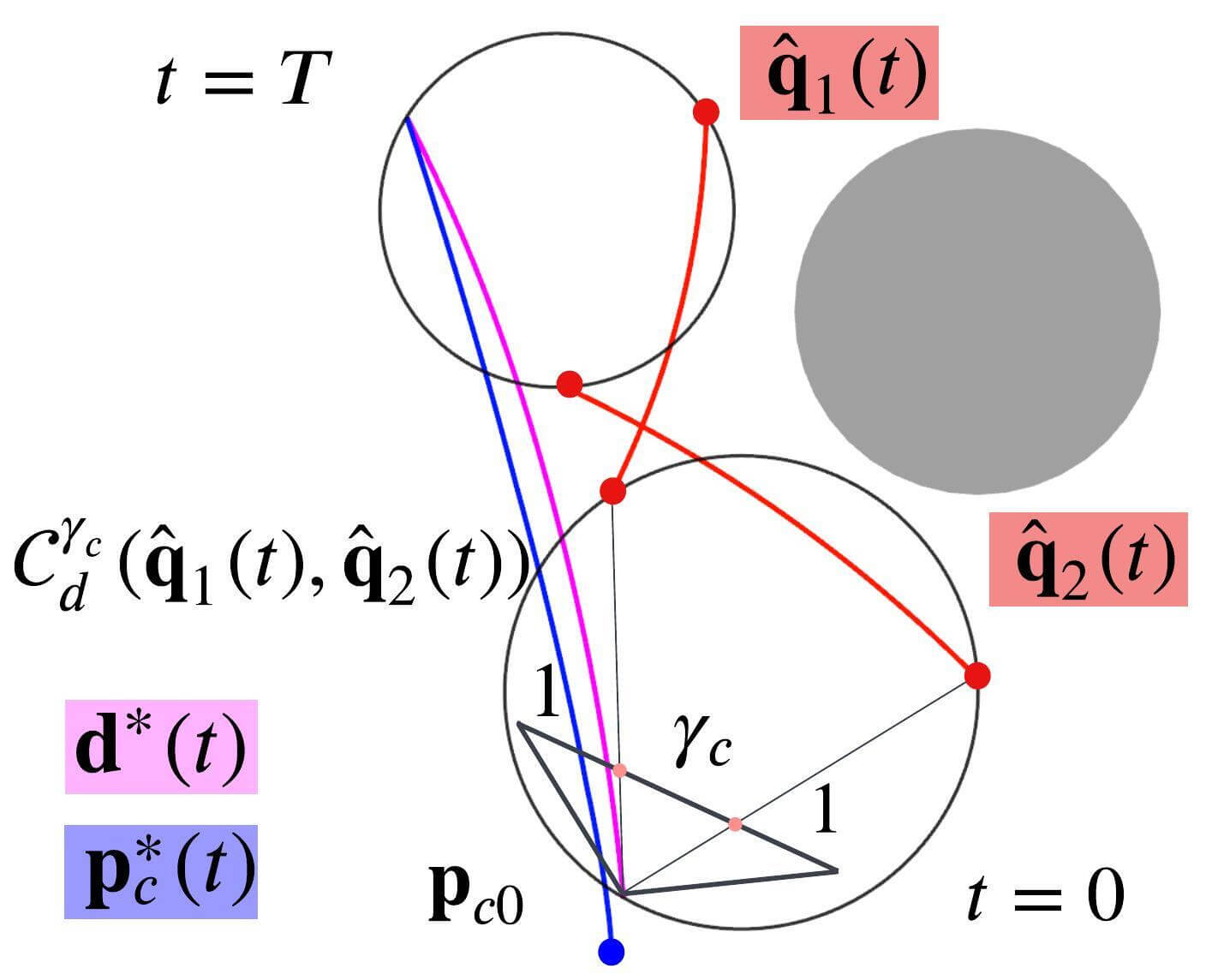}
\caption{Dual-target}
\label{fig:DualRefTraj}
\end{subfigure}
\caption{Reference trajectory design in single- and dual-target (red) cases among obstacles (grey). \textbf{(a)}: The points having the maximum visibility score $\psi$ in $\mathcal{C}_{s}(\hat{\textbf{q}}(t),r_d)$ (green), a  trajectory with the maximum $\psi$, $\textbf{s}^{*}(t)$  (magenta), and the reference trajectory $\textbf{p}^{*}_{c}(t)$ (blue line) considering current position of the drone (blue dot). \textbf{(b)}: $\mathcal{C}_{d}^{\gamma_{c}}(\hat{\textbf{q}}_{1}(t),\hat{\textbf{q}}_{2}(t))$ (black circle). The trajectory with  high $\psi$, $\textbf{d}^{*}(t)$ (magenta), and the reference trajectory $\textbf{p}_{c}^{*}(t)$ (blue line) considering current position of the drone (blue dot).}
\end{figure}

In designing the reference trajectory, we use the visibility score, as discussed in \cite{boseong_iros}.
The definition of the visibility score $\psi$ is the closest distance between all points in an $j$-th obstacle, $\mathcal{O}_j$, and the \textit{Line-of-Sight} connecting the target and the drone:
\begin{equation}
  \psi(\textbf{p}_{c}(t);\hat{\textbf{q}}(t),\mathcal{O}_{j}) =\min_{\substack{\textbf{x}\in \mathcal{L}(\textbf{p}_{c}(t),\hat{\textbf{q}}(t))\\\textbf{y}\in\mathcal{O}_{j}}}
  \|\textbf{x}-\textbf{y}\|_{2}.
  \label{eq:visibility_metric}
\end{equation}
In order to keep the projected size of the target on the camera image, we set the desired shooting distance $r_{d}$. With the desired shooting distance $r_d$, a viewpoint candidate set can be defined as  $\mathcal{C}_{s}(\hat{\textbf{q}}(t),r_d)$ $=\{\textbf{x}(t)| \|\textbf{x}(t)-\hat{\textbf{q}}(t)\|_{2}=r_{d} \}$. Under the assumption that the environment consists of cylindrical obstacles, half-circumference of $\mathcal{C}_{s}(\hat{\textbf{q}}(t),r_d)$ acquires the maximum visibility score, as illustrated in green in Fig. \ref{fig:SingleRefTraj}. Therefore, the following trajectory maintains the maximum $\psi(\textbf{p}_{c}(t);\hat{\textbf{q}}(t),\mathcal{O}_{j})$.
\begin{equation}
  \label{eq:ref_traj_single}
  \textbf{s}_{j}(t)=
    \textbf{q}(t)+r_{d}R(\mp\frac{\pi}{2})\bm{\delta}_{s,j}(t),\ \bm{\delta}_{s,j}(t)= \frac{\hat{\textbf{q}}(t)-\hat{\textbf{o}}_{j}(t)}{\|\hat{\textbf{q}}(t)-\hat{\textbf{o}}_{j}(t)\|_2}
\end{equation}
The upper and lower signs are for \textit{Class O1} and \textit{Class O2}, respectively. We define the reference trajectory as a weighted sum of $\textbf{s}_{j}(t)$.
\begin{equation}
\label{eq:single_ref_traj_weighted}
    \textbf{s}^{*}(t) =  \sum_{j=1}^{N_{o}}w_{s,j}\textbf{s}_{j}(t),\quad \text{where} \sum_{j=1}^{N_{o}}w_{s,j}=1
\end{equation}
$w_{s,j}$'s are weight functions that are inversely proportional to the current distance between the target and each obstacle.
\subsubsection{Dual-target case}
\label{subsubsec:reference_dual_target}
In order to make aesthetically pleasing scenes, we aim to place two targets in a ratio of $1:\gamma_{c}:1$ on the camera image. To do so, the drone must be within a set $\mathcal{C}_{d}^{\gamma_{c}}(\hat{\textbf{q}}_{1}(t),\hat{\textbf{q}}_{2}(t))$, which is defined as follows and illustrated as a black circle in Fig. \ref{fig:DualRefTraj}. 
\begin{equation}
    \label{fig:ref_set_dual}
    \begin{aligned}
    &\mathcal{C}_{d}^{\gamma_{c}}(\hat{\textbf{q}}_{1}(t),\hat{\textbf{q}}_{2}(t))= \{\textbf{x}(t)|\|\textbf{x}(t)-\textbf{f}_{c}(t)\|_{2}=r_{f,c}\},\\
    & \textbf{f}_{c}(t) = \frac{1}{2}(\hat{\textbf{q}}_{1}(t)+\hat{\textbf{q}}_{2}(t))\pm \kappa_1 R(-\frac{\pi}{2})
    {}_{\hat{\textbf{q}}_{1}}\hat{\textbf{q}}_{2}(t),\\
    & r_{f,c}(t)= \kappa_{2}\|{}_{\hat{\textbf{q}}_1}\hat{\textbf{q}}_{2}(t) \|_{2},\\
    & \kappa_1=\frac{\gamma_{c}+2}{4\gamma_{c}}\cot{\frac{\theta_{f}}{2}}\Big(1-\frac{\gamma_{c}^{2}}{(\gamma_{c}+2)^{2}}\tan^{2}{\frac{\theta_f}{2}}\Big),\\
    &\kappa_{2}=\frac{\gamma_{c}+2}{4\gamma_{c}}\cot{\frac{\theta_{f}}{2}}\Big(1+\frac{\gamma_{c}^2}{(\gamma_{c}+2)^{2}}\tan^{2}{\frac{\theta_f}{2}}\Big)
    \end{aligned}
\end{equation}
The upper and lower signs are for \textit{Class F1} and \textit{Class F2}.\\
AA\\
We define the reference trajectory $\textbf{d}^{*}(t)$ to acquire a high visibility score $\psi$ while maintaining the ratio.
\begin{equation}
\label{eq:ref_traj_dual}
\begin{aligned}
    &\textbf{d}^{*}(t) = \textbf{f}_{c}(t)+ r_{f,c}(t)\frac{\bm{\delta}_{d}(t)}{\|\bm{\delta}_{d}(t)\|_{2}},\\ &\bm{\delta}_{d}(t) =\bm{\delta}_{b}(t)+\sum_{j=1}^{N_{o}}w_{s1,j}\bm{\delta}_{s1,j}(t)+w_{s2,j}\bm{\delta}_{s2,j}(t),\\
    &\bm{\delta}_{b}(t) = R(\mp\frac{\pi}{2})\frac{\hat{\textbf{q}}_{2}(t)-\hat{\textbf{q}}_{1}(t)}{\|\hat{\textbf{q}}_{2}(t)-\hat{\textbf{q}}_{1}(t)\|_{2}}
\end{aligned}
\end{equation}
The upper and lower signs in $\bm{\delta}_{b}(t)$ are for \textit{Class F1} and \textit{Class F2}, respectively, and $w_{s1,j},\bm{\delta}_{s1,j}(t)$ and $w_{s2,j}, \bm{\delta}_{s2,j}(t)$ correspond to the $w_{s,j}$, $\bm{\delta}_{s,j}(t)$ in (\ref{eq:ref_traj_single}) and (\ref{eq:single_ref_traj_weighted}) for the target 1 and 2. $\bm{\delta}_{b}(t)$ is defined as the farthest direction from the two targets to maximize a metric, $\psi(\textbf{p}_{c}(t);\hat{\textbf{q}}_{1}(t),\mathcal{B}(\hat{\textbf{q}}_{2}(t),r_{q0}))$ + $\psi(\textbf{p}_{c}(t);\hat{\textbf{q}}_{2}(t),\mathcal{B}(\hat{\textbf{q}}_{1}(t),r_{q0}))$, representing the visibility score between two targets.

For the numerical stability of the QP solver, the reference trajectory is redefined by interpolating between $\boldsymbol{\mu}^{*}(t)$, $\boldsymbol{\mu}=\textbf{s},\textbf{d}$ and the current drone's position $\textbf{p}_{c0}$. With a non-decreasing polynomial function $\alpha(t)$ that satisfies $\alpha(0)=0$ and $\alpha(T)=1$, the reference trajectory is defined as follows.
\begin{equation}
  \textbf{p}_{c}^{*}(t) = \big(1-\alpha(t)\big)\textbf{p}_{c0}+\alpha(t)\boldsymbol{\mu}^{*}(t)
  \label{eq:ref_traj}
\end{equation}
 Since trigonometric terms for $\boldsymbol{\mu}^{*}(t)$, $\boldsymbol{\mu}=$ $\textbf{s}$ or $\textbf{d}$  in (\ref{eq:single_ref_traj_weighted}), (\ref{eq:ref_traj_dual}) are non-polynomial, the \textit{Lagrange interpolation} is used as (\ref{eq6:lagrangian_approximation}).
 The approximated $\textbf{p}_{c}^{*}(t)$ is denoted as $\tilde{\textbf{p}}_{c}^{*}(t)$.
 Based on the construction of reference trajectory and the target visibility constraints, we formulate the trajectory optimization problem as a QP problem.
\subsection{QP Formulation}
\label{subsec:QP_formulation}
\subsubsection{Trajectory segmentation}
\label{subsubsec:segmentation}
\begin{figure}[t]
     \centering
     \includegraphics[width = 0.9\linewidth]{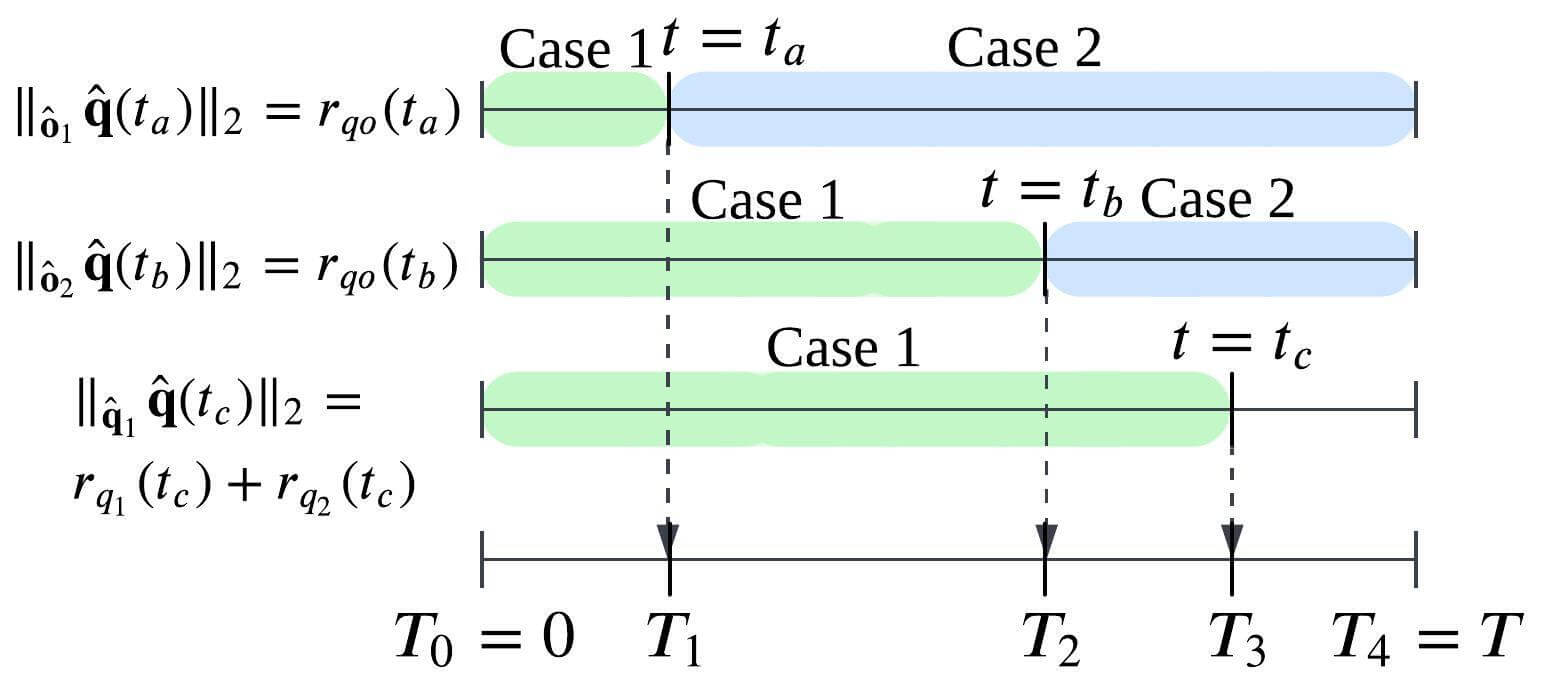}
     \caption{Time segmentation. Green and blue regions are time intervals when the (\ref{subeq:tvro1}) and (\ref{subeq:tvro2}) are adopted as the target visibility constraints, respectively.}
           \label{fig:trajectory_segmentation}
           \vspace{-4mm}
 \end{figure}

The constraint (\ref{eq:visibility_constraint_revised}) is divided into the cases: when the reachable areas are separated and when they overlap. To apply (\ref{eq:visibility_constraint_revised}) according to the situation, the roots of two equations $\|{}_{\hat{\textbf{o}}}\hat{\textbf{q}}(t)\|_{2}=r_{qo}(t)$ and $\|{}_{\hat{\textbf{q}}_{1}}\hat{\textbf{q}}_{2}(t)\|_{2}=r_{q1}(t)+r_{q2}(t)$ should be investigated for $t\in(0,T)$. We define $[T_{0},\ldots, T_{M}]$ by arranging the roots of the above equations in ascending order, and we set $M$ time intervals, as stated in (\ref{eq:segment_representation}). Fig. \ref{fig:trajectory_segmentation} visualizes the above process.
  Thanks to \textit{De Casteljau}'s algorithm \cite{bebot}, the single Bernstein polynomial such as $\hat{\textbf{q}}(t), \hat{\textbf{o}}(t), r_{q}(t), r_{o}(t)$ can be divided into $M$ Bernstein polynomials. Using an $M$-segment-polynomial representation, we formulate a QP problem.   
\subsubsection{Constraints}
\label{subsubsec:constraint}
The drone's trajectory is constrained to maintain the target visibility (\ref{eq4:visibility_constraint}) and (\ref{eq4:fov_constraint}), avoid collisions (\ref{eq4:collision_constraint}), and satisfy dynamic feasibility (\ref{eq4:dynamic_constraint}). To construct the QP problem, we formulate the constraints to be affine with respect to $\underbar{\textbf{c}}$.
By leveraging the properties of Bernstein polynomials, which include \textit{P1)} the first and last coefficients corresponding to the endpoints and \textit{P2)} the convex hull property, we can effectively formulate dynamic constraints.
Also, the $l$-th derivative of an $n$-th order Bernstein polynomial is an $(n-l)$-th order polynomial. The coefficients of the $l$-th derivative of a polynomial are obtained by multiplying the coefficients of the original polynomial by the following matrix.
\begin{equation}
    \begin{aligned}
        E_{n,m}^{(l)}&=\begin{cases}
            \frac{n-l+1}{T_{m}-T_{m-1}}E_{n-l}E_{n,m}^{(l-1)} & \text{if}\ l\geq1 \\
            \textbf{I}_{(n+1)\times (n+1)} & \text{if}\ l =0
        \end{cases},\ n\geq l \geq 0, \\
        E_{k}&=  \begin{bmatrix}
            -1 & 1 & 0 &\dotsi & 0\\
            0 & \ddots & \ddots &\ddots & \vdots \\
            \vdots & \ddots & \ddots &\ddots & 0 \\
            0 & \dotsi & 0 & -1 & 1
        \end{bmatrix}\in \mathbb{R}^{k\times (k+1)}
    \end{aligned}
\end{equation}
First, the trajectory is made to satisfy the initial conditions: the current position and velocity. Using \textit{P1)}, we can get
\begin{subequations}
    \label{eq:implementation_initial_state}
    \begin{align}
    &\textbf{e}_{n_{c}+1,1}^{\top}E_{n_{c},m}^{(0)}\textbf{C}_{m}=\textbf{p}_{c0}^{\top},\\
    &\textbf{e}_{n_{c},1}^{\top}E_{n_{c},m}^{(1)}\textbf{C}_{m}=\textbf{p}_{c0}'^{\top},
    \end{align}
\end{subequations}
where $\textbf{e}_{n,l}$ is an $n \times 1$ one-hot vector where the $l$-th element is 1, and all other elements are 0. Second, by using \textit{P1)}, the $\mathcal{C}_2$ continuity between consecutive segments is achieved by the following constraints.
\begin{equation}
    \label{eq:implementation_continuity}
    \begin{aligned}
        &\textbf{e}_{n_{c}+1-l,n_{c}+1-l}^{\top}E_{n_{c},m}^{(l)}\textbf{C}_{m} = \textbf{e}_{n_{c}+1-l,1}^{\top}E_{n_{c},m+1}^{(l)}\textbf{C}_{m+1},\\
        &l = 0, 1, 2,\ m = 1,2,\ldots, M-1
    \end{aligned}
\end{equation}
Third, by using \textit{P2)}, the constraints of velocity and acceleration limits, (\ref{eq4:dynamic_constraint}), are formulated as follows.
\begin{subequations}
\label{eq:implementation_actuator_limit}
    \begin{align}
        &\|\textbf{e}_{n_{c},k}^{\top}E_{n_{c},m}^{(1)}\textbf{C}_{m}\|_{\infty}\leq \frac{1}{\sqrt{2}}v_{\max},\\
       &\|\textbf{e}_{n_{c}-1,l}^{\top}E_{n_{c},m}^{(2)}\textbf{C}_{m}\|_{\infty}\leq \frac{1}{\sqrt{2}}a_{\max},\\
       &k=1,\ldots,n_{c},\ l=1,\ldots,n_{c}-1,\ m=1,\ldots,M \nonumber
    \end{align}
\end{subequations}
Since $\underbar{\textbf{c}}$ can be acquired by flattening $[\textbf{C}_{1},\ldots,\textbf{C}_{M}]$ , the dynamic constraints (\ref{eq:implementation_initial_state})-(\ref{eq:implementation_actuator_limit}) are affine with respect to $\underbar{\textbf{c}}$. Then, the constraints in (\ref{eq:affine_constraint}) are applied to ensure the target visibility and the drone's safety.
\subsubsection{Costs}
\label{cost}
The optimization cost is divided into two terms: $J_j$ and $J_e$. First, $J_{j}$ penalizes the jerkiness of the drone's path.
\begin{equation}
    \begin{aligned}
        &J_{j} = \int_{0}^{T}\|\textbf{p}_{c}'''(t)\|_{2}^{2}dt,\\
        &\ \ \ = \sum_{m=1}^{M}\int_{0}^{T}\|\textbf{b}^{\top}_{n_{c}-3,m}(t)E_{n_{c},m}^{(3)}\textbf{C}_{m}\|_{2}^{2}dt\\
        &\ \ \  = \sum_{m=1}^{M} \text{tr}(\textbf{C}_{m}^{\top}Q_{n_{c},m}^{\int_{j}}\textbf{C}_{m}),\ \ \text{where tr($\cdot$) is a trace operator,}\\
        & Q_{n_{c},m}^{\int_{j}}=\frac{T_{m}-T_{m-1}}{2n_{c}-5}(E_{n_{c},m}^{(3)})^{\top}Q_{n_{c}}^{\int_{j}}E_{n_{c},m}^{(3)},\\
        & Q_{n_{c}}^{\int_{j}}=\{{q_{k,l}^{\int_{j}}}\}\in\mathbb{R}^{(n_{c}-2)\times(n_{c}-2)}, q_{k,l}^{\int_{j}}= \frac{\binom{n_{c}-3}{k}\binom{n_{c}-3}{l}}{\binom{2n_{c}-6}{k+l}}
    \end{aligned}
\end{equation}
Second, $J_{e}$ mimimizes tracking error to the designed reference trajectory $\tilde{\textbf{p}}_{c}^{*}(t)$.
\begin{equation}
    \begin{aligned}
        &J_{e} = \int_{0}^{T}\|\textbf{p}_{c}(t)-\tilde{\textbf{p}}_{c}^{*}(t)\|_{2}^{2}dt\\
        &\ \ \ = \sum_{m=1}^{M}\int_{0}^{T}\|\textbf{b}^{\top}_{n_{c},m}(t)(\textbf{C}_{m}-\tilde{\textbf{C}}_{m}^{*})\|_{2}^{2}dt\\
        &\ \ \  = \sum_{m=1}^{M} \text{tr}\big((\textbf{C}_{m}-\tilde{\textbf{C}}_{m}^{*})^{\top}Q_{n_{c},m}^{\int_{e}}(\textbf{C}_{m}-\tilde{\textbf{C}}_{m}^{*})\big)\\
        &\text{where}\ Q_{n_{c},m}^{\int_{e}}=\frac{T_{m}-T_{m-1}}{2n_{c}+1}Q_{n_{c}}^{\int_{e}},\\
        & Q_{n_{c}}^{\int_{e}}=\{{q_{k,l}^{\int_{e}}}\}\in\mathbb{R}^{(n_{c}+1)\times(n_{c}+1)},\ q_{k,l}^{\int_{e}}= \frac{\binom{n_c}{k}\binom{n_{c}}{l}}{\binom{2n_{c}}{k+l}}
    \end{aligned}
\end{equation}
Similarly to the constraint formulation, the cost, $J=w_{j}J_{j}+w_{e}J_{e}$, can be expressed with  \underbar{\textbf{c}} and becomes quadratic in \underbar{\textbf{c}}, where $w_{j}$ and $w_{e}$ are nonnegative weight factors.

Since all the constraints are affine, the cost is quadratic with respect to $\underbar{\textbf{c}}$, and Hessian matrices of the cost, $w_{j}Q_{n_{c},m}^{\int_{j}}+w_{e}Q_{n_{c},m}^{\int_{e}}$, $m=1, \ldots, M$, are positive semi-definite, the proposed trajectory optimization problem can be formulated as a convex QP problem. 
\begin{equation}
    \label{eq:qp_final}
    \begin{aligned}
    &\underset{\underbar{\textbf{c}}}{\text{min}} && \underbar{\textbf{c}}^{\top}Q_{c}\underbar{\textbf{c}} +\textbf{g}_{c}^{\top}\underbar{\textbf{c}} \\
    &\text{s.t.} && A_{c}\underbar{\textbf{c}}\geq\textbf{b}_{c}\\
    \end{aligned}
\end{equation}
\begin{table*}[t!]
    \vspace{2mm}
    \centering
    \caption{Computation time in QP solver}
    \label{tab:planning_time}
    \begin{tabular}{rrr|rrrrr|rrrrr|rrrrr}
    \toprule
       & Computation & & \multicolumn{5}{c}{$n_{c}=4$} & \multicolumn{5}{c}{$n_{c}=5$} & \multicolumn{5}{c}{$n_{c}=6$}\\
       & time [ms] & & \multicolumn{5}{c}{\# polynomial segment\ $(M)$} & \multicolumn{5}{c}{\# polynomial segment\ $(M)$} & \multicolumn{5}{c}{\# polynomial segment\ $(M)$}\\ 
         &  & &1 &2 & 3 & 4 & 5 &1 &2 & 3 & 4 & 5 &1 &2 & 3 & 4 & 5\\ \hline
     \multirow{4}{*}{{\rotatebox[origin=c]{90}{Single}}} 
     &\# obstacle
     & 1 & $0.17$ & $0.54$ & $\cdot$ & $\cdot$ & $\cdot$
     & $0.19$ & $0.77$ & $\cdot$ & $\cdot$ & $\cdot$
     & $0.21$ & $1.16$ & $\cdot$ & $\cdot$ & $\cdot$\\
     & $(N_o)$ & 2 & $0.18$ & $0.71$ & $2.07$  & $\cdot$ & $\cdot$
     & $0.22$ & $1.12$ & $2.95$ & $\cdot$ & $\cdot$
     & $0.24$ & $1.66$ & $4.45$ & $\cdot$ & $\cdot$\\
     & & 3 & $0.29$ & $0.98$ & $2.91$ & $6.05$ & $\cdot$
     & $0.34$ & $1.23$ & $4.58$ & $8.05$ & $\cdot$
     & $0.43$ & $1.82$ & $5.62$ & $12.3$ & $\cdot$\\
     & & 4 & $0.32$ & $1.01$ & $2.75$ & $7.81$ & $12.2$
     & $0.39$ & $1.38$ & $4.24$ & $12.6$ & $18.8$
     & $0.55$ & $1.88$ & $6.81$ & $16.8$ & $24.8$\\ 
     \midrule
     \multirow{3}{*}{{\rotatebox[origin=c]{90}{Dual}}} 
     & \# obstacle
     & 1 & $0.18$ & $0.66$ & $1.94$ & $5.34$ & $\cdot$
     & $0.20$ & $0.91$ & $2.95$ & $7.61$ & $\cdot$
     & $0.26$ & $1.38$ & $4.46$ & $10.5$ & $\cdot$\\
     & $(N_o)$& 2 & $0.21$ & $1.21$ & $3.94$  & $5.82$ & $13.6$
     & $0.46$ & $1.72$ & $6.78$ & $9.48$ & $21.4$
     & $0.55$ & $2.41$ & $10.8$ & $17.8$ & $36.3$\\
     & & 3 & $0.43$ & $1.67$ & $3.86$ & $8.54$ & $17.8$
     & $0.78$ & $2.51$ & $5.62$ & $13.5$ & $28.6$
     & $0.81$ & $3.72$ & $8.91$ & $21.8$ & $43.6$\\
     \bottomrule
    \end{tabular}%
    \vspace{-2mm}
\end{table*}
\begin{table}[t!]
    \centering
    \caption{Chasing Planner Comparison}
    \label{tab:comparison_summary}
    \begin{tabular}{c|cccc}
        \toprule
            &\cite{sphere} &\cite{Bonatti}  &\cite{Nageli} &\textbf{Proposed}\\ \midrule
          Avg. of jerk [$\text{m}/\text{s}^{3}$]  & 733.9 & 430.4 & 774.1 & \textbf{2.379} \\ \hline
          \begin{tabular}[c]{@{}c@{}}\underbar{(Safe duration)}\\ {(Total duration)}\end{tabular}
           & 1.0 & 0.986 & 0.938 & \textbf{1.0} \\ \hline
         \begin{tabular}[c]{@{}c@{}}\underbar{(Visible duration)}\\ {(Total duration)}\end{tabular}  & 0.587 & 0.092 & 0.046 & \textbf{1.0} \\ \hline
         Computation time\ [ms] & 15.5  & 5.67 & 28.7 & \textbf{0.42}\\ 
        \bottomrule
    \end{tabular}
    \vspace{-2mm}
\end{table} 
\begin{figure}[t!]
\centering
\begin{subfigure}[t]{0.2\textwidth}
\centering\includegraphics[width=\textwidth]{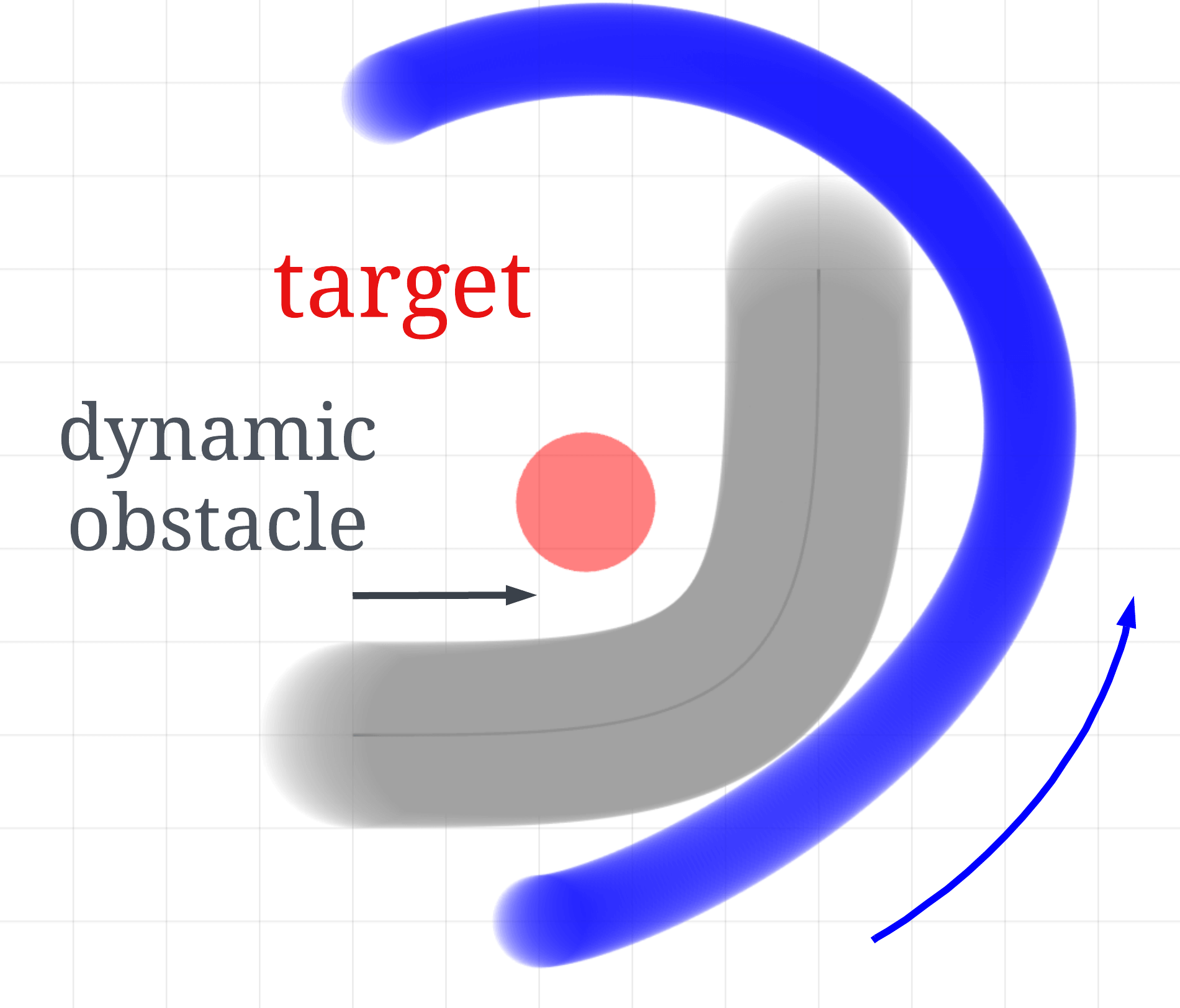}
\caption{Proposed}
\label{fig:comparison_qp}
\end{subfigure}
\hfill
\begin{subfigure}[t]{0.2\textwidth}
\centering\includegraphics[width=\textwidth]{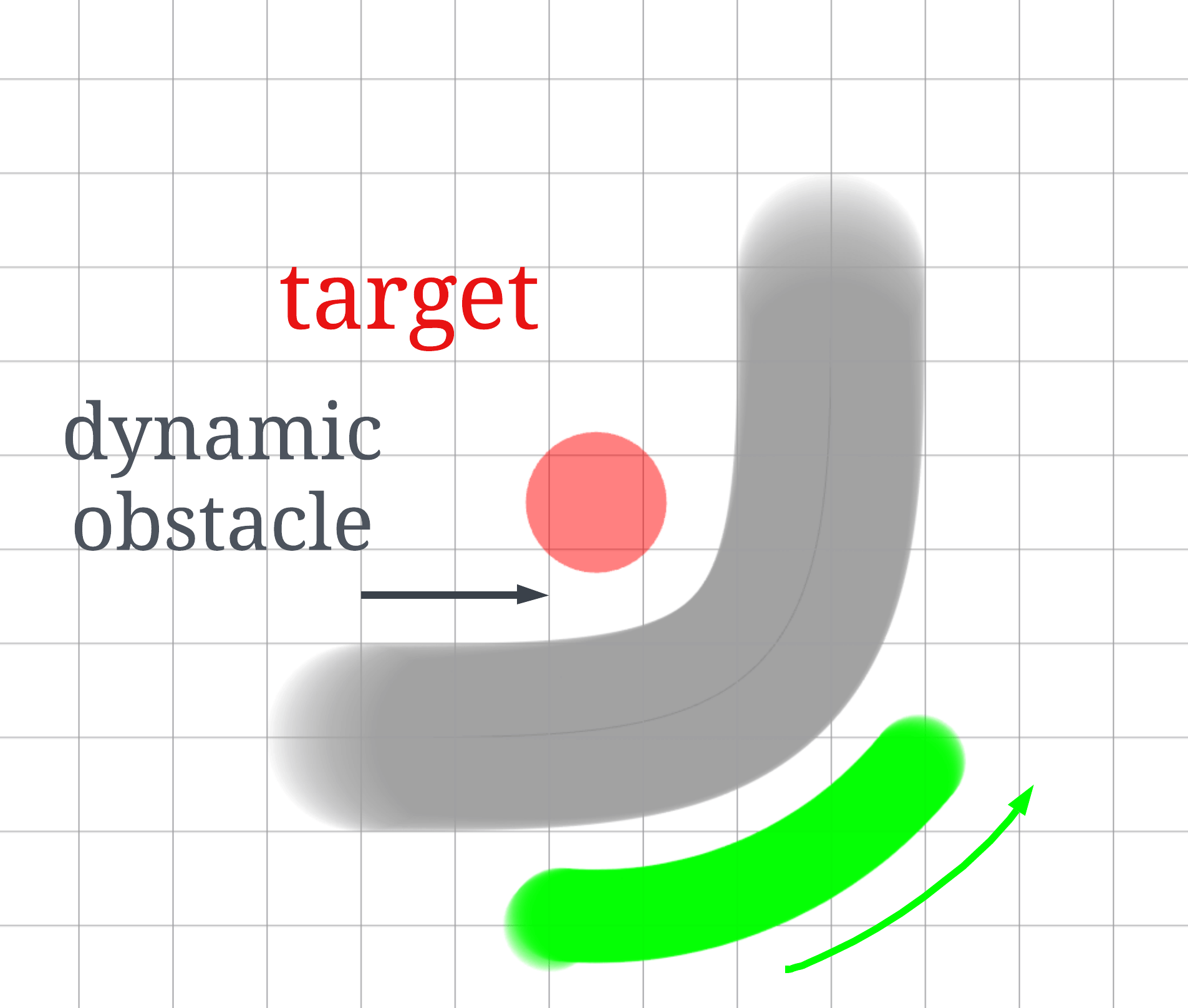}
\caption{Penin et al. \cite{sphere}}
\label{fig:comparison_penin}
\end{subfigure}
\hfill
\begin{subfigure}[t]{0.2\textwidth}
\centering\includegraphics[width=\textwidth]{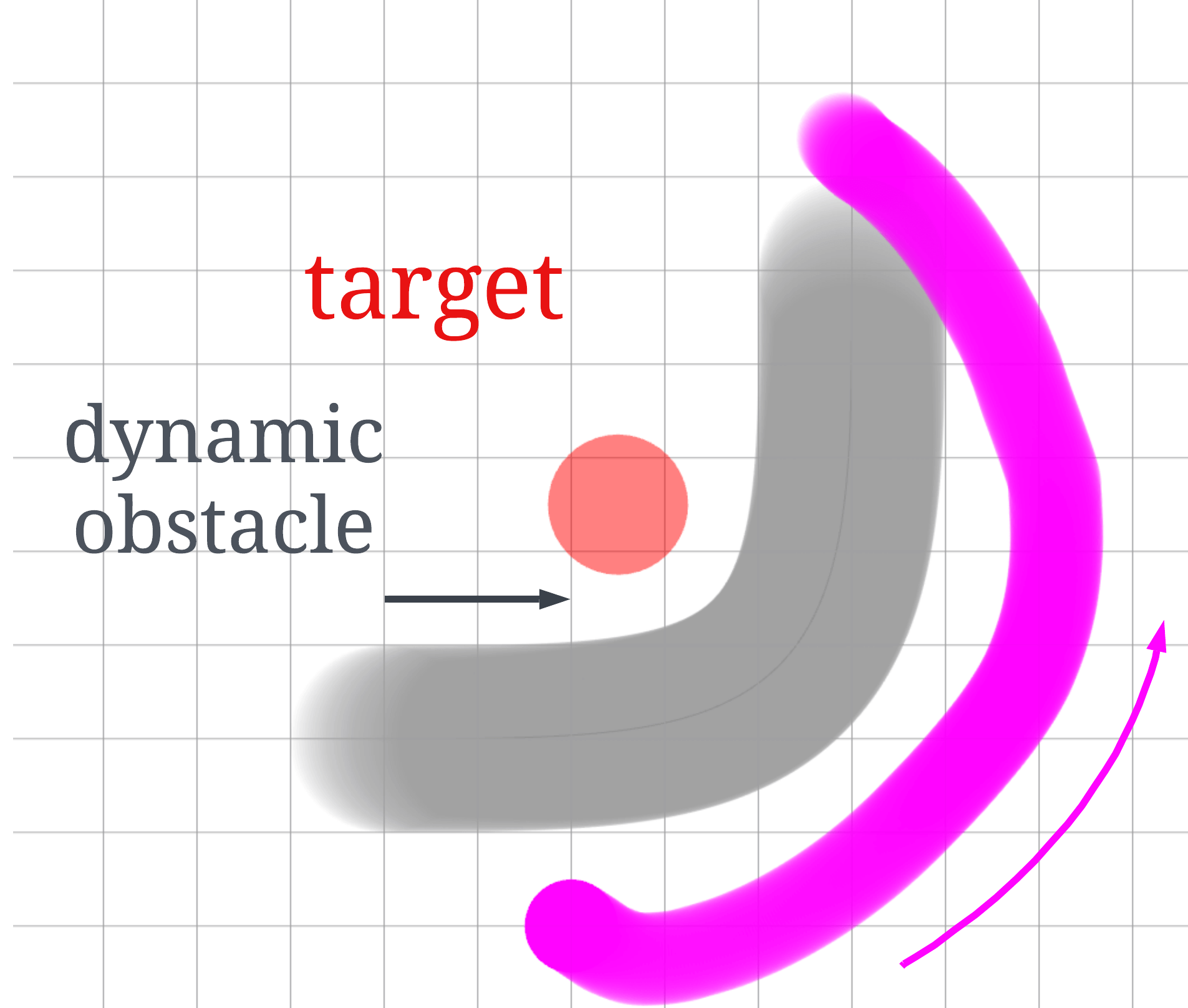}
\caption{Bonatti et al. \cite{Bonatti}}
\label{fig:comparison_bonatti}
\end{subfigure}
\hfill
\begin{subfigure}[t]{0.2\textwidth}
\centering\includegraphics[width=\textwidth]{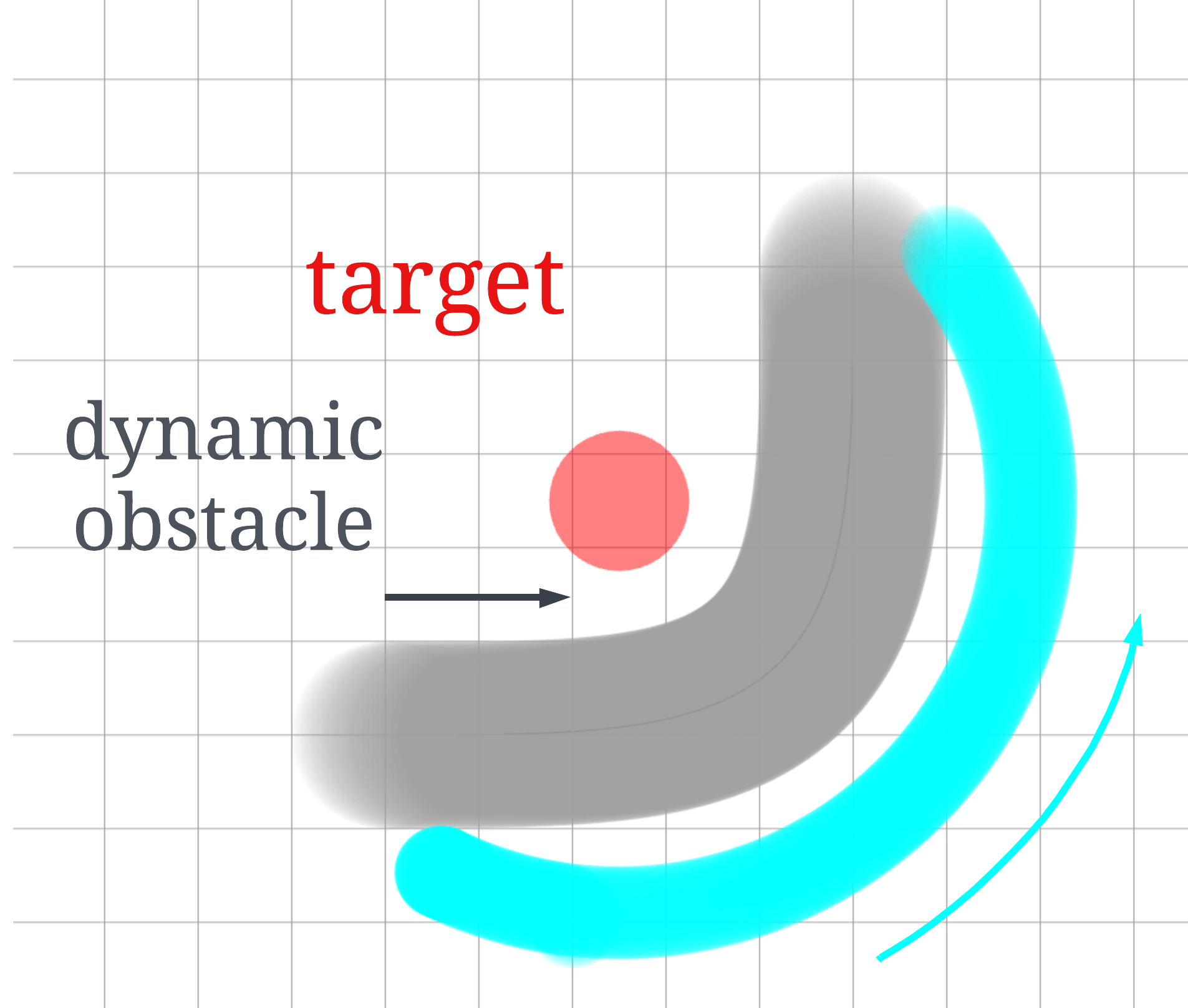}
\caption{Nägeli et al. \cite{Nageli}}
\label{fig:comparison_nageli}
\end{subfigure}
\hfill
\caption{Comparison results with relevant chasing planners: proposed (blue), \cite{sphere} (green), \cite{Bonatti} (magenta), and \cite{Nageli} (cyan).
}
\label{fig:comparison}
\end{figure}
\subsection{Evaluation}
\label{subsec:planning_evaluation}
\subsubsection{Computation time}
\label{subsubsec:computation_time}
We set $n_c$ as 4, 5, 6 and test 5000 times for different scenarios $N_{o}$ $=1,2,3,4$. qpOASES \cite{qpoases} is utilized as a QP solver, and the execution time to solve the QP problem is summarized in Table \ref{tab:planning_time}. As shown in the table, the computation time increases as the number of segments and obstacles increases.\\ 
In the realistic simulations and hardware experiments presented in Section \ref{sec:validation}, the number of polynomial segments is reported to be at most $M = 3$, and a planning frequency exceeding 40 Hz is achieved under the setting with $n_c = 6$. However, in extreme scenarios where multiple obstacles aggressively approach the target, the number of active constraints can significantly increase. Additionally, an increase in M results in a higher number of optimization variables. These factors combined may lead to a sharp increase in the planner's runtime.  
To satisfy real-time criteria (10 Hz) even as the number of close obstacles increases, we reduce $r_{q}(t)$ by defining $N_{\epsilon}$ furthest primitives as outliers, removing them from $\mathcal{P}_{s}$, and recalculating $\mathcal{R}_{\textbf{q}}(t)$. We determine the number of outliers $N_{\epsilon}=\epsilon_{p}|\mathcal{P}_{s}|$ with a factor $\epsilon_{p}$. The effect of $\epsilon_{p}$ is visualized in Fig. \ref{subfig5:ReachSet80}, and it can lower the number of segments $M$. 
Although there may be a potential reduction in responsiveness to unforeseen target movements, real-time planning is achievable even in complex and dense environments with this strategy.
\begin{figure}[t!]
   \centering
   \includegraphics[width=1.0\linewidth]{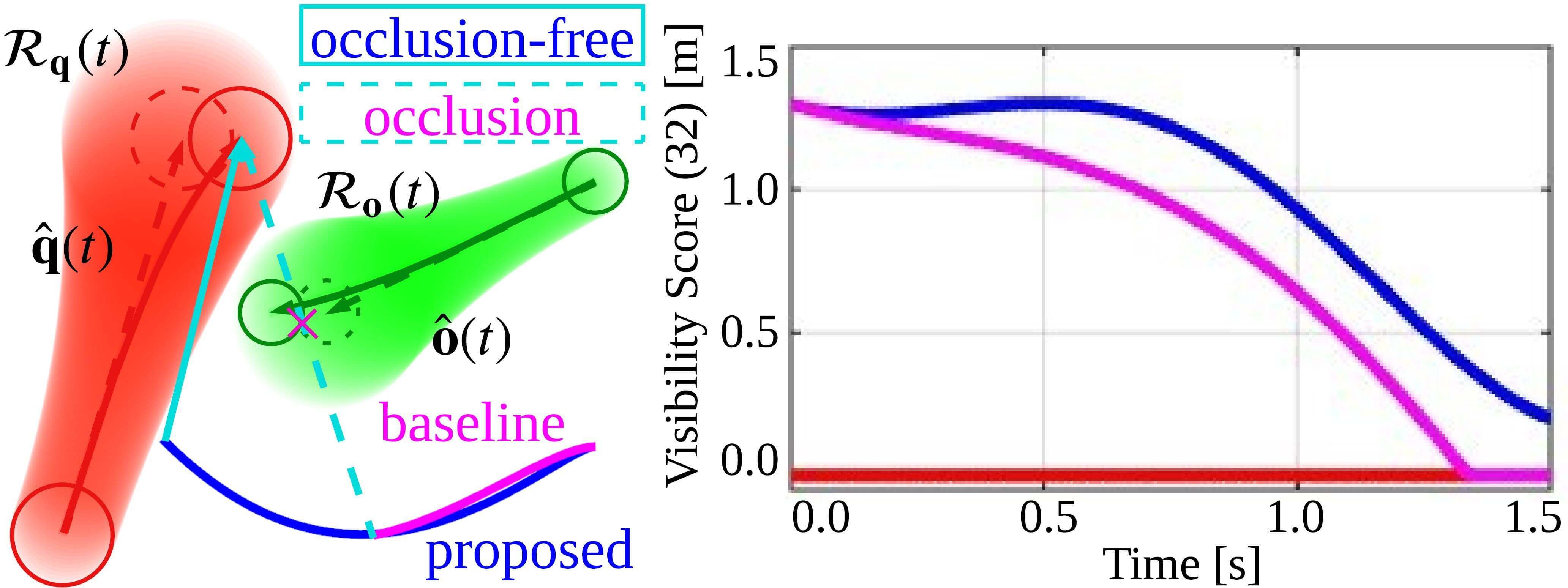}
   \caption{Comparison of planning with and without consideration of the path prediction error. \textbf{Left:} A blue spline is a generated trajectory by the proposed planner, considering $\mathcal{R}_{\textbf{q}}(t)$ (red-shaded) and $\mathcal{R}_{\textbf{o}}(t)$ (green-shaded). A magenta spline is a trajectory generated by the baseline \cite{ijcas}, which fully trusts $\hat{\textbf{q}}(t)$ (red-dashed) and $\hat{\textbf{o}}(t)$ (green-dashed).
   If a target and an obstacle move along red and green solid lines respectively, the \textit{Line-of-Sight} (cyan, solid) connecting our trajectory and the target's trajectory does not collide with the obstacle, whereas the baseline's \textit{Line-of-Sight} (cyan, dashed) collides, indicating occlusion. \textbf{Right:} The plot of visibility score (\ref{eq:visibility_metric}) when the target and obstacle move along the solid splines. A value equal to the red line represents occlusion.}
   \label{fig:uncertainty_planning}
\end{figure}
\subsubsection{Benchmark test}
\label{subsubsec:benchmark_test}
We now demonstrate the proposed trajectory planner's capability to maintain the target visibility against dynamic obstacles. The advantages of the proposed planner are analyzed by comparing it with non-convex optimization-based planners: \cite{sphere}, \cite{Bonatti}, and \cite{Nageli}. We consider a scenario where the target is static and a dynamic obstacle cuts in between the target and the drone. The total duration of the scenario is 10 seconds, and the position and velocity of the obstacle are updated every 20 milliseconds. We set $n_c$ as 5 and the planning horizon as 1.5 seconds for all planners. 
Fig. \ref{fig:comparison} visualizes the history of the positions of the target, the drone, and the obstacles, and Table \ref{tab:comparison_summary} summarizes the performance indices related to jerkiness, collisions, and occlusions.

All other nonlinear optimization-based planners fail to avoid occlusion and collision because the constraints are applied softly or the optimization converges to a sub-optimal solution. In contrast, our planner considers future obstacle movements and successfully finds trajectories by treating the drone's safety and the target visibility as hard constraints in a single QP. In addition, the computation efficiency of our method is an order of magnitude greater than other methods.

We also compare our planner with a motion-primitive-based planner \cite{ijcas} that does not account for the path prediction error. 
Fig. \ref{fig:uncertainty_planning} illustrates a situation where occlusion may occur, highlighting that the success of planning can depend on whether the prediction error is considered. Even if the results of the planning method that does not consider the error are updated quickly, occlusion can still occur if the error accumulates. In contrast, our planner generates relatively large movements, which are suitable for maintaining the target visibility when obstacles approach.
\subsubsection{Weight factor effect}
\label{subsusbsec:weight_factor_effect}
\begin{figure}[t!]
   \centering
   \includegraphics[width=0.95\linewidth]{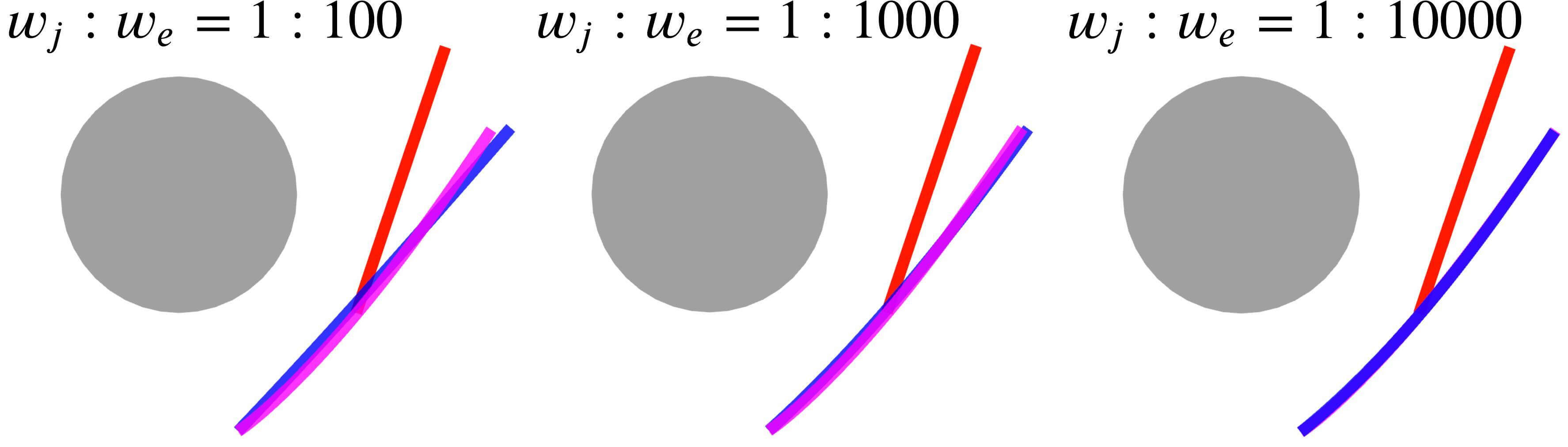}
   \caption{Effect of weight factors ($w_{j}$ and $w_{e}$) on the optimization result. The grey circle represents an obstacle, while the red, magenta, and blue splines indicate the target trajectory, reference trajectory, and optimization result, respectively.}
   \label{fig:weight_factor}
\end{figure}

We test the influence of the two weight factors, $w_{j}$ and $w_{e}$, in the QP problem (\ref{eq:qp_final}) on the tracking trajectory. Fig. \ref{fig:weight_factor} shows the resulting paths according to the ratio of the two weight factors, and Table \ref{tab6:weight_factor} presents the quantitative results.
A higher $w_{j}$ reduces jerkiness, while a higher $w_{e}$ decreases the tracking error. Although there are differences in the visibility score (\ref{eq:visibility_metric}), they are not significant enough to affect mission success. Within this range, the weights can be set according to the user’s preference.
\subsubsection{Critical analysis} \label{subsubsec:critical_analysis}
We now study the harsh situations that can fail to maintain the target visibility.
The optimization cannot be solved for the following reasons: 1) conflicts between the target visibility constraints and actuator limit constraints and 2) incompatibility among the target visibility constraints. For example, when a dynamic obstacle cuts in between the target and the drone at high speed, the huge variation in the TVR over time conflicts with the actuator limits.
Next, a situation where two dynamic obstacles cut in simultaneously from opposite directions may incur conflicts between the target visibility constraints.
These obstacles force the drone to move in opposite directions. Target occlusion inevitably occurs when the target and obstacles align. In these cases, we formulate the optimization problem considering constraints, excluding the target visibility constraint. While occlusion may occur, drone safety is the top priority, and planning considering the target visibility resumes once conditions return to normal.
\begin{table}[t!]
    \centering
    \caption{Impact of Weight Factors on Optimization}
    \label{tab6:weight_factor}
\begin{tabular}{c|ccc}
\hline
$w_{j}:w_{e}$ & $J_j$ [$\text{m}^{2}/\text{s}^{5}$] &$J_e$ [$\text{m}^{2}\cdot\text{s}$] &$\psi$ (\ref{eq:visibility_metric}) [m] (min/mean)\\ 
\hline
1:100 & 0.1721 & 0.011 & 0.5897/0.7982 \\ 
\hline
1:1000 & 3.1809 & 0.0032 & 0.6071/0.8128 \\ 
\hline
1:10000 & 9.3761 & 0.0097 & 0.6201/0.8234 \\ 
\hline
\end{tabular}
\end{table}
\section{Chasing Scenario Validation}
\label{sec:validation}
In this section, the presented method is validated through realistic simulators and real-world experiments.
\subsection{Implementation Details}
\label{subsec:implementation_details}
We perform AirSim \cite{airsim} simulations for a benchmark test of our method with \cite{ijcas} and \cite{access}.
In real-world experiments, we install two ZED2 cameras facing opposite directions.
The front-view camera is used for object detection and tracking, while the rear-view camera is used for the localization of the robot. Since the rear-view camera does not capture multiple moving objects in the camera image, the localization system is free from visual interference caused by dynamic objects, thereby preventing degradation in localization performance. Additionally, we utilize 3D human pose estimation to determine the positions of the humans. The humans in experiments are distinguished by color as described in \cite{access}. By using depth information and an algorithm from \cite{obstacle_detector}, we build a static map.
To distribute the computational burden, we use the Jetson Xavier NX to process data from the ZED2 cameras and the Intel NUC to build maps and run the QP-Chaser algorithm. To manage the payload on the drone, we utilize a coaxial octocopter and deploy Pixhawk4 as the flight controller.

The parameters used in validations are summarized in Table \ref{table:parameter}. 
As the planning result is frequently updated to respond to dynamic obstacles and targets, a long planning horizon is unnecessary; therefore, $T$ is set to a short value of 1.5 seconds.
The desired distance $r_d$ is set to 4 meters, as this value yields visually pleasing target capture with the ZED2 camera's intrinsic parameters, as determined empirically.
Additionally, we set the screen ratio, $\gamma_{c}$, to 1 to pursue the rule of thirds in cinematography. Lastly, the QP weight parameters were set to $w_e:w_j=1:1000$ based on the results in Section \ref{subsusbsec:weight_factor_effect}, in order to achieve an appropriate balance of visibility and smoothness.
\begin{figure*}[t!]
\centering
\begin{subfigure}[t!]{0.497\textwidth}
\centering \includegraphics[width=\textwidth]{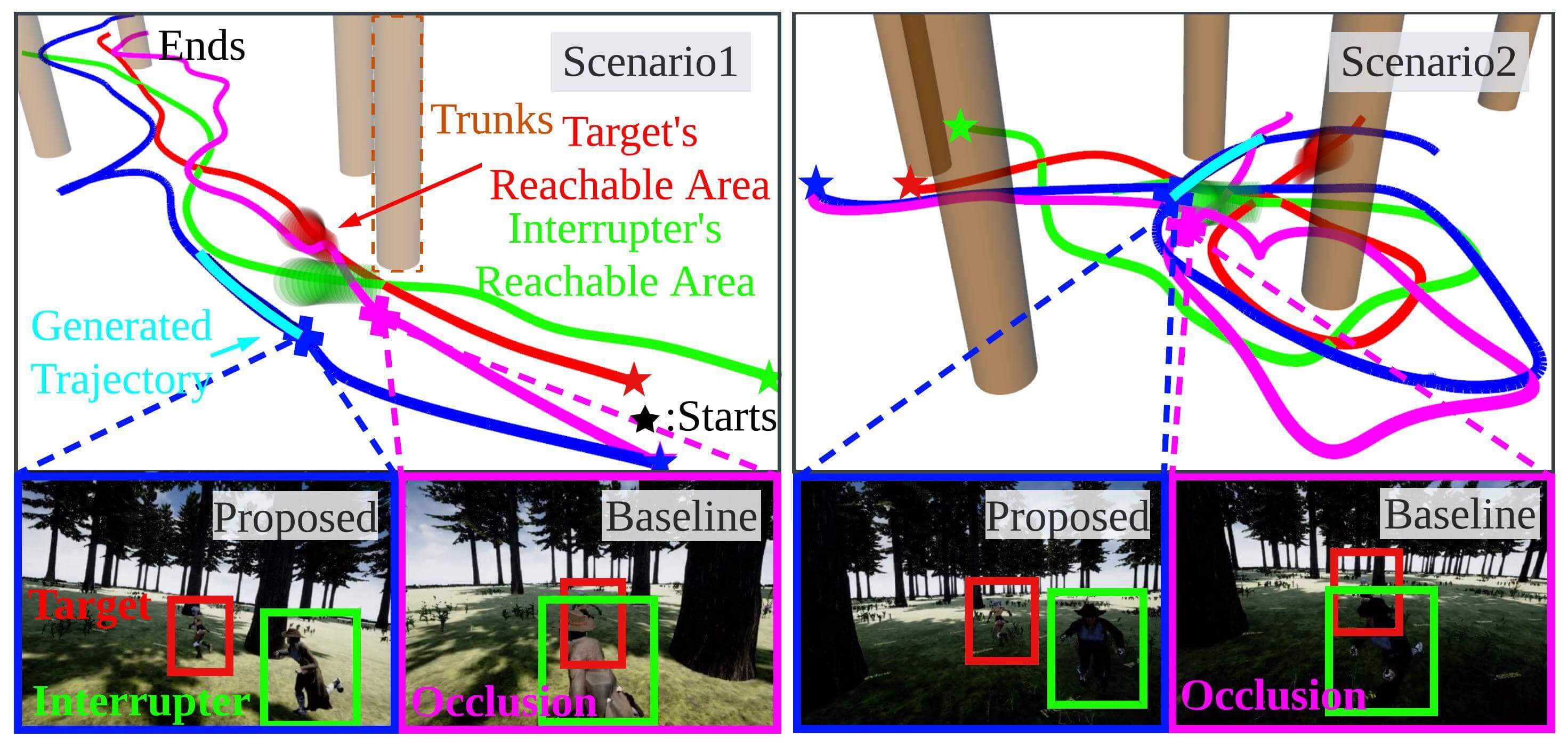}
\caption{Single target scenarios}
\label{subfig:2d_simulation}
\end{subfigure}
\begin{subfigure}[b!]{0.497\textwidth}
\centering \includegraphics[width=\textwidth]{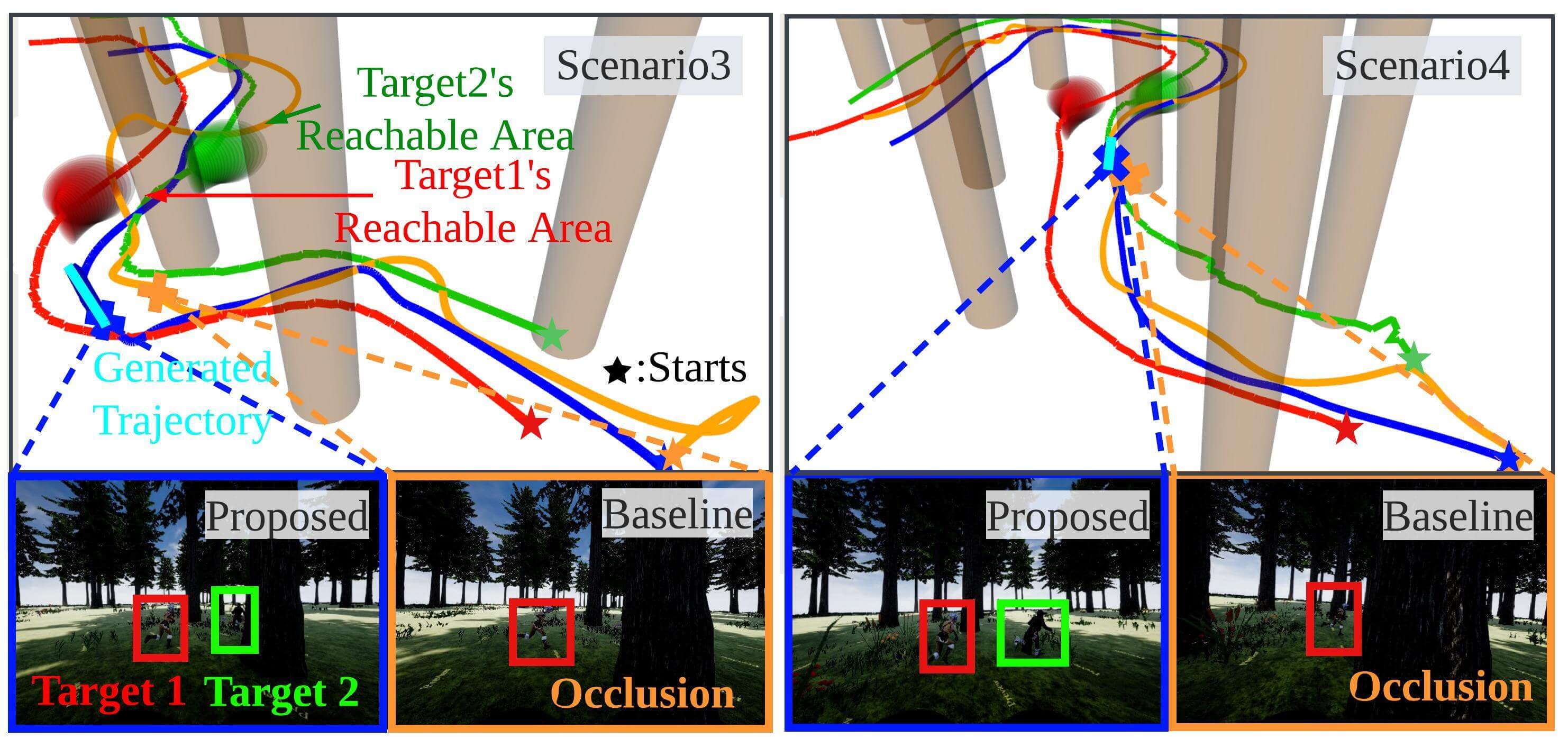}
\caption{Dual target scenarios}
\label{subfig:3d_simulation}
\end{subfigure}
\caption{Aerial tracking in AirSim environments. Scenarios 1-2 and 3-4 show simulation results for single-target and dual-target scenarios, respectively. Top figures present total position histories of moving targets and chasing drones.
The bottom figures are snapshots of the camera images when the interrupter intersects the target’s path and tries to conceal
the target’s body. The proposed planner (a blue cross) generates the trajectory (cyan) maintaining the target visibility. At the same time, \cite{ijcas} and \cite{access}
(magenta and orange crosses) fails to avoid target occlusion by the interrupter.}
\label{fig:airsim_simulation}
\end{figure*}
\begin{table}[t!]
\caption{Problem Settings}
\resizebox{\linewidth}{!}
{\begin{tabular}{lcc}
\toprule 
\multicolumn{3}{c}{\textbf{Scenario Settings}}\\
Name  &Single Target &Dual Target\\ \hline
drone radius ($r_c$) [m] & 0.4 &0.4 \\
camera FOV $(\theta_{f})$ $[{}^{\circ}]$ &120 &120 \\\hline
\multicolumn{3}{c}{\textbf{Prediction Parameters}}\\
Name  &Single Target &Dual Target\\ \hline
time horizon ($T$) [s] & 1.5 & 1.5 \\
$\#$ sampled points ($N_{samp}$) &2000 & 2000\\\hline
\multicolumn{3}{c}{\textbf{Planning Parameters}}\\
Name  &Single Target &Dual Target\\
\hline
polynomial degree $(n_{c})$ & 6 &6 \\
max velocity ($v_{\text{max}}$) [m/s] &4.0 &4.0 \\
max acceleration ($a_{\text{max}}$) [$\text{m}/\text{s}^2$] &5.0 &5.0 \\
shooting distance ($r_{d}$) [m] &$4.0$ &$\cdot$\\
screen ratio ($\gamma_{c}$) &$\cdot$ &1.0\\
tracking weight ($w_e$) &10.0 &10.0 \\
jerk weight ($w_{j}$) &0.01 &0.01 \\
\bottomrule
\end{tabular}}
\label{table:parameter}
\vspace{-2mm}
\end{table}
\subsection{Evaluation Metrics}
\label{subsec:evaluation_metrics}
In the validations, we evaluate the performance of the proposed planner using the following metrics: drone safety and target visibility. 
We measure the distances between the drone and targets (\ref{eq:distance_target}), and the minimum distance between the drone and obstacles (\ref{eq:min_dinstance_obstacle}), to evaluate drone's safety against targets and obstacles, respectively. To evaluate the target visibility, the visibility score (\ref{eq:visibility_metric}) is computed against all obstacles, and we calculate the minimum value (\ref{eq:min_visibility_score}) to assess the degree of occlusion in Cartesian coordinates, as discussed in \cite{boseong_iros}.
\begin{subequations}
    \label{eq:evaluation_metrics}
    \begin{align}
        \label{eq:distance_target}
        &\chi_{1}(t) := \|{}_{\textbf{q}}\textbf{p}_{c}(t)\|_{2}-r_{c}-r_{q}\\
        \label{eq:min_dinstance_obstacle}
        &\chi_{2}(t) := \underset{j:\mathcal{O}_{j}\in\boldsymbol{\mathcal{O}}}{\min}\|{}_{\textbf{o}_{j}}\textbf{p}_{c}(t)\|_{2}-r_{o_{j}}-r_{c}\\
        \label{eq:min_visibility_score}
        &\psi_{1}(t) := \underset{j:\mathcal{O}_{j}\in \boldsymbol{\mathcal{O}}}{\min}  \min_{\substack{\textbf{x}\in \mathcal{L}(\textbf{p}_{c}(t),\hat{\textbf{q}}(t))\\\textbf{y}\in\mathcal{O}_{j}}}
  \|\textbf{x}-\textbf{y}\|_{2}.
    \end{align}
\end{subequations}
Also, we assess the tracking performance in the image plane. First, we measure multi-object tracking accuracy (MOTA) and IDF1 to evaluate tracking accuracy. Then, we measure the visibility proportion (\ref{eq:visbility_percentage}) to evaluate the degree of occlusion caused by obstacles in the image plane. The score $\psi_{2}(t)$ means the proportion of the unobstructed part of the targets' bounding box. Since the bounding boxes can generally be obtained by most object detection algorithms, this metric is employed for generality and is formulated as follows:
\begin{equation}
    \label{eq:visbility_percentage}
    \psi_{2}(t) :=  \begin{cases} 1-\frac{b_{q}(t)\cap b_{o}(t)}{b_{q}(t)} \ (\text{if}\ \|{}_{\textbf{q}}\textbf{p}_{c}(t)\|_{2} \geq \|{}_{\textbf{o}}\textbf{p}_{c}(t)\|_{2})\\
        1 \quad \ \ \qquad \qquad \text{(otherwise)}
    \end{cases}    
\end{equation}
where $b_{q}(t)$ and $b_{o}(t)$ represent the bounding boxes of the targets and obstacles in the camera image, respectively. In the dual-target chasing scenarios, one target can be an obstacle occluding the other target, and the above metrics are measured accordingly. Throughout all validations, we use YOLO11 \cite{yolo11} for the object tracking.
\begin{table*}[t!]
    \vspace{2mm}
    \centering
    \caption{Comparison Between the Proposed Planner and Baselines (Simulation)}
    \label{tab:comparison}
    \begin{tabular}{cc|cc|cc}
    \toprule
       \multirow{2}{*}{Metrics} & \multirow{2}{*}{Planner}&\multicolumn{2}{c}{\textbf{Single Target}} &\multicolumn{2}{c}{\textbf{Dual Target}}\\
        & & Scenario 1 & Scenario 2 & Scenario 3 & Scenario 4\\
       \hline
    \multirow{2}{*}{Target Distance (\ref{eq:distance_target}) [m] $(\dagger)$} &proposed&1.595/3.726 &0.995/2.562 &1.213/2.204 &1.505/3.135\\
    &baseline &1.198/3.298 &-0.131/3.280 &1.060/3.610 &0.890/2.611\\
    \hline
    \multirow{2}{*}{Obstacle Distance (\ref{eq:min_dinstance_obstacle}) [m] $(\dagger)$}&proposed &0.872/4.070 &0.714/3.320 &0.428/1.873 &0.315/2.847\\
    &baseline &-0.264/1.715 &-0.364/2.520 &0.302/1.901 &0.059/3.207\\
    \hline
    \multirow{2}{*}{Visibility Score (\ref{eq:min_visibility_score}) [m] $(\dagger)$}&proposed &0.687/2.369 &0.028/2.529 &0.193/1.187 &0.151/1.975\\
    &baseline &0.0/1.183 &0.0/1.649 &0.0/0.834 &0.0/2.103\\
    \hline
    \multirow{2}{*}{MOTA $(\uparrow)$} &proposed
     &0.987 &0.917 &0.913 &0.934\\ &baseline
     &0.939 &0.808 &0.882 &0.879\\
    \hline
    \multirow{2}{*}{IDF1 $(\uparrow)$} &proposed
    &0.994 &0.990 &0.997 &0.997\\ &baseline
    &0.969 &0.894 &0.974 &0.972\\
    \hline
    \multirow{2}{*}{Visibility Proportion (\ref{eq:visbility_percentage}) $(\uparrow)$}&proposed &1.0/1.0 &0.523/0.996 &0.726/0.998 &0.951/0.999\\
    &baseline&0.0/0.954 &0.0/0.926 &0.0/0.942 &0.0/0.951\\
    \hline
    \multirow{2}{*}{Computation Time [ms]}&proposed
    &13.64 &15.41 &20.53 &21.22\\&baseline
    &29.45 &32.12 &79.61 &81.37\\
    \hline
    \end{tabular}%
    \vspace{1mm}
    \begin{minipage}{\textwidth}
    The upper and lower data in each metric represent the reported metrics of the proposed planner and the baselines (single target: \cite{ijcas}, dual target: \cite{access}), respectively. The values for the first, second, third, and sixth metrics indicate the (minimum/mean) performance. $\dagger$ means that a value eqaul to or below 0 indicates a collision or occlusion. $\uparrow$ means that higher is better. 
    \end{minipage}
    \vspace{-2mm}
\end{table*}
\subsection{Simulations}
\label{subsec:simulations}
We test the proposed planner in scenarios where two actors run in the forest. To show the robust performance of the planner, we bring the following four scenarios. In the first two scenarios, one actor is the target and the other is the interrupter. In the remaining two scenarios, both actors are targets.  
\begin{itemize}
    \item \textbf{Scenario 1}: The interrupter runs around the target, ceaselessly disrupting the shooting.
    \item \textbf{Scenario 2}: The interrupter intermittently cuts in between the target and the drone. 
    \item \textbf{Scenario 3}: The two targets run while maintaining a short distance between them. 
    \item \textbf{Scenario 4}: The two targets run while varying the distance between them.
\end{itemize}
\begin{table*}[t!]
    \vspace{2mm}
    \centering
    \caption{Reported Performance in Experiments}
    \label{tab:experiments}
    \begin{tabular}{c|cccc|ccc}
    \toprule
    \multirow{2}{*}{Metrics}
       & \multicolumn{4}{c}{\textbf{Single Target}} &\multicolumn{3}{c}{\textbf{Dual Target}}\\
       & Scenario S1 & Scenario S2 & Scenario S3 & Scenario S4 & Scenario D1 & Scenario D2 & Scenario D3\\
       \hline
    Target Distance (\ref{eq:distance_target}) [m] ($\dagger$)&1.668/2.303 &1.301/2.153 &1.793/2.382 &0.805/1.739 &1.336/2.037 &1.061/3.349 &1.567/2.358\\
    \hline
    Obstacle Distance (\ref{eq:min_dinstance_obstacle}) [m] ($\dagger$)&0.534/2.289 &0.701/2.285 &1.405/2.109 &0.545/1.862 &2.316/2.877 &1.094/3.277 &0.209/1.889\\
    \hline
    Visibility Score (\ref{eq:min_visibility_score}) [m] ($\dagger$)&0.393/2.111 &1.020/1.678 &1.302/1.831 &0.765/1.461 &2.403/2.809 &1.397/3.099 &0.534/2.289\\
    \hline
    MOTA ($\uparrow$)&0.998 &0.990 &0.987 &0.990 &0.991 &0.844 &0.990\\
    \hline
    IDF1 ($\uparrow$)&0.999 &0.995 &0.995 &0.993 &0.996 &0.990 &1.0\\
    \hline
    Visibility Proportion (\ref{eq:visbility_percentage}) ($\uparrow$)&1.0/1.0 &1.0/1.0 &1.0/1.0 &1.0/1.0 &1.0/1.0 &1.0/1.0 &0.901/0.995\\
    \hline
    Computation Time [ms] &12.61 &14.63 &16.12 &15.32 &14.61 &13.94 &20.81\\
    \hline
    \end{tabular}%
    \vspace{1mm}
    \begin{minipage}{\textwidth}The values for the first, second, third, and sixth metrics indicate the minimum/mean performance.
        $\dagger$ means that a value below 0 indicates a collision or occlusion. $\uparrow$ means that higher is better. 
    \end{minipage}
\end{table*}

Of the many flight tests, we extract and report some of the 30-40 seconds long flights. We compare the proposed planner with the other state-of-the-art planners \cite{ijcas} in single-target chasing scenarios and \cite{access} in dual-target chasing scenarios. 
Fig \ref{fig:airsim_simulation} show the comparative results with key snapshots, and Table \ref{tab:comparison} shows that the proposed approach makes the drone follow the target while maintaining the target visibility safely, whereas the baselines cause collisions and occlusions. There are rare cases in which partial occlusion occurs with our planner due to the limitations of the PD controller provided by the Airsim simulator. However, in the same situation, full occlusions occur with the baselines.
\begin{figure*}[t!]
    \centering
    \includegraphics[width = 0.96\textwidth]{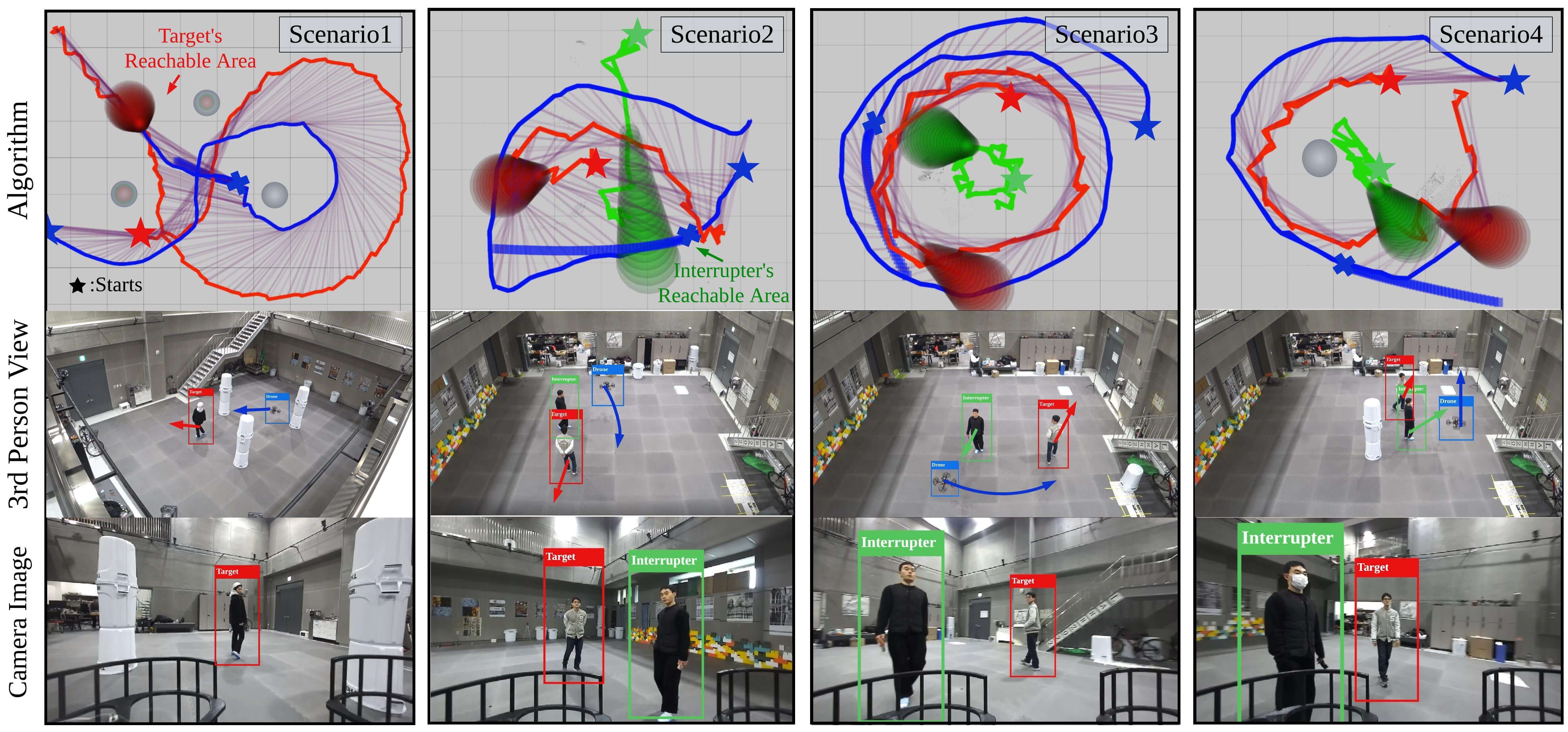}  
    \caption{Autonomous aerial tracking in an indoor environment with single-target scenarios. Blue, red, and green curves represent reported paths moved by the drone, the target, and the interrupter, respectively.
    Blue crosses mean the position of the drone at captured moments, and purple segments represent lines-of-sight between the target and the drone. The short, thick blue splines, which start from blue crosses, represent the generated trajectory for 1.5 seconds}
    \label{fig:single_exp}
    \vspace{-2mm}
\end{figure*}
\begin{figure*}[t!]
    \centering
    \includegraphics[width = 0.8\textwidth]{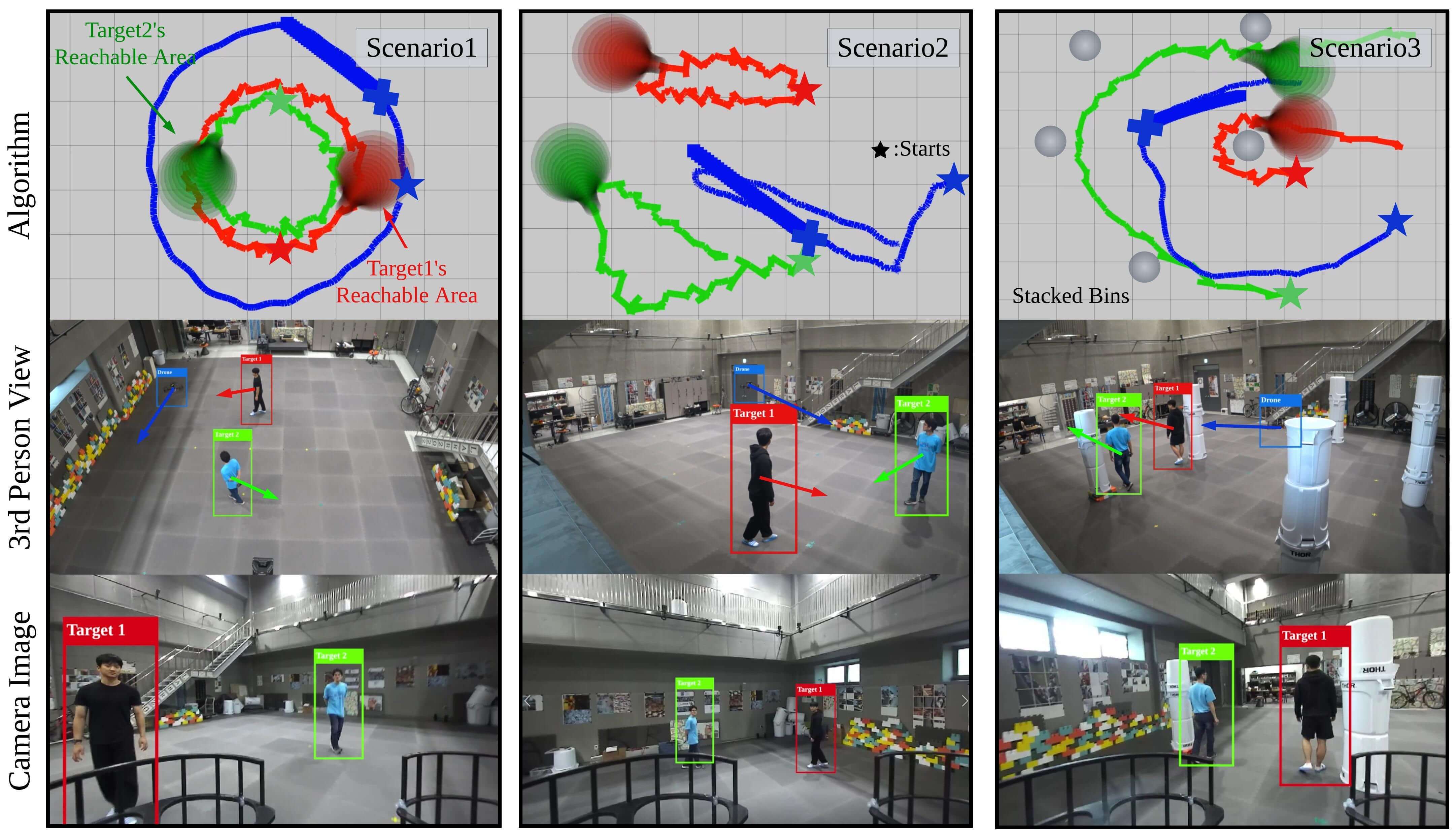}
    \caption{Autonomous aerial tracking in an indoor environment with dual-target scenarios. Blue, red, and green curves represent reported paths moved by the drone, target 1 and target 2, respectively. Blue crosses mean the position of the drone at captured moments. The short, thick blue splines, which start from blue crosses, represent the generated trajectory for 1.5 seconds.}
    \label{fig:dual_exp}
    \vspace{-4mm}
\end{figure*}

In single-target scenarios, \cite{ijcas} fails because it alters the chasing path homotopy when obstacles are close to the drone.
In dual-target scenarios, \cite{access} calculates good viewpoints considering the visibility score (\ref{eq:min_visibility_score}) and camera FOV, but the path between these viewpoints does not guarantee the target visibility. In contrast, our planner succeeds in chasing because we consider these factors as hard constraints. Moreover, despite the heavy computational demands of the physics engine, the planning pipeline is executed within 25 milliseconds, which is faster than the baselines.
\subsection{Experiments}
\label{subsec:experiments}
To validate our chasing strategy, we conduct real-world experiments with several single- and dual-target tracking scenarios.
 Two actors move in an $8\times11$ [m$^{2}$] indoor space with stacked bins. As in the simulations, one actor is the target, and the other is the interrupter in single-target scenarios, while the two actors are both targets in dual-target scenarios.\\
 For the single-target chasing, we bring the following scenarios.
\begin{itemize}
    \item \textbf{Scenario S1}: The drone senses the bins as static obstacles and follows the target.
    \item \textbf{Scenario S2}: The target moves away from the drone, and the interrupter cuts in between the target and the drone. 
    \item \textbf{Scenario S3}: Two actors rotate in a circular way, causing the target to move away from the drone, while the interrupter consistently disrupts the visibility of the target.
    \item \textbf{Scenario S4}: The interrupter intentionally hides behind bins and appears abruptly to obstruct the target.
\end{itemize}
For the dual-target chasing, we bring the following scenarios.
\begin{itemize}
    \item\textbf{Scenario D1}: The targets move in a circular pattern. To capture both targets in the camera view, the drone also moves in circles.
    \item \textbf{Scenario D2}: The targets move while repeatedly increasing and reducing their relative distance. The drone adjusts its distance from the targets to keep them within the image.
    \item \textbf{Scenario D3}: The drone detects the bins and follows the targets among them.
\end{itemize}
Figs. \ref{fig:single_exp} and \ref{fig:dual_exp} show the histories with key snapshots of the single- and dual-target experiments, respectively. The planner generates chasing trajectories in response to the targets' diverse movements and obstacles' interference. Table \ref{tab:experiments} confirms that the drone successfully followed the targets without collision and occlusion.

We use compressed RGB and depth images in the pipeline to record the camera images in limited onboard computer storage and to transfer images between computers. Since the compressed data is acquired at a slow rate, information about moving objects is updated slowly.
Therefore, as shown in Figs. \ref{fig:single_exp} and \ref{fig:dual_exp}, due to intermittent data, the planning module interprets the target as making sudden movements. Nevertheless, the drone keeps the target in view and ensures its own safety by quickly updating its trajectory, taking prediction error into account. This implicitly shows that the proposed system can effectively handle situations where the target actually exhibits abrupt motions.
\subsection{Adaptability, Scalability, and Practicality}
\label{subsec:scalability_and_adaptability}
As shown in the validations, our planner can be used in environments such as forests or crowds, where obstacles can be modeled as cylinders. Extension to environments with ellipsoidal objects is manageable, but the QP-based approach that can handle unstructured obstacles should be studied further. Extension to multi-target chasing is also feasible. As in dual-target missions, we apply occlusion avoidance and FOV constraints to all pairs of targets, but the reference trajectory to keep high visibility of all targets needs further study.

In real-world experiments, we confirmed that despite the path prediction error, the drone successfully tracked the targets by quickly updating its chasing trajectory. However, to handle greater sensor noise and more severe unexpected target behaviors, not only is a fast planning algorithm required, but also greater maneuverability of the drone. Two cameras used for stable localization and two onboard computers to meet the pipeline’s real-time constraints significantly increase the overall weight of our system. Reducing the drone's weight will be beneficial for its agility.
\section{Conclusion}
\label{Conclusion}
We propose a real-time target-chasing planner among static and dynamic obstacles. First, we calculate the reachable areas of moving objects in obstacle environments using Bernstein polynomial primitives. Then, to prevent target occlusion, we define a continuous-time target-visible region (TVR) based on path homotopy while considering the camera field-of-view limit. The reference trajectory for target tracking is designed and utilized with TVR to formulate trajectory optimization as a single QP problem. The proposed QP formulation can generate dynamically feasible, collision-free, and occlusion-free chasing trajectories in real-time. We extensively demonstrate the effectiveness of the proposed planner through challenging scenarios, including realistic simulations and indoor experiments. In the future, we plan to extend our work to chase multiple targets in environments with moving unstructured obstacles.
\appendices
\section{Implementation of Collision Check}
\label{appendix:collision_check}
The collision between the trajectories of the moving object and the $j$-th obstacle in (\ref{eq:collision_constraint_pred}) can be determined using coefficients of Bernstein polynomials.  
\begin{equation}
\label{eq:collision_const_bernstein}
    \begin{aligned}
        &\|\hat{\textbf{p}}_{i}(t)-\hat{\textbf{o}}_{j}(t) \|_{2}^{2}-(r_{p0}+r_{o_{j}}(t))^{2} \geq 0 \Longleftrightarrow\\
        &\sum_{k=0}^{2n_{p}}
        \binom{2n_p}{k}^{-1}{}^{ij}C_{k} b_{k,2n_{p}}(t;0,T) \geq 0, \ {}^{\forall}j\in \{1,\ldots,|\mathcal{O}|\}, \\
        &\text{where}\\
        &{}^{ij}C_{k}\ =  
        \sum_{l=\max(0,k-n_p)}^{\min(k,n_p)}
        \begin{pmatrix}
        n_p \\ l
        \end{pmatrix}
        \begin{pmatrix}
            n_p \\ k-l
        \end{pmatrix} 
        \Big ( {}^{ij}_{o}\textbf{p}_{(x)}[l]\ {}^{ij}_{o}\textbf{p}_{(x)}[k-l]\\
        & \quad \quad \quad \quad \quad \quad \quad+{}^{ij}_{o}\textbf{p}_{(y)}[l]\ {}^{ij}_{o}\textbf{p}_{(y)}[k-l]
        - {}^{ij}_{o}\textbf{r}[l]\ {}^{ij}_{o}\textbf{r}[k-l]
        \Big ),\\
        &{}^{ij}_{o}\textbf{p}_{(x)}=\textbf{p}_{i(x)}-\textbf{o}_{j(x)},  {}^{ij}_{o}\textbf{p}_{(y)}=\textbf{p}_{i(y)}-\textbf{o}_{j(y)}, 
        {}^{ij}_{o}\textbf{r}=\textbf{r}_{o_{j}}+r_{p0}
    \end{aligned}
\end{equation}
$[l]$ represents $l$-th elements of a vector, and $\textbf{r}_{o_{j}}$ is the Bernstein coefficient representing the radius of the $j$-th obstacle. The condition ${}^{ij}C_{k}\geq 0,\ k=0,\ldots,2n_{p}$ makes the moving objects not collide with obstacles during $[0, T]$.
\section{Proof of the Proposition1}
\label{appendix:prediction_proof}
Motion primitives for object prediction derived from (\ref{eq:primitive_optimization_problem}) are represented as follows.
\begin{equation}
\label{eq:appendix:primitive_solution}
\begin{aligned}
    &\hat{\textbf{p}}_{i}(t)= \textbf{P}_{i,n_{p}}^{\top}\textbf{b}_{n_p}(t,T)\\
    &\text{where}\ \textbf{P}_{i,n_{p}}=U_{3,n_{p}}^{\top}\textbf{P}_{i,3}^{*},\\
    &\textbf{P}_{3}^{*}=[\hat{\textbf{p}}_{0},\hat{\textbf{p}}_{0}+\frac{1}{3}\hat{\textbf{p}}_{0}'T,\frac{2}{3}\hat{\textbf{p}}_{0}+\frac{1}{3}\textbf{s}_{p,i}+\frac{1}{3}\hat{\textbf{p}}_{0}'T, \textbf{s}_{p,i}]^{\top}, \\
    &\textbf{b}_{n_p}(t,T) = [b_{0,n_{p}}(t,0,T),\ldots,b_{n_{p},n_{p}}(t,0,T)]^{\top},\ n_{p}\geq3,\\
    &U_{m,n}=\{u_{j,k}\}\in \mathbb{R}^{(n+1)\times(m+1)},\ m\geq n, \\
    & u_{j,j+k} = \frac{\binom{n}{j}\binom{m-n}{k}}{\binom{m}{j+k}}
\end{aligned}
\end{equation}
The difference between $\hat{\textbf{p}}_{i}(t)$ and $\hat{\textbf{p}}_{j}(t)$, $\|\hat{\textbf{p}}_{i}(t)-\hat{\textbf{p}}_{j}(t)\|=\|\textbf{s}_{p,i}- \textbf{s}_{p,j}\|\frac{t^{2}}{T^{2}}$. Therefore, in order to minimize distance sum in (\ref{eq:best_primitive}), it is sufficient to investigate sampled endpoints $\textbf{s}_{p,i}$.
\section{TVR Formulation}
\label{appendix:tvr_formulation}
We derive TVR-O, which are represented as (\ref{eq:TVR}) and (\ref{eq:TVR_except}), and TVR-F, which are represented as (\ref{eq:fov_constraint}).  For the mathematical simplicity, we omit time $t$ to represent variables.
\subsection{TVR-O Formulation}
\label{appendix:tvro_formulation}
 As mentioned in Section. \ref{subsec:target_visible_region}, there are two cases where the reachable areas of the target and obstacles \textbf{Case 1}: do not overlap and \textbf{Case 2}: overlap.
 \subsubsection{Case 1: Non-overlap}
 Fig. \ref{fig:tvro_1_math} shows the TVR-O, when the relation between the target, the obstacle, and the drone corresponds to \textit{Class O1}. TVR-O is made by the tangential line, and its normal vector is represented as $\textbf{n}_{qo}$. TVR-O is represented as follows:
\begin{equation}
    \label{eq:tvro_math_case1}
    \textbf{n}_{qo}^{\top}(\textbf{x}-\hat{\textbf{o}})\geq r_{o}
\end{equation}
$\textbf{n}_{qo}$ is perpendicular to the tangential line, and the line is rotated by $\theta_{qo}$ from a segment connecting the centers of the reachable areas. By using rotation matrices, $\textbf{n}_{qo}$ becomes as follows. 
\begin{subequations}
    \label{eq:tvro_math_case1_normal}
    \begin{align}
    &\textbf{n}_{qo}=R(-\frac{\pi}{2})R(\theta_{qo}) \frac{\hat{\textbf{q}}-\hat{\textbf{o}}}{\|\hat{\textbf{q}}-\hat{\textbf{o}}\|},\\
    &\sin\theta_{qo} = \frac{r_{q}+r_{o}}{\|\hat{\textbf{q}}-\hat{\textbf{o}}\|},\  \cos\theta_{qo}=\sqrt{1-\sin^{2}\theta_{qo}}
    \end{align}
\end{subequations}
By multiplying $\|\hat{\textbf{q}}-\hat{\textbf{o}}\|^{2}_{2}$ to (\ref{eq:tvro_math_case1}), we can acquire (\ref{eq:TVR}).

\subsubsection{Case 2: Overlap}
\begin{figure}[t!]
\centering
\begin{subfigure}[t]{0.20\textwidth}
\centering
\includegraphics[width=\textwidth]{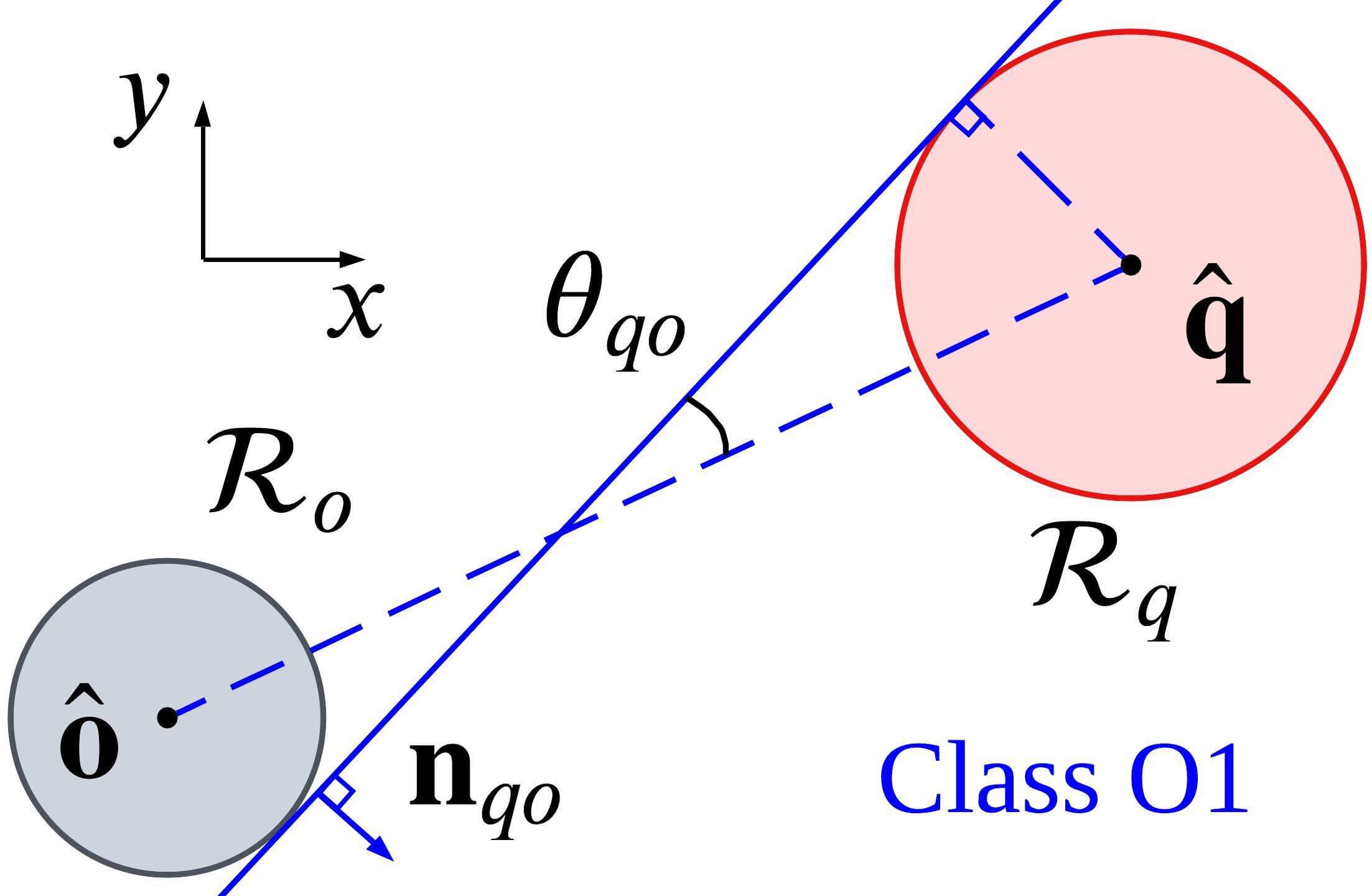}
\caption{\textbf{Case 1}}
\label{fig:tvro_1_math}
\end{subfigure}
\hfill
\begin{subfigure}[t]{0.20\textwidth}
\centering
\includegraphics[width=\textwidth]{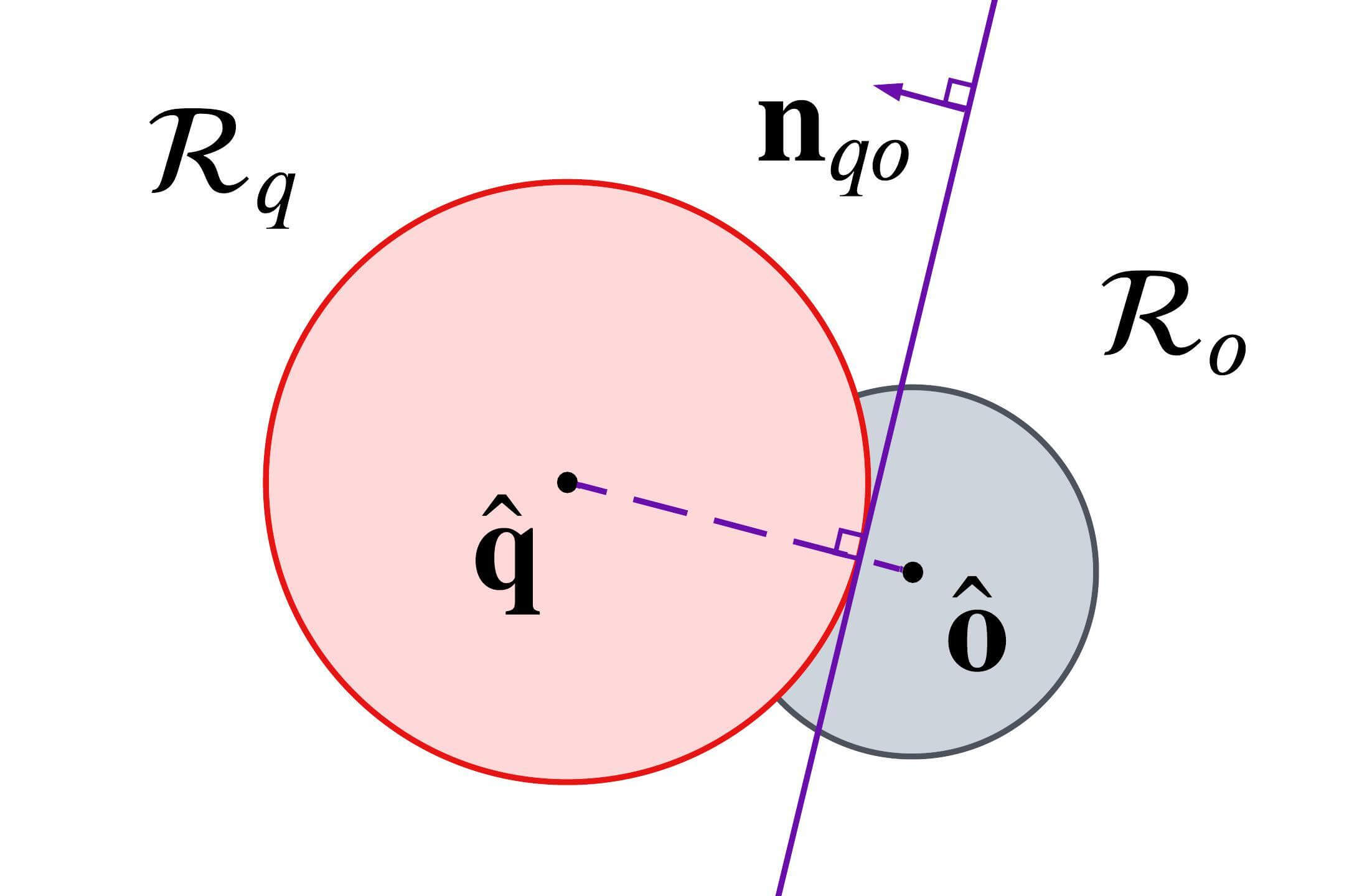}
\caption{\textbf{Case 2}}
\label{fig:tvro_2_math}
\end{subfigure}
\caption{TVR-O Formulation}
\label{fig:tvro_math}
\end{figure}

 Fig. \ref{fig:tvro_2_math} shows the TVR-O, when the reachable areas overlap. TVR-O is made by the tangential line that is perpendicular to the segment connecting the centers of the reachable areas. TVR-O is represented as follows:
 \begin{equation}
     \label{eq:tvro_math_case2}
     \textbf{n}_{qo}^{\top}(\textbf{x}-\hat{\textbf{q}}) +r_{q}\geq0,\ \textbf{n}_{qo} = \frac{\hat{\textbf{q}}-\hat{\textbf{o}}}{\|\hat{\textbf{q}}-\hat{\textbf{o}}\|}
 \end{equation}
(\ref{eq:tvro_math_case2}) is equivalent to (\ref{eq:TVR_except}).

\subsection{TVR-F Formulation}
\label{appendix:tvrf_formulation}
 \begin{figure}[t]
     \centering
     \includegraphics[width = 0.45\linewidth]{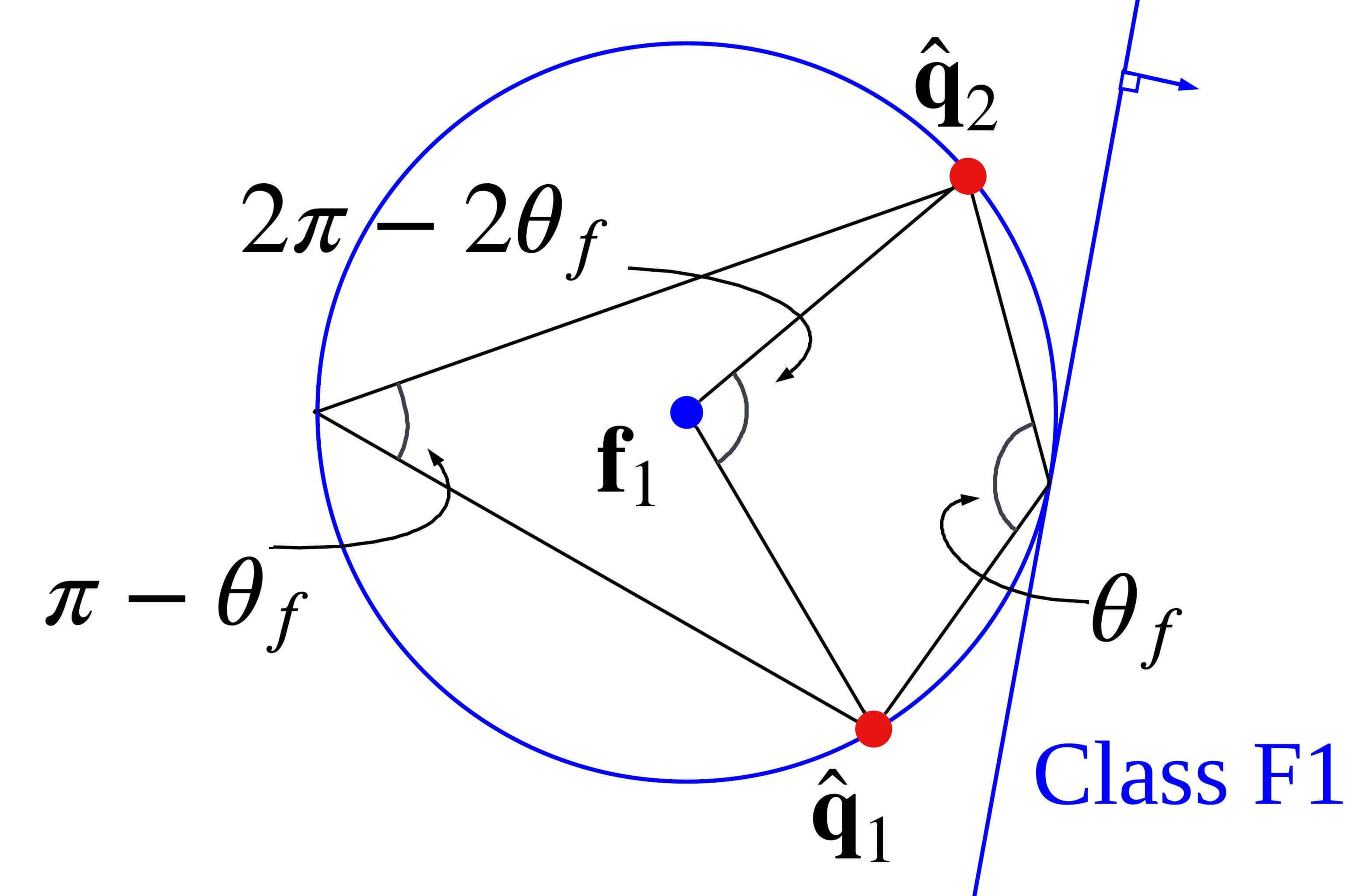}
     \caption{TVR-F Formulation}
    \label{fig:tvrf_math}
 \end{figure}
  Fig. \ref{fig:tvrf_math} shows the TVR-F, when the relation between the targets and the drone corresponds to \textit{Class F1}. The blue circle has inscribed angles of an arc tracing two points at $\hat{\textbf{q}}_{1}$ and $\hat{\textbf{q}}_{2}$ equals camera FOV $\theta_{f}$. The position of its center, $\textbf{f}_{1}$, is obtained by a relation: rotation of the segment connecting $\textbf{f}_{1}$ and $\hat{\textbf{q}}_{1}$ by an angle $2\pi-2\theta_{f}$ becomes the segment connecting  $\textbf{f}_{1}$ and $\hat{\textbf{q}}_{2}$.
  \begin{equation}
    \label{eq:tvrf_math_circle}
    R(2\pi-2\theta_{f})
    (\hat{\textbf{q}}_{1}-\textbf{f}_{1})=\hat{\textbf{q}}_{2}-\textbf{f}_{1}
\end{equation}
The angle $2\pi-2\theta_{f}$ comes from the property of the inscribed angle.  The radius of the circle. $r_{f}$, is equal to the distance between the points at $\hat{\textbf{q}}_{1}$ and $\textbf{f}_{1}$.
\begin{equation}
    \label{eq:tvrf_math_radius}
    \begin{aligned}
    r_{f} &= \|\textbf{f}_{1}-\hat{\textbf{q}}_{1}\|=\frac{1}{2}\bigg\|\begin{bmatrix}
        1 & \cot{\theta_{f}}\\-\cot{\theta_{f}} & 1
    \end{bmatrix}(\hat{\textbf{q}}_{2}-\hat{\textbf{q}}_{1})\bigg\| \\   &= \frac{1}{2\sin{\theta_{f}}}\|{}_{\hat{\textbf{q}}_{1}}\hat{\textbf{q}}_{2}\|     
    \end{aligned}
\end{equation}

TVR-F is made by the tangential line that is parallel to the segments connecting the two targets, and is represented as follows:
\begin{equation}
    \label{eq:tvrf_math}
    (\textbf{x}-\textbf{f}_{1})^{\top}
    R(-\frac{\pi}{2})\frac{\hat{\textbf{q}}_{2}-\hat{\textbf{q}}_{1}}{\|\hat{\textbf{q}}_{2}-\textbf{q}_{1}\|}\geq r_{f}
\end{equation}
By substituting $\textbf{f}_{1}$ and $r_{f}$ in (\ref{eq:tvrf_math}) with terms represented by $\hat{\textbf{q}}_{1}$ and $\hat{\textbf{q}}_{2}$, we can acquire (\ref{eq:fov_constraint}). 

\normalem
\bibliographystyle{IEEEtran}
\bibliography{my_bib}

\begin{IEEEbiography}[{\includegraphics[width=1in,height=1.25in,clip,keepaspectratio]{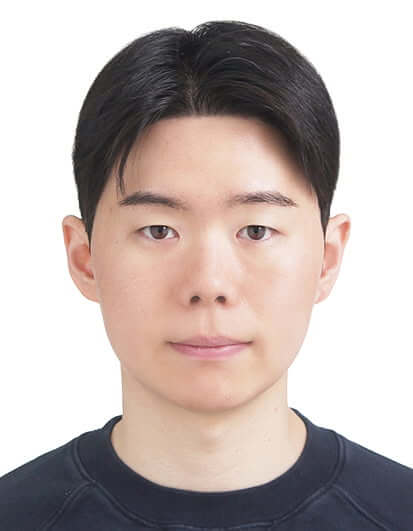}}]{Yunwoo Lee}
received the B.S. degree in electrical and computer engineering and Ph.D. degree in mechanical and aerospace engineering at Seoul National University, South Korea in 2019 and 2025, respectively. He is currently a Postdoctoral Fellow at Carnegie Mellon University, United States, after working at the Artificial Intelligence of Seoul National University, South Korea. His current research interests include aerial tracking and multi-robot systems.
\end{IEEEbiography}
\begin{IEEEbiography}[{\includegraphics[width=1in,height=1.25in,clip,keepaspectratio]{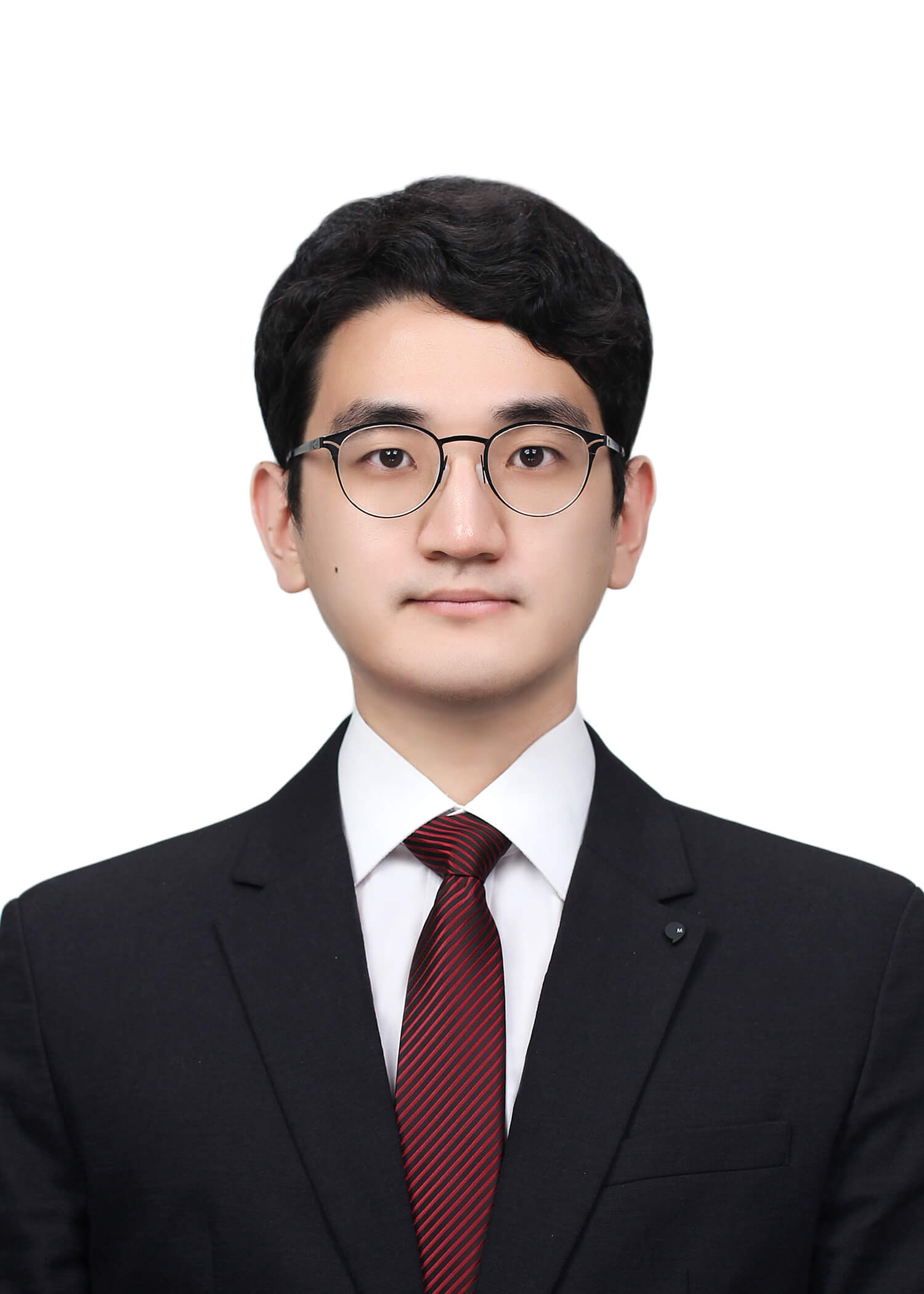}}]{Jungwon Park}
received the B.S. degree in electrical and computer engineering in 2018, and the M.S and Ph.D. degrees in mechanical and aerospace engineering at Seoul National University, South Korea in 2020 and 2023, respectively. He is currently an Assistant Professor at Seoul National University of Science and Technology, South Korea. His current research interests include path planning and task allocation for distributed multi-robot systems. His work was a finalist for the Best Paper Award in Multi-Robot Systems at ICRA 2020 and won the top prize at the 2022 KAI Aerospace Paper Award.
\end{IEEEbiography}
\begin{IEEEbiography}[{\includegraphics[width=1in,height=1.25in,clip,keepaspectratio]{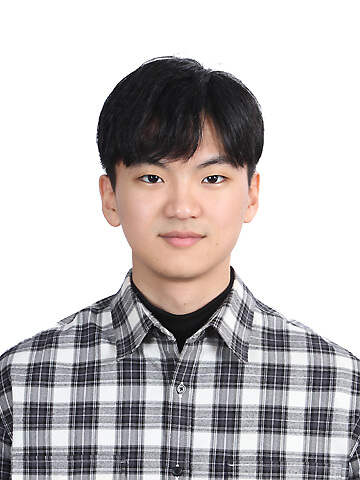}}]{Seungwoo Jung}
received the B.S. degree in mechanical engineering and artificial intelligence in 2021 from Korea University, South Korea. He is currently pursuing the intergrated M.S./Ph.D. degree in Aerospace engineering as member of the Lab for Autonomous Robotics Research under the supervision of H. Jin Kim. His current research interests include learning-based planning and control of unmanned vehicle systems.
\end{IEEEbiography}
\begin{IEEEbiography}[{\includegraphics[width=1in,height=1.25in,clip,keepaspectratio]{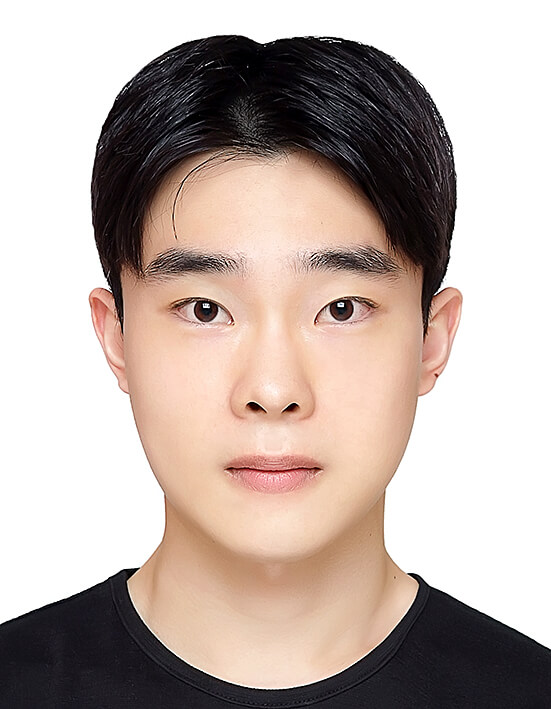}}]{Boseong Jeon}
 received the B.S. degree in mechanical engineering and  Ph.D. degrees in
aerospace engineering from Seoul National University, Seoul, South Korea, in 2017, and 2022
respectively.
He is currently a Researcher with Samsung Research, South Korea. His research interests
include VLA and planning.
\end{IEEEbiography}
\begin{IEEEbiography}[{\includegraphics[width=1in,height=1.25in,clip,keepaspectratio]{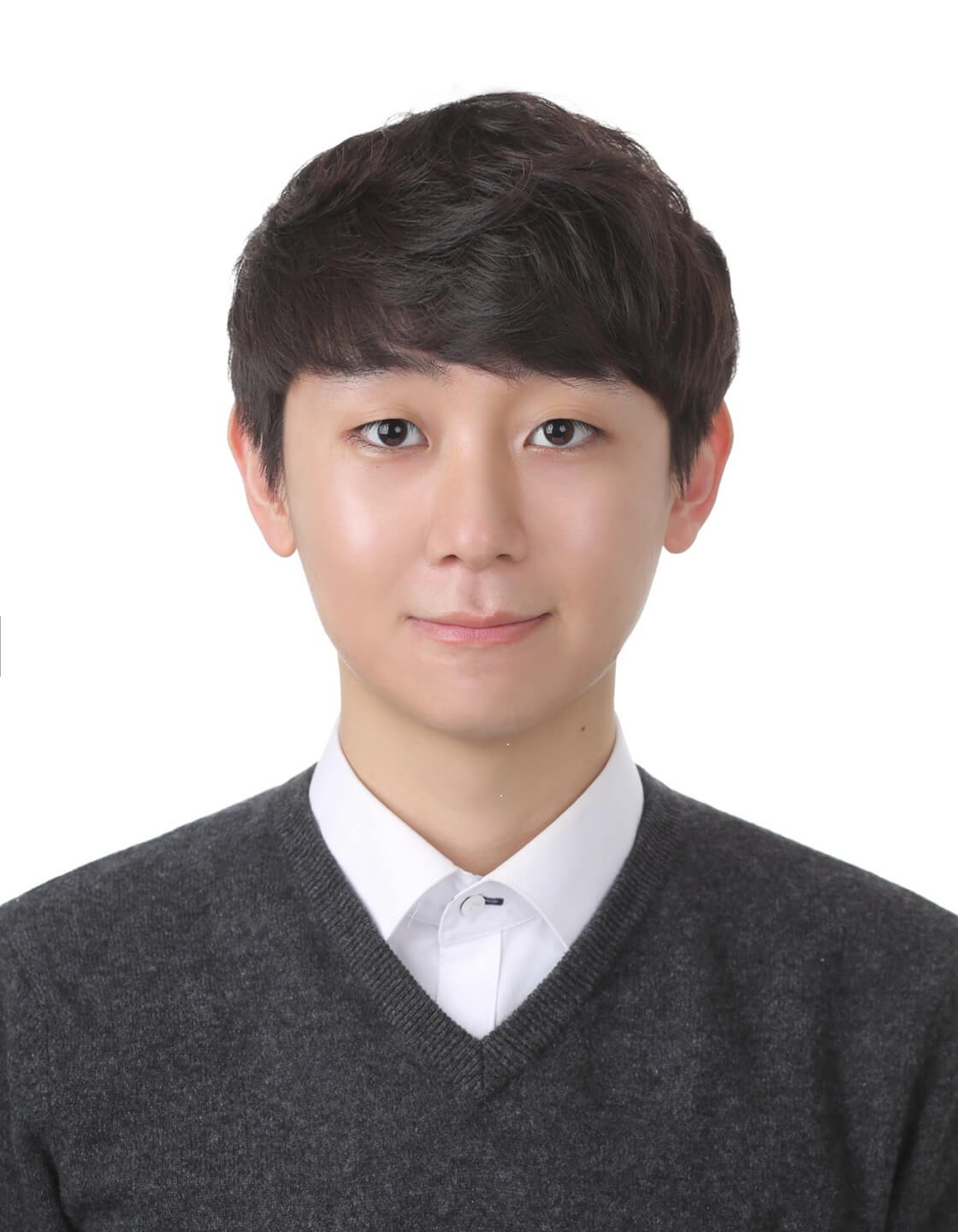}}]{Dahyun Oh}
 received a B.S. in Mechanical Engineering in 2021 from Korea University, South Korea. He is currently pursuing on an Ph.D. degree in aerospace engineering as a member of the Lab for Autonomous Robotics Research under the supervision of H. Jin Kim. His current research interests include reinforcement learning for multi-agents. 
\end{IEEEbiography}
\begin{IEEEbiography}[{\includegraphics[width=1in,height=1.25in,clip,keepaspectratio]{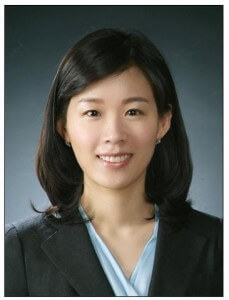}}]{H. Jin Kim}
 received the B.S. degree from the Korean Advanced Institute of Technology,
Daejeon, South Korea, in 1995, and the M.S. and Ph.D. degrees from the University of California,
Berkeley, Berkeley, CA, USA, in 1999 and 2001, respectively, all in mechanical engineering. From 2002 to 2004, she was a Postdoctoral Researcher with the Department of Electrical Engineering and Computer Science, University of California, Berkeley. In 2004, she joined the School of Mechanical and Aerospace Engineering, Seoul National University, Seoul, South Korea, where she is currently a Professor. Her research interests include navigation and motion planning of autonomous robotic systems.
\end{IEEEbiography}
\end{document}